\documentclass[10pt,twocolumn,letterpaper]{article}

\usepackage[final]{iccv}      %

\newcommand{\ours}{LaMAria}  %

\usepackage{graphicx}
\usepackage{booktabs}  %
\usepackage[accsupp]{axessibility}  %
\usepackage[utf8]{inputenc} %
\usepackage{amsfonts}       %
\usepackage{nicefrac}       %
\usepackage{microtype}      %
\usepackage{gensymb}
\usepackage{tabularx}
\usepackage{multirow}
\usepackage{svg}
\usepackage{cuted}
\usepackage{tabu}
\usepackage{pifont}  %
\usepackage{mwe}
\usepackage{bm}  %
\usepackage{textcomp}
\usepackage{makecell}  %
\usepackage{grffile} %
\usepackage{rotating}  %
\usepackage[percent]{overpic} %

\usepackage{colortbl}

\usepackage{siunitx}  %
\renewcommand{\*}[1]{\bm{\mathrm{#1}}}

\newcommand{\red}[1]{\textcolor{red}{#1}}

\newcommand{\blue}[1]{\textcolor{blue}{#1}}

\newcommand{\mytilde}{{\raise.17ex\hbox{$\scriptstyle\sim$}}}

\renewcommand{\paragraph}[1]{\vskip4pt \noindent\textbf{#1}}

\newcommand{\fail}{$\times$}%
\newcommand{\na}{--}%
\newcommand{\mono}[1]{\textcolor{orange}{#1}}

\newcommand{\monoinertial}[1]{\textcolor{OliveGreen}{#1}}
\newcommand{\binoinertial}[1]{\textcolor{blue}{#1}}

\definecolor{tabfirst}{rgb}{1, 0.7, 0.7} %
\definecolor{tabsecond}{rgb}{1, 0.85, 0.7} %
\definecolor{tabthird}{rgb}{1, 1, 0.7} %

\newlength{\pwidth}
\newlength{\lwidth}

\newlength{\bwidth}
\newlength{\iwidth}

\newcommand{\transp}{^{\top}}
\newcommand{\real}{\mathbb{R}}

\newcommand{\camproj}[1]{\Pi\left(#1\right)}
\newcommand{\covar}{\*\Sigma}
\newcommand{\residual}{\*r}

\newcommand{\norm}[1]{\left\lVert#1\right\rVert}

\newcommand{\set}[1]{\{ #1 \}} %

\newcommand{\pose}[2]{{}_{#1}\*T_{#2}}

\newcommand{\camera}{\*C}

\newcommand{\localf}{\mathsf{L}}
\newcommand{\worldf}{\mathsf{W}}
\newcommand{\pointw}{{\*P}}
\newcommand{\pointim}{{\*p}}
\newcommand{\obs}[1]{\mathcal{V}(#1)}
\newcommand{\imset}{\mathcal{I}}

\newif\ifrebuttal
\rebuttaltrue
\usepackage{lipsum}

\definecolor{cvprblue}{rgb}{0.21,0.49,0.74}
\usepackage[pagebackref,breaklinks,colorlinks,allcolors=cvprblue,hyperfootnotes=false]{hyperref}

\def\paperID{8642} %
\def\confName{ICCV}
\def\confYear{2025}

\newif\ifproceedings  %
\proceedingstrue
\newif\ifaddappendix
\addappendixtrue

\begin{document}

\title{Benchmarking Egocentric Visual-Inertial SLAM at City Scale}

\author{Anusha Krishnan$^{1}$\thanks{indicates equal contribution}\hspace{.12in}
Shaohui Liu$^{1}$\footnotemark[1]\hspace{.12in}
Paul-Edouard Sarlin$^{2}$\footnotemark[1]\hspace{.12in}
Oscar Gentilhomme$^{1}$\hspace{.12in}\\
David Caruso$^{3}$\hspace{.12in}
Maurizio Monge$^{3}$\hspace{.12in}
Richard Newcombe$^{3}$\hspace{.12in}
Jakob Engel$^{3}$\hspace{.12in}
Marc Pollefeys$^{1,4}$
\vspace{0.05in}\\
$^{1}$ETH Zurich\hspace{.2in} $^{2}$Google\hspace{.2in} $^{3}$Meta Reality Labs Research\hspace{.2in} $^{4}$Microsoft Spatial AI Lab
}

\twocolumn[{%
\renewcommand\twocolumn[1][]{#1}%
\maketitle
\vspace{-20pt} %
\begin{center}
    \centering
    \includegraphics[width=0.90\linewidth]{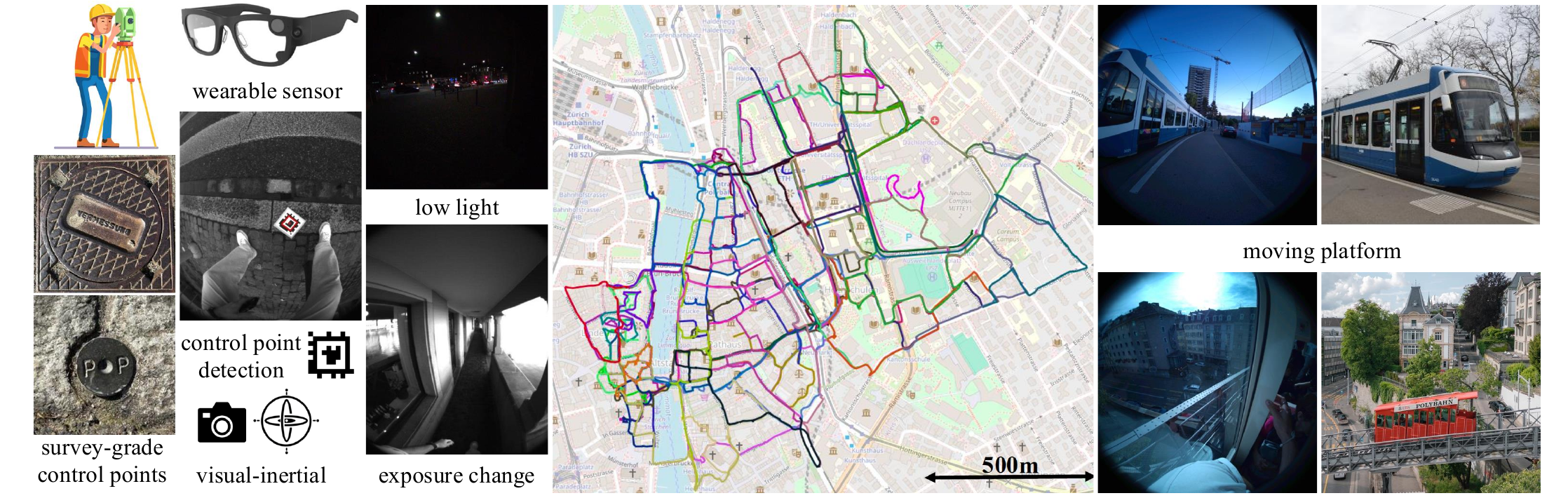} 
    \captionsetup{type=figure}
    \caption{
    \textbf{We introduce a dataset to benchmark multi-sensor VIO/SLAM with egocentric data at city scale.}
    We record hours and kilometers of trajectories with Project Aria devices~\cite{aria}, with pose annotations derived from centimeter-accurate surveyed control points.
    We cover unique challenges of egocentric data, including low light, exposure changes, moving platforms, time-varying calibration, \etc.
    }\label{fig:teaser}%
\end{center}

}]
{
  \renewcommand{\thefootnote}%
    {\fnsymbol{footnote}}
  \footnotetext[1]{indicates equal contribution}
}

\begin{abstract}
Precise 6-DoF simultaneous localization and mapping (SLAM) from onboard sensors is critical for wearable devices capturing egocentric data, which exhibits specific challenges, such as a wider diversity of motions and viewpoints, prevalent dynamic visual content, or long sessions affected by time-varying sensor calibration.
While recent progress on SLAM has been swift, academic research is still driven by benchmarks that do not reflect these challenges or do not offer sufficiently accurate ground truth poses.
In this paper, we introduce a new dataset and benchmark for visual-inertial SLAM with egocentric, multi-modal data.
We record hours and kilometers of trajectories through a city center with glasses-like devices equipped with various sensors.
We leverage surveying tools to obtain control points as indirect pose annotations that are metric, centimeter-accurate, and available at city scale.
This makes it possible to evaluate extreme trajectories that involve walking at night or traveling in a vehicle.
We show that state-of-the-art systems developed by academia are not robust to these challenges and we identify components that are responsible for this.
In addition, we design tracks with different levels of difficulty to ease in-depth analysis and evaluation of less mature approaches.
The dataset and benchmark are available at \href{https://www.lamaria.ethz.ch}{lamaria.ethz.ch}.

\end{abstract}
\vspace{-28pt} %
\newcommand{\good}{\cellcolor{green!25}}%
\newcommand{\bad}{\cellcolor{red!25}}%
\newcommand{\ok}{\cellcolor{gray!25}}%
\begin{table*}[t]
\centering
\scriptsize
\setlength\tabcolsep{3pt}%
\begin{tabular}{lccccccccccc}
\toprule
\multirow{2}{*}{dataset} & \multicolumn{3}{c}{data} & \multicolumn{3}{c}{sensors} & \multicolumn{2}{c}{ground-truth} & \multicolumn{3}{c}{challenges}\\
\cmidrule(lr){2-4} \cmidrule(lr){5-7} \cmidrule(lr){5-7} \cmidrule(lr){8-9} \cmidrule(lr){10-12}
& motion & environment & multi-seq & multi-cam & IMU & others & source & accuracy & duration & dynamics & lighting\\
\midrule
EuRoC~\cite{euroc} & \bad drone & \bad small & no\bad & yes\good & yes\good & no 
& mocap & $\sim$cm\good & $<$\qty{3}{\minute}\bad & no\bad & partial\ok\\
TartanAir~\cite{tartanair} & random\bad & \good large & no\bad & yes\good & no\bad & depth,LiDAR 
& synthetic & perfect\good & $\rightarrow$\qty{20}{\minute}\good  & yes\good & yes\good\\
4Seasons~\cite{wenzel2020fourseasons} & \bad car & \good large & \good yes & yes\good & yes\good & GNSS 
& VI+GNSS & $>$dm\bad & \qty{350}{\kilo\meter}\good & moderate\ok & yes\good\\
VBR~\cite{vbr} & \bad car,handheld & \good large & partial\ok & yes\good & yes\good & LiDAR, GNSS
& \makecell{LiDAR+IMU\\+RTK-GNSS} & $\sim$cm\good & $\rightarrow$\qty{50}{\minute}\good & moderate\ok & partial\ok\\
\midrule
TUM-RGBD~\cite{sturm2012benchmark} & \ok handheld & small\bad & no\bad & no\bad & no\bad & depth 
& mocap & $<$cm\good & $<$\qty{3}{\minute}\bad & no\bad & no\bad\\
TUM-VI~\cite{tumvi} & \ok handheld & \ok medium & no\bad & yes\good & yes\good & no
& mocap & $<$cm\good & $\rightarrow$\qty{25}{\minute}\good & no\bad & no\bad\\
ADVIO~\cite{advio} & \ok handheld & \ok medium & \good yes & no\bad & yes\good & no & VI-SLAM & $\sim$dm\bad & $\sim$\qty{3}{\minute}\bad & \ok moderate & \bad no \\
ETH3D-SLAM~\cite{badslam} & \ok handheld & \bad small & \bad no & yes\good & yes\good & depth 
& mocap & $<$cm\good & $<$\qty{4}{\minute}\bad & moderate\ok & yes\good\\
NewerCollege~\cite{newercollege,zhang2021multicamera} & handheld\ok & \ok medium & yes\good & yes\good & yes\good 
& LiDAR & LiDAR-SLAM & $\sim$cm\good & $\rightarrow$\qty{26}{\minute}\good & no\bad & no\bad \\
Hilti-Oxford~\cite{hilti2022} & handheld\ok & \ok medium & \good yes & \good yes & \good yes & LiDAR & surveying & $<$cm\good & $\rightarrow$\qty{17}{\minute}\good & \bad no & \ok partial \\
Hilti-UZH~\cite{hilti2023} & \bad robot,handheld & \ok medium & yes\good & yes \good & yes \good & LiDAR & surveying & $<$cm\good & $\rightarrow$\qty{12}{\minute}\good & \bad no & \ok partial \\
\midrule
LaMAR~\cite{sarlin2022lamar} & \good\makecell{head-mounted\\handheld} & \ok medium & yes\good & yes\good & \ok uncalibrated & GNSS, WiFi, BT
& \makecell{V-SLAM\\+LiDAR} & $\sim$dm\bad & $\sim$\qty{5}{\minute}\bad & \ok moderate & yes\good\\
\bf{\ours~(ours)} & \good\makecell{head-mounted\\handheld} & large\good & yes\good & yes\good & yes ($\times$2)\good & GNSS, WiFi, BT 
& surveying %
& $\sim$cm\good & $\rightarrow$\qty{45}{\minute}\good & \makecell{people,tram\\funicular}\good & yes\good\\
\bottomrule
\end{tabular}
\caption{\textbf{Overview of existing datasets.} \ours\ is the first egocentric dataset that is recorded in city-scaled large environments, has multiple calibrated camera and IMU sensors, includes long sequences up to 48 minutes, and covers all challenges including dynamic environments, moving platforms, and varying lighting conditions, while still providing centimeter accurate pose annotations from surveying. 
}
\label{tbl:datasets}%
\end{table*}

\section{Introduction}
Estimating the precise location of a camera over time is a fundamental problem in computer vision.
Algorithms like Visual-Inertial Odometry (VIO) or Simultaneous Localization and Mapping (VI-SLAM) can estimate a 6 Degrees-of-Freedom (DoF) pose for each image of a sequence, often aided by inertial sensors.
Positioning plays a crucial role in ensuring the persistence of digital content over time and enabling seamless sharing across devices, which is especially important for applications like AI assistants and augmented reality.
Progress in mobile computing has fueled the development of wearable devices that are equipped with various sensors, including multiple color or depth cameras, inertial units, and radio receivers.
The egocentric, multi-modal data that they capture presents unique challenges that are often overlooked in computer vision research, which typically relies on curated datasets with controlled viewpoints and motions tailored to algorithms or visual content of interest.

Differently, egocentric data is passive and accidental: it does not constrain the user's actions but rather endures them.
As a result, this data exhibits significantly more diversity in motion patterns, viewpoints, and environments than typically found in computer vision datasets.
Moreover, egocentric devices aspire to be all-day wearables that capture data over extended durations, in which factors like sensor calibration can change over time.
Finally, the wearability and consumer adoption of these devices limits the size, weight, and cost of these sensors, and thus their quality.

Academic research in VIO/SLAM is mainly driven by benchmarks that do not exhibit the characteristics of egocentric data.
Often originating from the robotics community~\cite{wenzel2020fourseasons,euroc}, their data is recorded by expensive, industrial-grade sensors mounted on robots with limited locomotion capabilities.
The robot's motion can also often be adapted to accommodate the limitations of the perception algorithms, as in active perception.
Additionally, datasets that offer sufficiently accurate ground-truth (GT) camera poses are often limited to smaller environments than the ones found in egocentric applications~\cite{tumvi,badslam,hilti2022,hilti2023}.
The datasets that offer egocentric data~\cite{sarlin2022lamar,lv2024aria,banerjee2024introducing,ma2024nymeria} do not have sufficiently accurate GT poses to measure improvements in VIO/SLAM algorithms without saturation.

In this paper, we introduce \ours, a new dataset and benchmark\footnote{All data collection, storage, and hosting was performed by ETH Zurich.} to track progress in egocentric SLAM (\cref{fig:teaser}).
We record data with Project Aria devices~\cite{aria}, which capture rich multi-sensor streams in a glasses-like form-factor, such that they can be worn over extended durations and distances without impeding the wearer's motion.
The dataset thus exhibits all key characteristics of egocentric data, with a focus on challenges that break existing algorithms: extremely low illumination, fast motion, large distances, transition between indoors and outdoors, time-varying calibration, and dynamic content -- the wearer's own body, other people, or even moving environments such as elevators and vehicles.
The trajectories cover the large area of a city center, with some of them spanning kilometers.
They benefit from a metric, centimeter-accurate ground-truth based on sparse control points (CPs) widely used in the surveying community.

We evaluate state-of-the-art VIO/SLAM systems with over \qty{22}{\hour} and \qty{70}{\kilo\meter} of egocentric data under different sensor configurations and across different difficulty levels and types of challenges. 
Our results suggest that the top methods developed by academia are still far from solving this benchmark, while exhibiting a significant gap against 
Aria's SLAM API.
Additional sequences offer gradually increasing difficulty levels between controlled hand-held motion, as exhibited by most academic datasets, and challenging unrestricted head-mounted motion.
All evaluated methods perform well with controlled motion but significantly break down as it becomes more natural and egocentric.

Our results shed light on the limitations of existing systems while our dataset opens new avenues for multi-sensor SLAM. 
The dataset and benchmark is publicly released to ease tracking progress in this direction.

\section{Related work}
\label{sec:related-work}%

\begin{figure*}[tb]
\centering
\setlength{\pwidth}{0.002\linewidth}
\setlength{\bwidth}{0.01\linewidth}
\setlength{\iwidth}{\dimexpr(0.999\linewidth - 4\pwidth - 3\bwidth)/8 \relax}
\setlength{\lwidth}{\dimexpr(0.999\linewidth - 3\bwidth)/4 \relax}
\begin{minipage}[b]{\lwidth}
\centering{\small long outdoor trajectories}
\end{minipage}%
\hspace{\bwidth}%
\begin{minipage}[b]{\lwidth}
\centering{\small exposure changes}
\end{minipage}%
\hspace{\bwidth}%
\begin{minipage}[b]{\lwidth}
\centering{\small low light}
\end{minipage}%
\hspace{\bwidth}%
\begin{minipage}[b]{\lwidth}
\centering{\small moving platforms}
\end{minipage}%
\vspace{-3mm}%

\includegraphics[height=\iwidth,angle=-90]{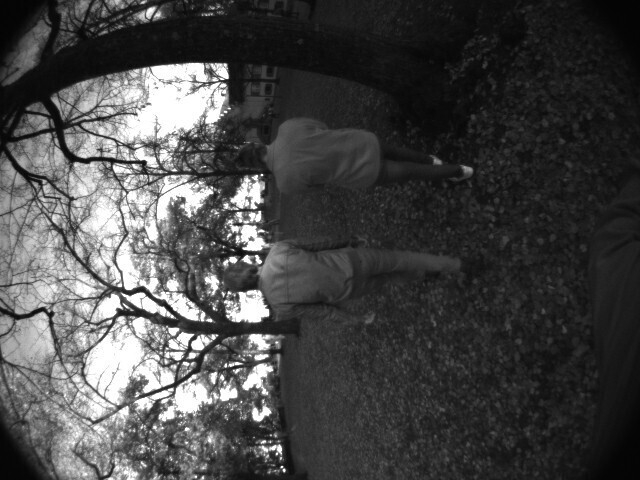}%
\hspace{\pwidth}%
\includegraphics[height=\iwidth,angle=-90]{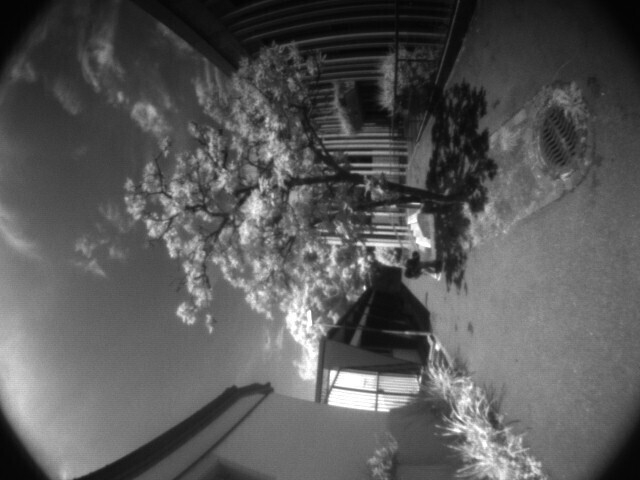}%
\hspace{\bwidth}%
\includegraphics[height=\iwidth,angle=-90]{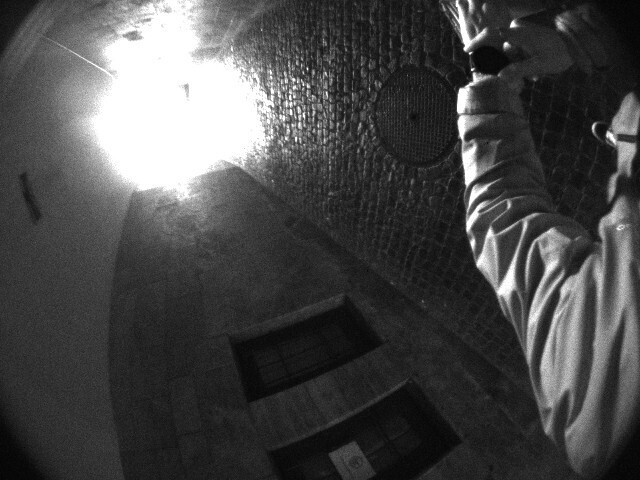}%
\hspace{\pwidth}%
\includegraphics[height=\iwidth,angle=-90]{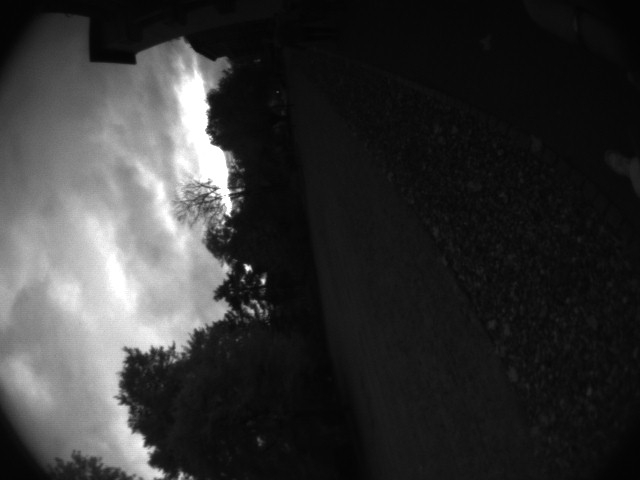}%
\hspace{\bwidth}%
\includegraphics[height=\iwidth,angle=-90]{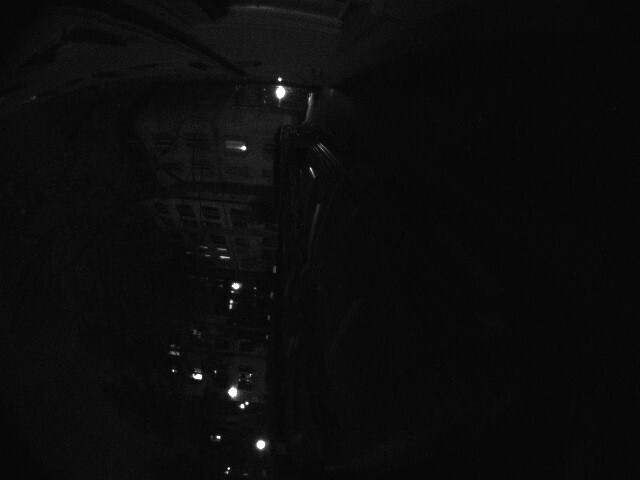}%
\hspace{\pwidth}%
\includegraphics[height=\iwidth,angle=-90]{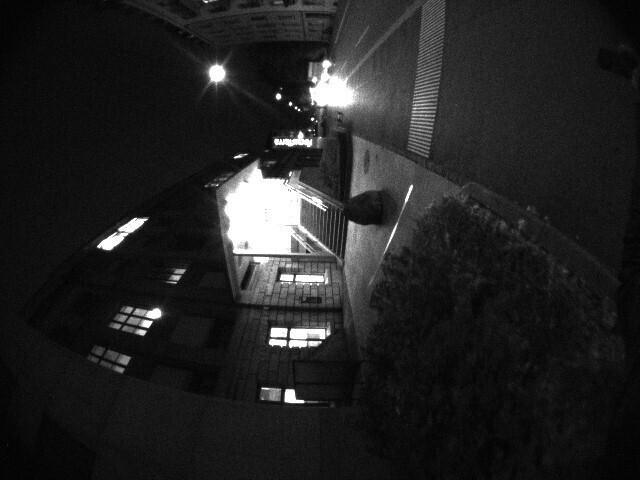}%
\hspace{\bwidth}%
\includegraphics[height=\iwidth,angle=-90]{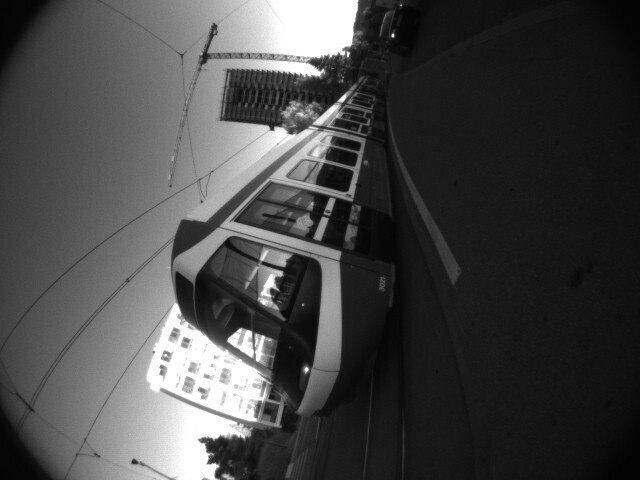}%
\hspace{\pwidth}%
\includegraphics[height=\iwidth,angle=-90]{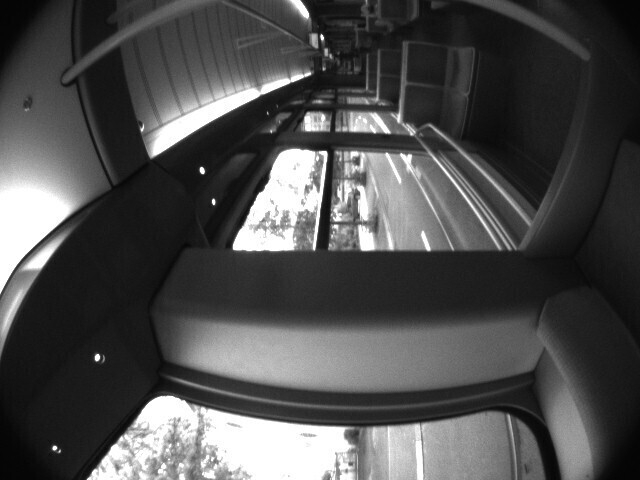}%
\vspace{\pwidth}%

\includegraphics[height=\iwidth,angle=-90]{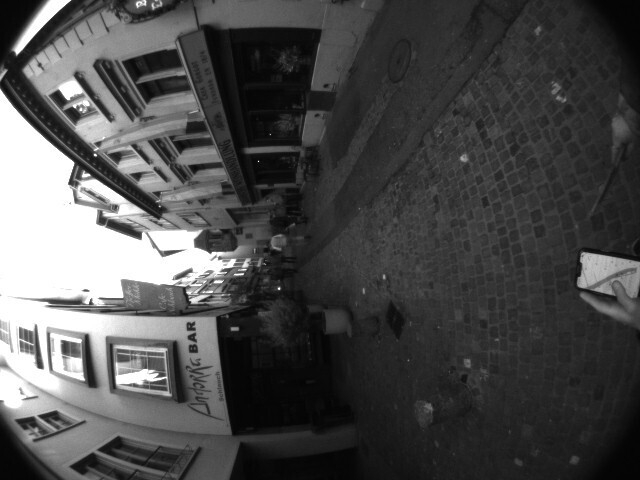}%
\hspace{\pwidth}%
\includegraphics[height=\iwidth,angle=-90]{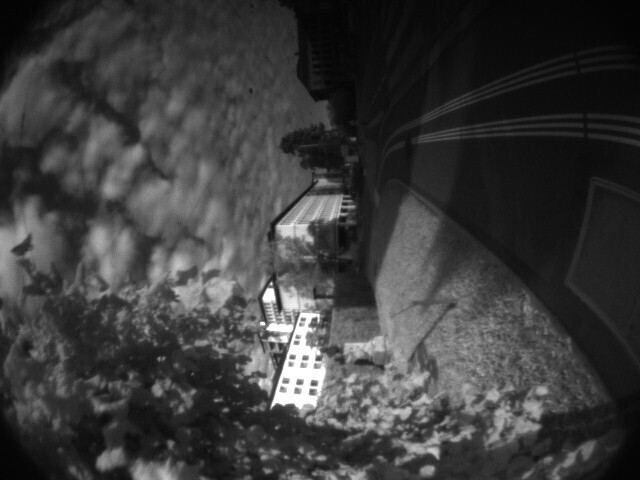}%
\hspace{\bwidth}%
\includegraphics[height=\iwidth,angle=-90]{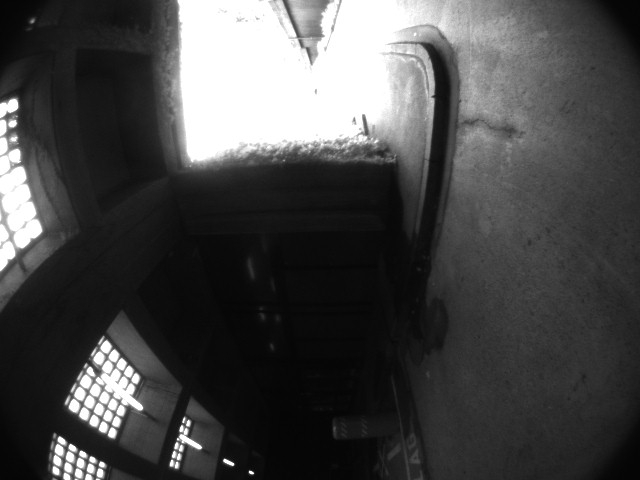}%
\hspace{\pwidth}%
\includegraphics[height=\iwidth,angle=-90]{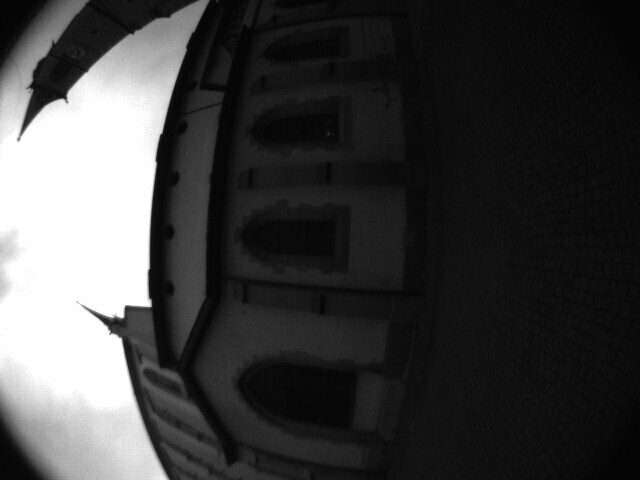}%
\hspace{\bwidth}%
\includegraphics[height=\iwidth,angle=-90]{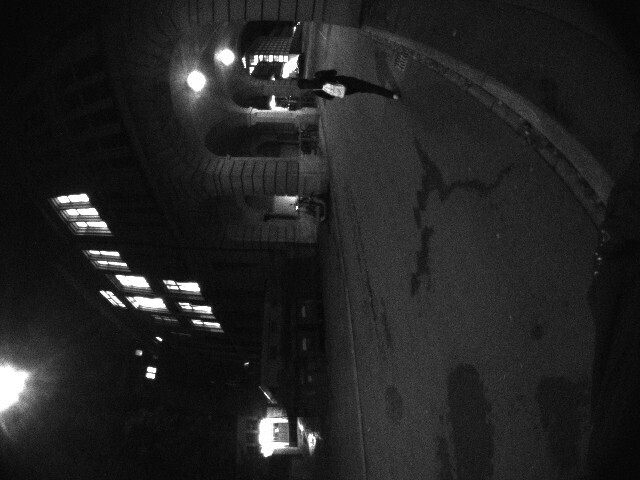}%
\hspace{\pwidth}%
\includegraphics[height=\iwidth,angle=-90]{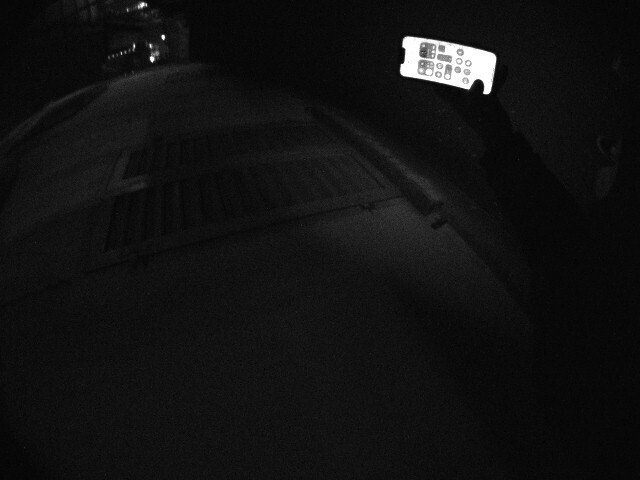}%
\hspace{\bwidth}%
\includegraphics[height=\iwidth,angle=-90]{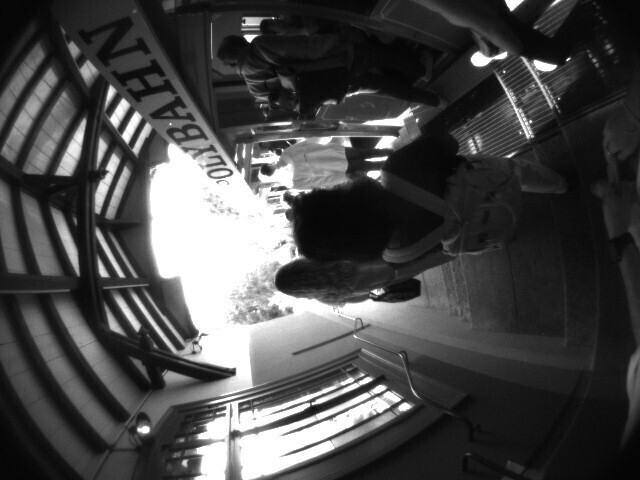}%
\hspace{\pwidth}%
\includegraphics[height=\iwidth,angle=-90]{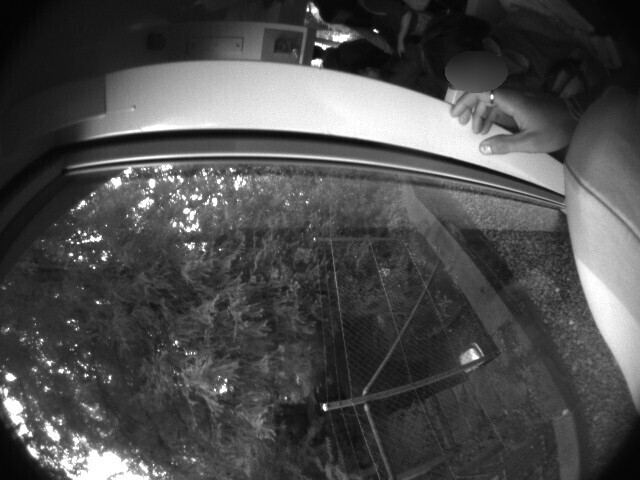}%
\caption{%
\textbf{Challenges:} \ours~includes sensor data recorded by head-mounted devices following outdoor and indoor trajectories in diverse conditions and environments that impair the perceived visual information and are thus challenging for existing algorithms.
}
\label{fig::nominal_images}
\end{figure*}

\paragraph{Odometry and SLAM:}
Visual odometry (VO) \cite{nister2004visual} estimates motion incrementally in a causal manner, while SLAM extends it by incorporating loop closure to correct drift and can be formulated as a batch optimization problem. 
Both are related to Structure-from-Motion (SfM)~\cite{agarwal2010bundle,schoenberger2016sfm,pan2024global,duisterhof2024mast3r}, which more generally handles unordered image collections in an offline manner.
While early visual SLAM methods \cite{davison2007monoslam} employ Extended Kalman Filtering (EKF), PTAM \cite{klein2007parallel} and early ORB-SLAM systems \cite{mur2015orb,mur2017orb} have utilized bundle adjustment with non-linear least square optimization and achieved improved accuracy.
While these approaches rely on image matching via sparse interest points~\cite{lowe2004distinctive,Rublee2011ORB,calonder2010brief}, others
align pixel intensities~\cite{newcombe2011dtam,engel14ECCV} and achieve good stability when a good initialization is available.
These have been later developed into sparse SLAM systems \cite{forster2014svo,engel2017direct}.

With the reduction in size and cost of inertial sensors,
visual-inertial odometry and SLAM have gained significant attention.
Early methods were mainly filter-based, pioneered by MSCKF~\cite{mourikis2007multi} and later extended into full-fledged systems~\cite{bloesch2017iterated,geneva2020openvins,schneider2018maplab}.
Other approaches rely on factor graph optimization \cite{kummerle2011g,ceres-solver,kaess2012isam2,dellaert2012factor} to achieve higher performance with batch VI optimization~\cite{leutenegger2015keyframe,qin2017vins,orbslam3,rosinol2020kimera,leutenegger2022okvis2}.
Direct approaches have also been shown to benefit from inertial constraints~\cite{stumberg18vidso,dmvio}.
As multi-camera system become ubiquitous, many systems also support stereo (horizontally aligned/rectified) \cite{wang2017stereo,rosinol2020kimera} or binocular \cite{geneva2020openvins,leutenegger2022okvis2} modes.

Deep learning too can increase the robustness of VO/SLAM using learned geometric priors \cite{codeslam,czarnowski2020deepfactors,murai2024_mast3rslam,li2024megasam} and differentiable optimization \cite{teed2021droid,teed2023deep,lipson2024deep,chen2024leap}, especially in dynamic environments~\cite{bescos2018dynaslam,zhao2022particlesfm,chen2024leap,zhang2024monst3r}.
It can also benefit systems that rely on inertial constraints~\cite{liu2020tlio,zuo2021codevio,buchanan2022deep}.
While many of these approaches exhibit higher robustness and accuracy on short videos, they are computationally expensive and cannot scale to trajectories that span kilometers.

Our dataset makes it possible to evaluate different sensor configurations that include one or multiple cameras and inertial sensors, on both short and long trajectories.

\paragraph{Datasets and benchmarks:}
There exists many well-established datasets for VIO/SLAM, as shown in \cref{tbl:datasets}.
Most are captured with sensors mounted on cars, robots, or handheld devices, and thus exhibit significantly different motion than devices head-worn in everyday activities.
While mocap systems and indoor surveying instruments are very accurate, they severely limit the scale of the data capture.
Few datasets cover large scale environments, and they generally are automotive and rely on bulky GNSS receivers~\cite{wenzel2020fourseasons,vbr}.
Moreover, egocentric data exhibits a large number of unique challenges, such as time varying calibration, low-light conditions, dynamic environments, moving platforms, \etc.
None of the existing VIO/SLAM datasets reflects the constraints and opportunities of the egocentric setup.
LaMAR~\cite{sarlin2022lamar} offers egocentric data but is mainly designed for benchmarking offline localization and mapping.
Its ground truth poses and sensor calibration are thus not sufficiently accurate to evaluate VIO/SLAM.
On the other hand, existing datasets recorded with Aria devices focus on semantic tasks in small indoor spaces~\cite{lv2024aria,banerjee2024introducing,ma2024nymeria}.
In this paper, we present a city-scale dataset that covers typical challenges found in egocentric data, with centimeter-level pose annotations to reliably evaluate modern VIO/SLAM systems.

\section{Dataset}
\label{sec:dataset}%

We now give an overview of the content of our dataset.

\paragraph{Device and sensors:}
We leverage data collection devices of Project Aria~\cite{aria}, which embed multiple sensors in a glasses-like form-factor. 
These sensors include two synchronized grayscale global-shutter cameras~(640$\times$480, 20\,FPS), a rolling-shutter RGB camera~(1408$\times$1408, 10\,FPS), two inertial measurement units~(IMUs, \qty{1}{\kilo\hertz} and \qty{800}{\hertz}), a magnetometer~(\qty{10}{\hertz}), a barometer~(\qty{50}{\hertz}), a thermometer, a GNSS receiver~(\qty{1}{\hertz}), and WiFi and Bluetooth transceivers~(\qty{0.1}{\hertz}).
They are mounted on a rigid frame, factory-calibrated, and accurately timestamped.
This makes the resulting data ideal for multi-sensor odometry and SLAM.
The recording is controlled by a user-friendly mobile app and is based on the efficient VRS file format~\cite{vrs}, which is optimized for long recordings.

\paragraph{Environment and setup:}
We recorded the dataset in the city center of Zurich, a mid-sized European city, over the course of 6 months.
It spans an area of approximately \qty{1.5}{\square\kilo\meter} and over \qty{50}{\meter} of elevation.
It includes an old town, river banks, several busy tram stations, and a university campus.
Participants were given devices and asked to record multiple sequences through the city.
Each trajectory observes 5 to 30 unique fiducial markers located on control points, whose positions are accurately known.
We rely on them to compute GT device poses, as later explained in~\cref{sec:gt}.
We obtained 63 sequences, each spanning on average \qty{1.5}{\kilo\meter} and \qty{26}{\minute}.
The longest one reaches \qty{2.87}{\kilo\meter} and \qty{48}{\minute}.

\paragraph{Challenges:}
Participants were not familiar with the robustness of existing VIO/SLAM algorithms, ensuring that the data distribution is not biased towards benchmarking.
They most often walked but occasionally also traveled in a tram or a funicular.
This \emph{moving platform} scenario is very challenging as it introduces a discrepancy between visual and inertial constraints but is rarely found in academic benchmarks.
Sequences were recorded at different times of the day but also at night, yielding low-light, uninformative images.
The device is head-worn in most sequences but is occasionally hand-held, with participants removing the device and putting it back on.
This introduces deformations of the frame that may be accounted for with time-varying extrinsic calibration.
The intrinsic calibration can also vary over long sequences as the device temperature increases, which particularly affects the focal length.
Finally, some sequences also traverse indoor buildings and exhibit indoor-outdoor transitions that often introduce over or underexposure (\cref{fig::nominal_images}).

\paragraph{Controlled experimental set:}
Alongside the main dataset, we collect a curated experimental set that serves as an entry-level testbench for VIO/SLAM systems.
This helps understand why egocentric data is much more challenging than existing benchmarks.
This set contains sequences with 4 difficulty levels, as follows:
\begin{itemize}
    \item \textbf{Level I:} platform-based data, with controlled motion and only in-plane rotation.
    \item \textbf{Level II:} platform-based data, with controlled motion and both in-plane and out-of-plane rotation.
    \item \textbf{Level III:} platform-based data, with fast and complex motion / rotation but controlled initial motion.
    \item \textbf{Level IV:} egocentric data, with controlled initial motion.
\end{itemize}
Level II that has four sequences while the other levels each have three sequences. 
The gradually increasing difficulty helps identify system failure cases and bridges the gap between academic datasets and our full egocentric data.

\paragraph{Privacy:}
We blur visible faces and license plates using Aria's EgoBlur~\cite{egoblur}.

\section{Ground-truth generation}
\label{sec:gt}%

We can sparsely evaluate any trajectory based on highly-accurate but sparse ground control points measured independently.
We can further recover a pseudo-ground-truth for the device trajectory through sensor fusion.

\begin{figure}[t]
    \setlength{\pwidth}{0.005\linewidth} 
    \setlength{\iwidth}{0.725\linewidth}
    \setlength{\lwidth}{\dimexpr(0.999\linewidth - \pwidth - \iwidth) \relax}
    \centering
    \begin{subfigure}[t]{\iwidth}
        \centering
        \includegraphics[width=\textwidth]{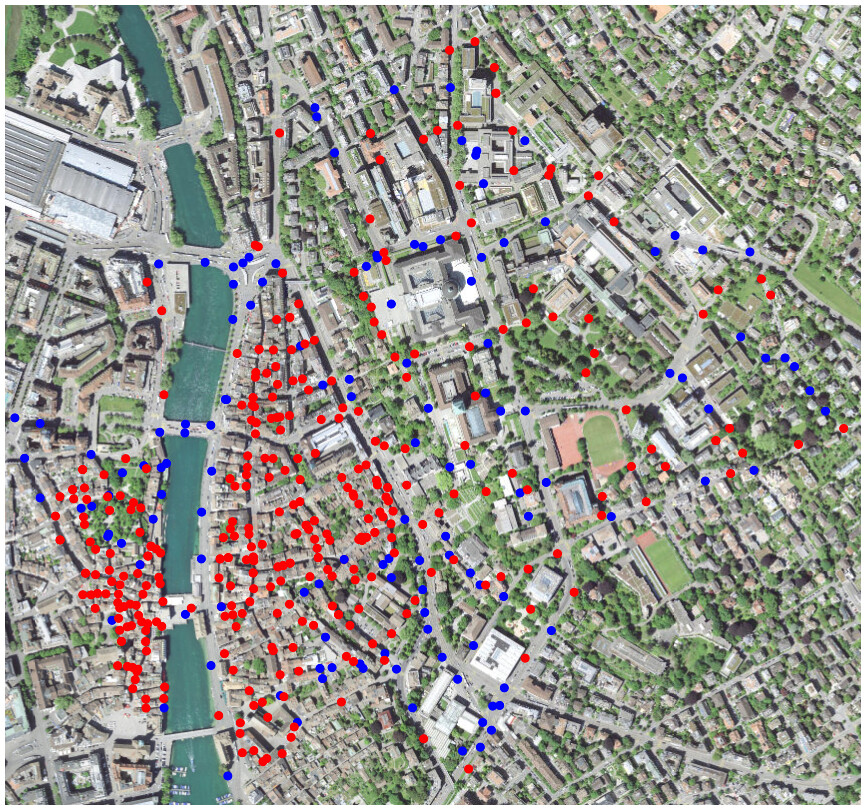}
        \caption{483 \red{2D} and \blue{3D} control points cover the area.}%
        \label{fig:cp:map}%
    \end{subfigure}%
    \hspace{\pwidth}%
    \begin{minipage}[b]{\lwidth}
    \centering
    \begin{subfigure}[t]{\textwidth}
        \centering
        \includegraphics[width=\textwidth]{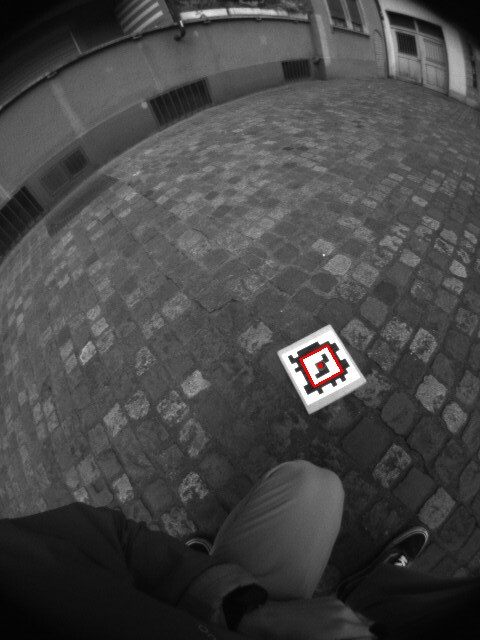}
        \caption{}%
        \label{fig:cp:detect}%
    \end{subfigure}\\
    \begin{subfigure}[t]{\textwidth}
        \centering
        \includegraphics[width=\textwidth]{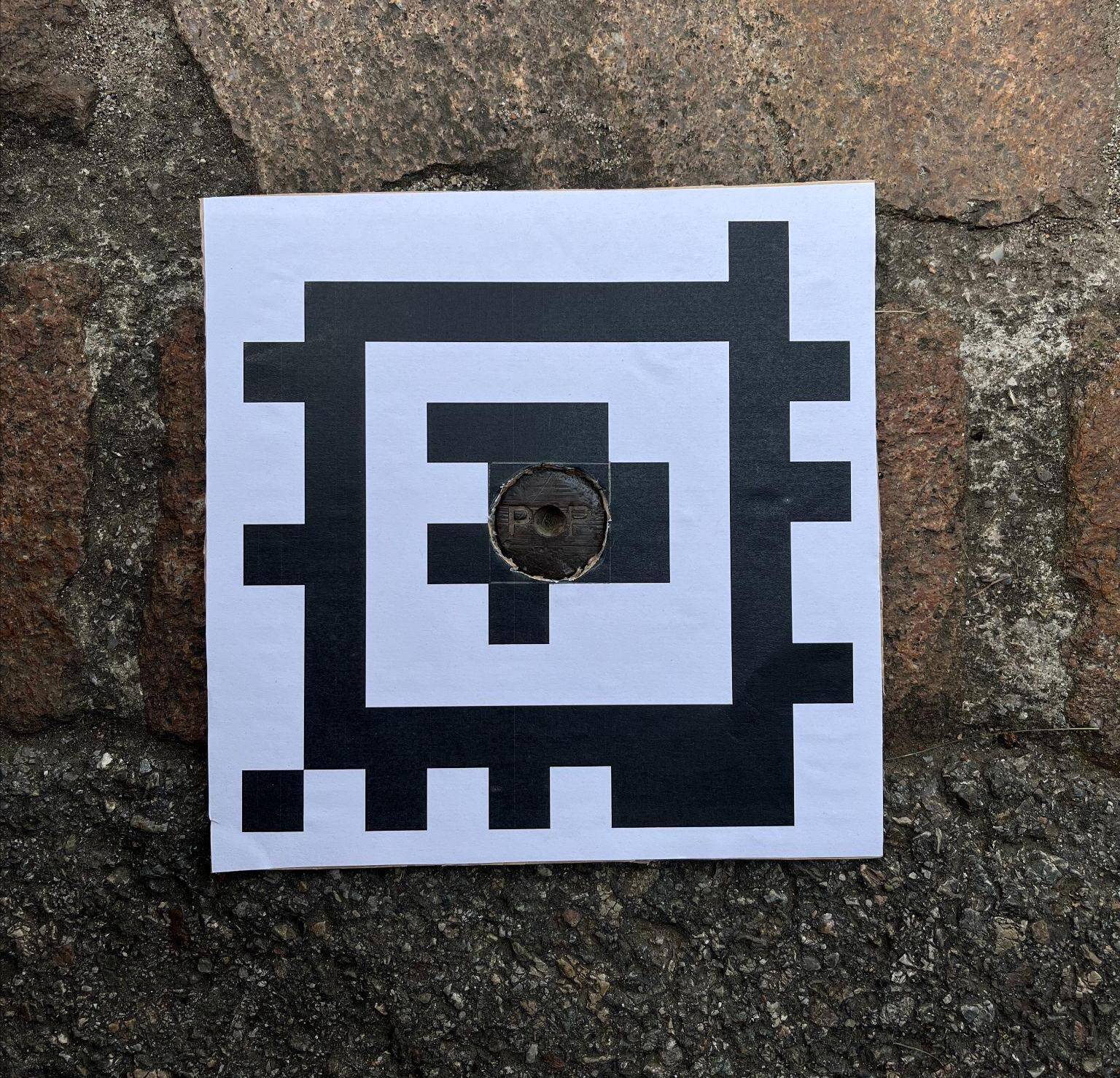}
        \caption{}%
        \label{fig:cp:marker}%
    \end{subfigure}%
    \end{minipage}%
    \caption{\textbf{Control points} (a) are measured with centimeter-accuracy by surveying instruments and are (b) automatically detected in Aria's images using (c) fiducial markers.
    }%
    \label{fig:cp}%
\end{figure}

\subsection{Control points}
\label{sec:gt:cp}

\paragraph{Characteristics:}
Control points (CPs) are 3D points commonly used in the fields of surveying and photogrammetry to anchor sensor measurements in a common reference system across time.
Their position is measured in a geographic coordinate system with a survey instrument like a GNSS-RTK rover, which corrects the error of a GNSS measurement using a nearby base station.
Their uncertainty includes the uncertainty of the base station and increases with the distance between the rover and the station.
As with GNSS, the measurement is reliable only when there is a clear line-of-sight with multiple satellites.
As such, it is not reliable in urban canyons, under overpasses, and indoors.
In these scenarios, surveyors typically propagate constraints from nearby reliable CPs using total stations, which measure absolute distances and relative orientations between CPs.

\paragraph{Public points:}
In many countries, public administrations maintain a registry of CPs in each city.
They are typically used to anchor construction work on public or private ground.
These CPs are generally defined on urban features that are standardized, easy to recognize, and stable over time, such as marked stones or metal bolts sealed in cement.
The position of these CPs is publicly available and regularly updated.
Their measurement process is often documented and their accuracy is guaranteed.
In the area in which our dataset is captured, public CPs are located on average every \qty{35}{\meter}, with an horizontal uncertainty of 1 cm. 

\paragraph{Usage:}
Some public CPs do not provide height information but only a 2D horizontal constrain.
We measured this missing information with a GNSS-RTK rover (Emlid Reach RS3), when reliable.
We added and measured additional 3D CPs in areas where public CPs have a low density.
In total, this results in 483 CPs (134 3D and 349 2D), shown in \cref{fig:cp:map}.
Our measurements have an uncertainty of $\sim$\qty{1.5}{\centi\meter} horizontally and \qty{3}{\centi\meter} vertically.
It is close to identical for all CPs because the distance of the base station is approximately constant~(\qty{4.5}{\kilo\meter}).
We measure each CP three times, on different days, to further validate its uncertainties (see Appendix D for details).

\paragraph{Detection:}
To automatically detect control points in Aria imagery, we attach fiducial markers on top of them.
We choose the AprilTag markers~\cite{apriltag} as they offer a special layout at the center of which there is no information.
We thus pierce a hole in the center of each marker and use it to accurately align the marker with the CP (\cref{fig:cp:marker}).
Each marker is placed on its corresponding CP by a second helper participant shortly before being observed by the Aria device, ensuring that CP and markers are well-aligned.
After constructing the dataset, we blur the detected fiducial markers to avoid biasing the evaluated VIO/SLAM algorithms. 

\begin{figure}[t]
    \centering
    \includegraphics[width=\linewidth]{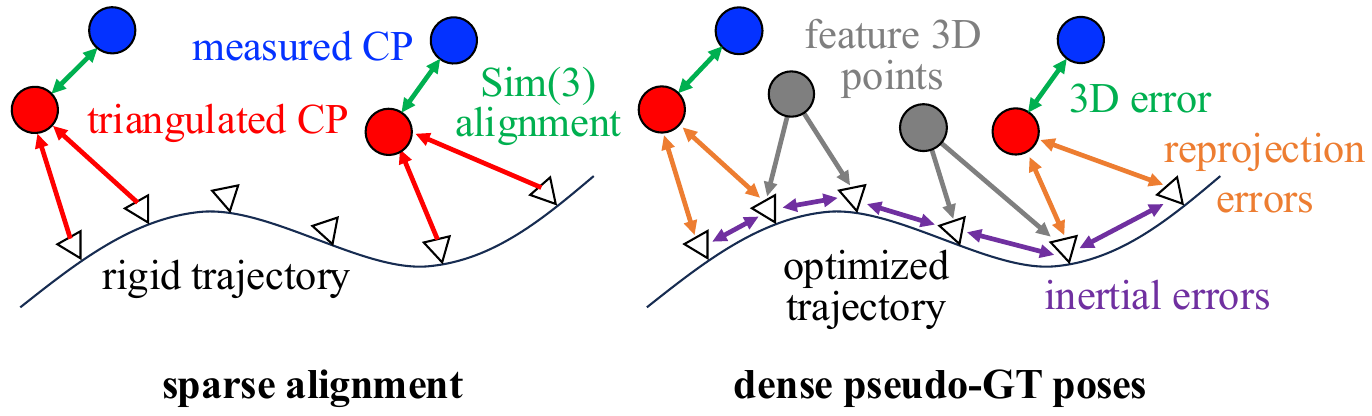}%
    \caption{\textbf{Types of ground-truth.}
    Left: Any trajectory can be evaluated with high accuracy via sparse alignment against the GT control points (CPs).
    Right: We also compute GT camera poses, which are denser but less accurate, via a joint multi-sensor optimization.
    }%
    \label{fig:alignments}%
\end{figure}

\definecolor{errColor}{rgb}{0.851, 0.373, 0.008}
\definecolor{totColor}{rgb}{0.459, 0.439, 0.702}
\definecolor{triColor}{rgb}{0.106, 0.620, 0.467}

\begin{figure}[tb]
\centering
\setlength{\pwidth}{0.495\linewidth}
\includegraphics[width=\pwidth]{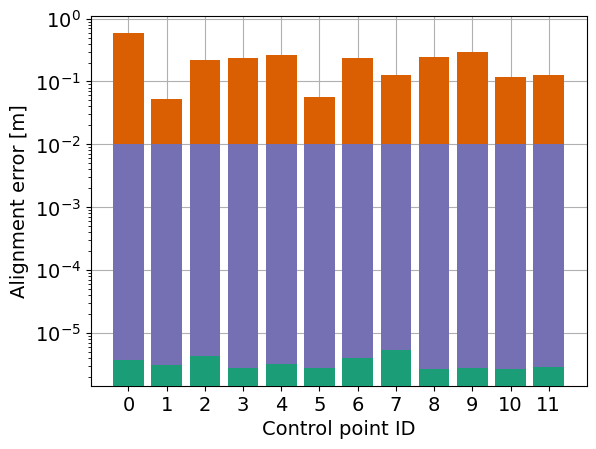}%
\hspace{0.008\linewidth}%
\includegraphics[width=\pwidth]{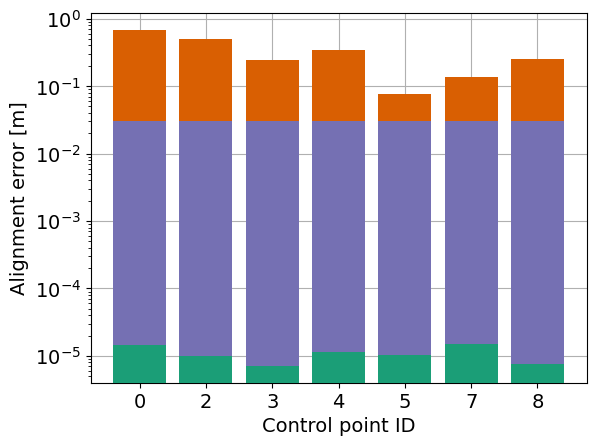}%
\centering
\caption{\textbf{Cross-validation of CPs for sparse alignment.}
The uncertainty of the \textcolor{triColor}{triangulations} and \textcolor{totColor}{CP measurements} are orders of magnitude smaller than the \textcolor{errColor}{CP error} that we evaluate, in both 2D (left) and 3D (right), validating that our sparse GT is sufficiently accurate for evaluation. 
}%
\label{fig::sparse_loocv}
\end{figure}

\subsection{Sparse alignment}
\label{sec:gt:sparse}
We can readily use the detected control points to evaluate any given trajectory and its camera calibration by rigidly aligning them~(\cref{fig:alignments}).
We denote $\mathcal{I}$ the set of image indices and, for an image $i \in \imset$, its pose $\pose{\localf}{i} \in \text{Sim}(3)$, which might include the composition of the device pose and rig extrinsics, in the local SLAM coordinate system $\localf$.

\paragraph{Process:}
For each of the $N$ CPs, we use the 2D detections $\set{\pointim^n_i\in\real^2\ \forall i \in \obs{n} \subset \imset}$ of fiducial markers to triangulate its 3D position $\pointw^n_\localf$ in $\localf$, by minimizing the reprojection error: 
\begin{equation}
    E_\text{tri}^n = \sum_{i \in \obs{n}} \norm{\camproj{\pose{\localf}{i}^{-1}\cdot\pointw^n_\localf, \camera_i} - \pointim^n_i}^2
    \enspace.
    \label{eq:cost:tri}
\end{equation}
Here $\camproj{\cdot}$ is the image projection function, given camera intrinsics $\camera_i$.
We first find an initial estimate using a LO-RANSAC scheme~\cite{fischler1981random,chum2003locally,schoenberger2016sfm} and subsequently refine it with non-linear least-squares.

Next, we aim to estimate the similarity transformation $\pose{\worldf}{\localf} \in \text{Sim}(3)$ that best aligns the triangulated $\set{\pointw^n_\localf}$ and measured $\set{\pointw^n_\worldf}$ CPs. Intuitively, this can be done by minimizing the alignment error:
\begin{equation}
    E_\text{sim} = \sum_{n=1}^N \norm{\pointw^n_\worldf - \pose{\worldf}{\localf} \cdot \pointw^n_\localf}^2\enspace.
    \label{eq:cost:sim}
\end{equation}
However, in practice triangulations may have different uncertainties and can get unreliable when the camera poses exhibit ill-posed configurations. To take this into consideration, we solve for the optimal similarity transformation $\pose{\worldf}{\localf}$ via joint optimization of $\pose{\worldf}{\localf}$ and proxy 3D points $\hat{\pointw}^n_\localf$ over the two factors defined in ~\cref{eq:cost:tri,eq:cost:sim}, weighted by their respective covariances. This formulation is able to account for uncertainties of the camera poses, although this is often not exposed by the algorithms that we evaluate.

\paragraph{Use for evaluation:}
We use the optimal $\pose{\worldf}{\localf} \in \text{Sim}(3)$ from the alignment to transform the original triangulated point and measure its error against the corresponding CP: $\norm{\pointw^n_\worldf - \pose{\worldf}{\localf} \cdot \pointw^n_\localf}$.
These alignment errors define the accuracy of the trajectory: more accurate camera poses better fit the CPs, up to their measurement uncertainty.
Intuitively, for a trajectory of \qty{1.5}{\kilo\meter} and CPs with an horizontal uncertainty of \qty{3}{\centi\meter}, we can measure scale drift of \qty{0.002}{\percent}, which is much lower than the error of state-of-the-art systems -- stereo ORB-SLAM3~\cite{orbslam3} and mono DM-VIO~\cite{dmvio} both report \qty{0.6}{\percent} on the EuRoC dataset~\cite{euroc}.

Our approach is different from other large-scale SLAM benchmarks~\cite{sarlin2022lamar,wenzel2020fourseasons,advio}, which have dense but less accurate GT device poses.
Hilti benchmarks~\cite{hilti2022,hilti2023} also have sparse CPs but those are not detected in the images. 
Instead, the capture device is directly positioned on each CP.
This makes its motion artificial, as the device remains static for a few seconds, and is tedious, as precise alignment takes time.
Our solution has minimal effect on the motion pattern and only requires to observe each CP from different angles to maximize the triangulation accuracy.
This can be easily performed by non-expert participants.

\paragraph{Reliability of the sparse evaluation:}
We aim to study whether the uncertainty in triangulation will affect the CP alignment error in our sparse evaluation.  
We use a trajectory estimated by the best approach (Aria's SLAM, as found later in our benchmark in \cref{sec:evaluation}).
We perform a Leave-one-out Cross Validation (LOOCV) in which we iteratively ignore one CP in the alignment process and compare its alignment error with its aggregated uncertainty:
\begin{equation}
\covar=\covar_{\mathrm{tri_{metric}}}{+}\hat{\covar},   
\end{equation}
where $\covar_{\mathrm{tri_{metric}}}$, $\hat{\covar}$ are the covariances of the transformed triangulation and the corresponding CP respectively. 
The original triangulation covariance $\covar_{\mathrm{tri}}$ is estimated via the inverse of the Gauss-Newton Hessian from the triangulation problem, and can be transformed with scale $s$ and rotation matrix $\mathrm{\*R}$ from $\pose{\worldf}{\localf} \in \text{Sim}(3)$, where $\covar_{\mathrm{tri_{metric}}} = s^2\mathrm{\*R}\covar_{\mathrm{tri}}\mathrm{\*R}^T$.

\cref{fig::sparse_loocv} shows the uncertainty (square root of the covariance spectral norm) of the aligned triangulation and CP measurement on a sequence with 12 CPs. The CP error is on average 70 times larger than its uncertainty, which confirms that our sparse GT is by far sufficiently accurate to benchmark the next generations of SLAM systems.

\subsection{Dense ground-truth poses}
\label{sec:gt:dense}
Having accurate camera poses instead of sparse points is however useful \eg, for fine-grained evaluation and analysis or other computer vision tasks like 3D reconstruction.
As such, we compute pseudo-GT poses and sensor calibrations by fusing visual, inertial, and CP information,
thus propagating the CP constraints to the poses connecting them.

\paragraph{Process:}
We start with an initial SLAM trajectory. While our approach is general, here we rely on the result of the proprietary VI-SLAM exposed by Aria's SLAM API since it performs the best compared to open-source systems (see \cref{sec:evaluation} for more details).
We then align the trajectory sparsely, as described in \cref{sec:gt:sparse},
perform steps typical in Structure-from-Motion: we select representative keyframes, extract local features~\cite{Zhao2023ALIKED}, match them based on sequential pairs and image retrieval~\cite{lindenberger2023lightglue,arandjelovic2016netvlad}, and finally triangulate a sparse 3D point cloud.
Finally, we fix the time-varying intrinsics and refine the camera poses, inertial biases and speeds, triangulated CPs, and 3D points by jointly minimizing 
a) the CP triangulation error (\cref{eq:cost:tri}), 
b) the error between measured and triangulated CPs, 
as well as c) feature reprojection errors and 
d) inertial pre-integration constraints typically used in VI-SLAM~\cite{imupreintegration}.
This is a standard non-linear least-squares optimization problem.

We weight these four terms by their respective covariances.
The covariances of the inertial terms are computed using pre-integration \cite{imupreintegration}, with factory-calibrated IMU noise parameters.
We use two different covariances for the detection of features and fiducial markers, which are iteratively estimated and refined based on the variance factor of the corresponding residuals~\cite{forstnerwrobel}.
Because we know that the CP measurements are unbiased, we deflate their covariance w.r.t.\ the one estimated by the surveying instrument.

For the most challenging sections, \eg, involving moving platforms, neither Aria's SLAM nor any other baseline provides a usable initialization because the visual information is not reliable.
In these cases, we rely only on the inertial and CP information, ignoring visual features.

\begin{figure}[tb]
\centering
\includegraphics[width=0.498\linewidth,trim=0 15pt 20pt 10pt,clip]{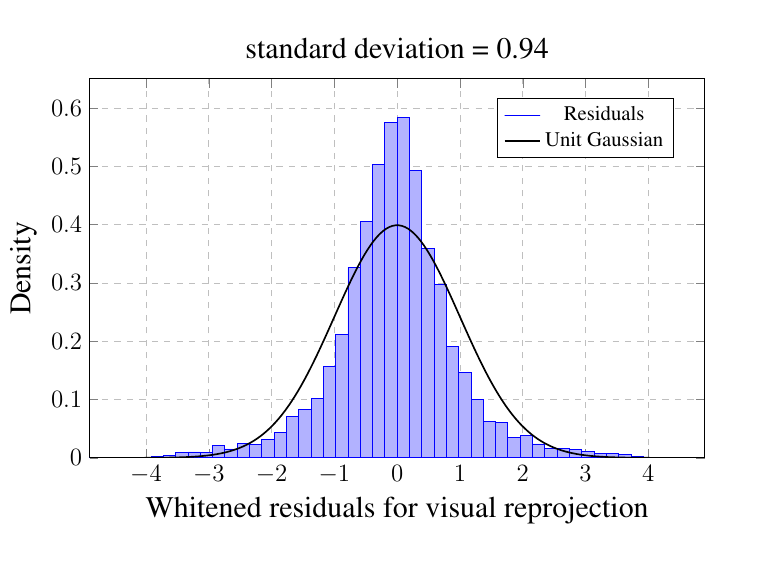}%
\includegraphics[width=0.498\linewidth,trim=0 15pt 20pt 10pt,clip]{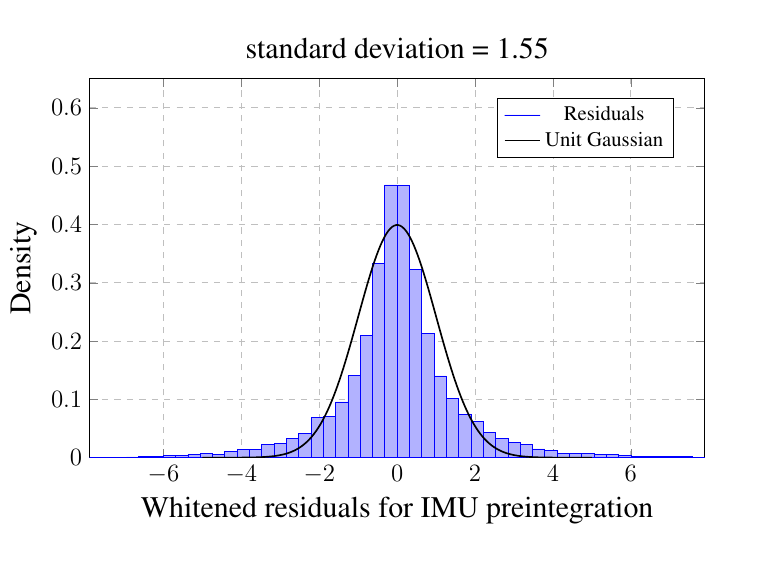}%
\centering
\caption{\textbf{Distribution of the whitened visual and IMU residuals.}}
\label{fig::visual_imu_residuals}
\end{figure}

\paragraph{Accuracy Validation: }
We consider the multi-factor optimization described above, which minimizes different costs $\residual\transp\hat{\covar}^{-1}\residual$, with residuals $\residual\in\real^d$ whitened by their measurement covariance $\hat{\covar}$. We collect all whitened residuals for visual and inertial factors respectively. 
\cref{fig::visual_imu_residuals} shows that the distribution of residuals is close to a unit Gaussian distribution for both types, which validates the appropriate weighting of the visual and inertial terms in the optimization \cite{forstnerwrobel}. 

We further validate this dense optimization by dropping the two CP-related factors and evaluating the trajectory sparsely (\cref{sec:gt:sparse}).
We find that our pseudo-ground-truth has an overall accuracy of $\sim$\qty{20}{\centi\meter} (see Appendix D for details), which is sufficiently accurate to measure keyframe errors larger than \qty{50}{\centi\meter} in trajectories spanning kilometers.

\section{Evaluation}
\label{sec:evaluation}

\begin{table*}[tb]
\centering
\scriptsize
\setlength\tabcolsep{5pt} %
\begin{tabular}{l l ccc cccc ccc ccc}
\toprule
\multirow{2}{*}{method} & \multirow{2}{*}{year} 
& \multicolumn{3}{c}{level I} 
& \multicolumn{4}{c}{level II} 
& \multicolumn{3}{c}{level III} 
& \multicolumn{3}{c}{level IV}\\
\cmidrule(lr){3-5} \cmidrule(lr){6-9} \cmidrule(lr){10-12} \cmidrule(lr){13-15} &
& seq 1 & seq 2 & seq 3
& seq 4 & seq 5 & seq 6 & seq 7
& seq 8 & seq 9 & seq 10
& seq 11 & seq 12 & seq 13 \\
\midrule
\mono{DSO}                                  & 2016 & 0.76 & 0.21 & \fail & 1.95 & 19.80 & \fail & 30.12 & \fail & \fail & \fail & \fail & \fail & \fail \\
\mono{ORB-SLAM3}                            & 2020 & 0.13 & 0.35 & 0.12 & 0.72 & 1.25 & \fail & 5.94 & \fail & \fail & \fail & \fail & \fail & \fail \\
\mono{DPVO}                                 & 2023 & 0.22 & 0.16 & 0.28 & 1.10 & 5.40 & 1.74 & 5.15 & 7.14 & 18.40 & 15.65 & 42.07 & 28.73 & 39.43 \\
\mono{DPV-SLAM}                             & 2023 & 0.12 & 0.20 & 0.29 & 5.35 & 2.93 & 2.95 & 8.10 & 2.83 & 19.45 & 19.48 & 28.02 & 29.43 & 50.41 \\
\midrule
\monoinertial{Kimera VIO}                   & 2020 & 0.82 & 2.38 & 0.74 & 7.23 & 4.06 & 11.06 & 12.59 & 9.70 & 32.24 & \fail & 7.98 & 13.11 & \fail \\
\monoinertial{ORB-SLAM3}                    & 2020 & 0.03 & 0.43 & 0.23 & 2.01 & 1.51 & 2.70 & 5.90 & 7.80 & 19.10 & 21.80 & \fail & 14.60 & \fail \\
\monoinertial{OpenVINS}                     & 2020 & 0.75 & 3.15 & 0.96 & 2.32 & 3.94 & 6.07 & 5.96 & 9.23 & 23.78 & 8.62 & 5.87 & 28.00 & 17.57 \\
\monoinertial{OpenVINS + Maplab}            & 2022 & 0.71 & 3.09 & 0.77 & 1.23 & 3.92 & 6.00 & 4.86 & 8.87 & 23.90 & 7.73 & 5.92 & 26.30 & 15.44 \\
\monoinertial{DM-VIO}                       & 2022 & 0.75 & 0.34 & 0.26 & 0.78 & 32.03 & \fail & \fail & \fail & \fail & \fail & 10.70 & \fail & \fail \\
\midrule
\binoinertial{OpenVINS}                     & 2020 & 0.66 & 2.36 & 0.68 & 0.94 & 1.43 & 1.35 & 2.96 & 4.25 & 4.31 & 8.01 & 1.04 & 18.72 & 10.35 \\
\binoinertial{OpenVINS + Maplab}            & 2022 & 0.65 & 2.30 & 0.68 & 1.05 & 1.22 & 1.19 & 2.01 & 3.97 & 4.29 & 8.22 & 1.62 & 16.59 & 8.37 \\
\binoinertial{OKVIS2}                       & 2024 & 0.02 & 0.72 & 0.03 & 1.36 & 0.80 & 3.78 & \fail & 6.81 & 5.32 & 7.06 & 1.85 & 16.55 & 6.65 \\

\bottomrule
\end{tabular}
\caption{\textbf{Results for the controlled experimental set.} 
We evaluate systems on \mono{monocular}, \monoinertial{monocular+inertial}, and \binoinertial{multi-camera+inertial} inputs.
We report the ATE RMSE \cite{grupp2017evo} (lower is better) in meters for each sequence.
Failures to output a valid trajectory are marked as \fail.
}
\label{tbl:results:experimental_set}
\end{table*}

\begin{figure}[tb]
\begin{minipage}{0.48\linewidth}
\vspace{10pt}\includegraphics[width=1.0\linewidth]{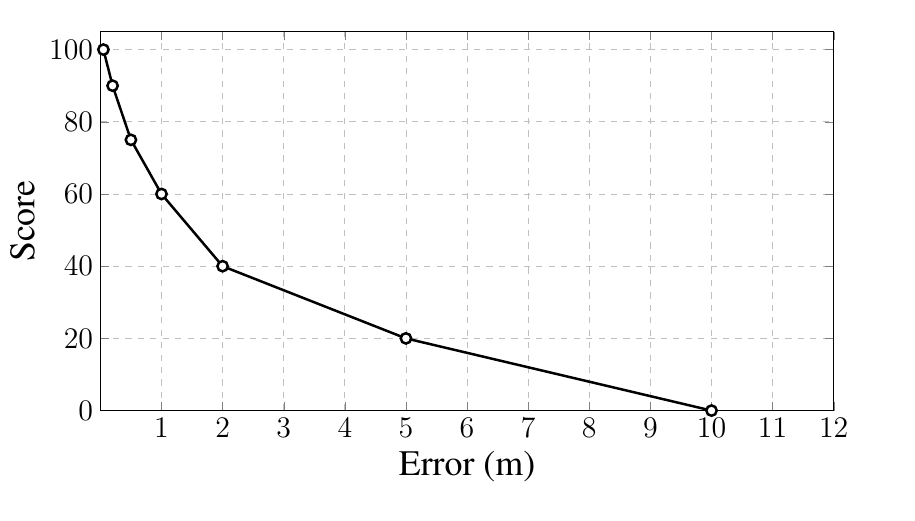}
\end{minipage}
\begin{minipage}{.48\linewidth}
\centering
\scriptsize
\setlength{\tabcolsep}{2pt}
\begin{tabular}{lcccc}
    \toprule
error $e$ & $\qty{5}{\centi\meter}$ & $\qty{20}{\centi\meter}$ & $\qty{50}{\centi\meter}$ & $\qty{1}{\meter}$ \\
\midrule
\midrule
score $s(e)$ & 100 & 90 & 75 & 60 \\
\midrule
error $e$ & $\qty{2}{\meter}$ & $\qty{5}{\meter}$ & $\qty{10}{\meter}$ & $>$$\qty{10}{\meter}$ \\
\midrule
\midrule
score $s(e)$ & 40 & 20 & 0 & 0 \\
\bottomrule
\end{tabular}
\end{minipage}
\caption{\textbf{Scoring function $s(e)$ for the sparse evaluation} given the measured alignment error.
The function is piecewise linear as shown in the left plot, with its anchor points in the right table.
}%
\label{fig::scoring_sparse}
\end{figure}

\begin{table*}[tb]
\centering
\scriptsize
\setlength\tabcolsep{1pt}%
\begin{tabular}{l c ccc c ccc c ccc c ccc c ccc}
\toprule
\multirow{2}{*}{method} & \multirow{2}{*}{causal} & \multicolumn{3}{c}{short} & & \multicolumn{3}{c}{medium} & &\multicolumn{3}{c}{long} & & \multicolumn{3}{c}{challenge -- low-light} & & \multicolumn{3}{c}{challenge -- moving platform} \\
\cmidrule(lr){3-5} \cmidrule(lr){7-9} \cmidrule(lr){11-13} \cmidrule(lr){15-17} \cmidrule(lr){19-21}
& & score\,$\uparrow$ & CP@1m\,$\uparrow$ & R@5m\,$\uparrow$
& & score\,$\uparrow$ & CP@1m\,$\uparrow$ & R@5m\,$\uparrow$
& & score\,$\uparrow$ & CP@1m\,$\uparrow$ & R@5m\,$\uparrow$
& & score\,$\uparrow$ & CP@1m\,$\uparrow$ & R@5m\,$\uparrow$
& & score\,$\uparrow$ & CP@1m\,$\uparrow$ & R@5m\,$\uparrow$\\
\midrule
\mono{DPVO}                         & \checkmark & 9.4 & 1.7 & 21.3 & & 5.2 & 1.0 & 10.8 & & 1.2 & 0.0 & 1.9 & & 3.4 & 0.2 & 7.5 & & 2.4 & 0.1 & \na \\
\mono{DPV-SLAM}                     & x & 7.5 & 1.5 & 14.8 & & 5.2 & 1.4 & 10.1 & & 0.4 & 0.0 & 0.7 & & 1.9 & 0.4 & 3.5 & & 1.7 & 0.0 & \na   \\
\midrule
\monoinertial{Kimera VIO}           & \checkmark & 6.3 & 2.9 & 12.6 & & 6.6 & 1.7 & 15.1 & & 6.3 & 1.7 & 14.3 & & 4.2 & 2.7 & 6.4 & & 7.1 & 1.6 & \na   \\
\monoinertial{ORB-SLAM3}            & x & 28.3 & 13.4 & 67.1 & & 20.3 & 4.4 & 57.0 & & 14.2 & 2.3 & 40.6 & & 6.2 & 0.6 & 12.5 & & 15.7 & 4.1 & \na  \\
\monoinertial{OpenVINS}             & \checkmark & 18.1 & 4.4 & 45.7 & & 10.9 & 2.3 & 27.9 & & 4.7 & 0.5 & 12.3 & & 7.9 & 2.4 & 17.6 & & 2.4 & 0.6 & \na \\
\monoinertial{OpenVINS + Maplab}    & x  & 22.9 & 8.1 & 50.8 & & 13.1 & 4.1 & 29.0 & & 5.8 & 1.3 & 13.3 & & 9.6 & 2.9 & 19.3 & & 3.7 & 1.2 & \na \\
\midrule
\binoinertial{OpenVINS}             & \checkmark & 22.2 & 6.2 & 57.9 & & 17.8 & 5.7 & 46.1 & & 10.6 & 1.7 & 25.8 & & 16.9 & 6.2 & 38.2 & & 11.5 & 2.4 & \na \\
\binoinertial{OpenVINS + Maplab}    & x & 26.0 & 9.5 & 61.1 & & 21.3 & 7.3 & 50.6 & & 12.6 & 1.9 & 30.3 & & 16.5 & 4.6 & 37.9 & & 13.0 & 3.0 & \na \\
\binoinertial{OKVIS2}               & x & 24.2 & 12.0 & 54.7 & & 13.6 & 6.8 & 28.2 & & 3.6 & 2.7 & 7.2 & & 15.4 & 5.4 & 38.6 & & 4.2 & 2.8 & \na  \\

\midrule
\midrule
\binoinertial{Aria's SLAM}           & x & 90.7 & 99.2 & \na & & 78.5 & 87.4 & \na & & 70.8 & 75.9 & \na & & 84.2 & 91.6 & \na & & 53.6 & 51.2 & \na  \\
\bottomrule
\end{tabular}
\caption{\textbf{Main evaluation.}
We evaluate systems on \mono{monocular}, \monoinertial{monocular+inertial}, and \binoinertial{multi-camera+inertial} inputs.
We also include the closed-source SLAM system exposed by Aria's SLAM API.
We report the average score (defined in Fig. \ref{fig::scoring_sparse}), recall w.r.t.\ control points at 1 meter (CP@1m), and recall w.r.t.\ pseudo-GT at 5 meters (R@5m), all of which are evaluated in 2D.
}
\label{tbl:results:long}%
\end{table*}

\begin{figure*}[tb]
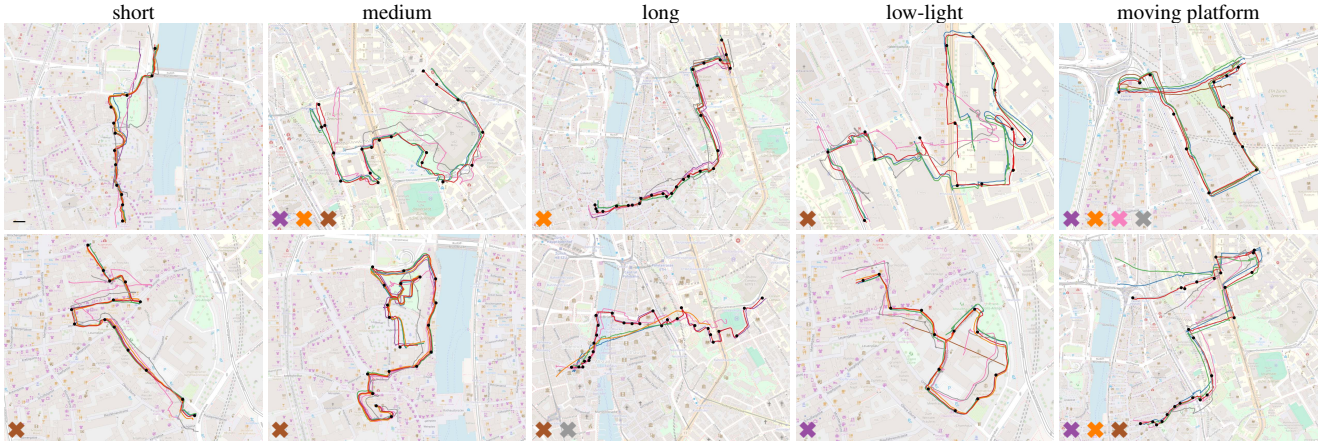

\centering
\definecolor{mps}{RGB}{228,26,28}
\definecolor{openvins}{RGB}{55,126,184}
\definecolor{maplab}{RGB}{77,175,74}
\definecolor{orbslam3}{RGB}{152,78,163}
\definecolor{okvis2}{RGB}{255,127,0}
\definecolor{kimera}{RGB}{166,86,40}
\definecolor{dpvo}{RGB}{247,129,191}
\definecolor{dpvslam}{RGB}{153,153,153}
\newcommand{\thickcross}{\ding{54}}
\newcommand{\soliddot}{{\large $\bullet$}}%
\newcommand{\imageoverlay}[2]{%
\begin{overpic}[width=\iwidth]{figures/visualization/trajs/#1}
\put (1,1) {\small #2}
\end{overpic}%
}
\def\ncols{5}
\setlength{\pwidth}{0.005\linewidth}
\setlength{\iwidth}{\dimexpr(0.999\linewidth - \ncols\pwidth + \pwidth)/\ncols \relax}
\begin{minipage}[b]{\iwidth}
\centering{\footnotesize short}
\end{minipage}%
\hspace{\pwidth}%
\begin{minipage}[b]{\iwidth}
\centering{\footnotesize medium}
\end{minipage}%
\hspace{\pwidth}%
\begin{minipage}[b]{\iwidth}
\centering{\footnotesize long}
\end{minipage}%
\hspace{\pwidth}%
\begin{minipage}[b]{\iwidth}
\centering{\footnotesize low-light}
\end{minipage}%
\hspace{\pwidth}%
\begin{minipage}[b]{\iwidth}
\centering{\footnotesize moving platform}
\end{minipage}%

\imageoverlay{DMA_LIN_1.pdf}{\ --}%
\hspace{\pwidth}%
\imageoverlay{DMA_ETH_2.pdf}{%
\textcolor{orbslam3}{\thickcross}
\textcolor{okvis2}{\thickcross}
\textcolor{kimera}{\thickcross}
}%
\hspace{\pwidth}%
\imageoverlay{lindenhof_1.pdf}{\textcolor{okvis2}{\thickcross}}%
\hspace{\pwidth}%
\imageoverlay{DMA_ETH_5.pdf}{\textcolor{kimera}{\thickcross}}%
\hspace{\pwidth}%
\imageoverlay{DMA_ETH_4_2.pdf}{%
\textcolor{orbslam3}{\thickcross}
\textcolor{okvis2}{\thickcross}
\textcolor{dpvo}{\thickcross}
\textcolor{dpvslam}{\thickcross}
}%
\vspace{0.004\linewidth}%

\imageoverlay{RW_1.pdf}{\textcolor{kimera}{\thickcross}}%
\hspace{\pwidth}%
\imageoverlay{DMA_LIN_8.pdf}{\textcolor{kimera}{\thickcross}}%
\hspace{\pwidth}%
\imageoverlay{lindenhof_21.pdf}{%
\textcolor{kimera}{\thickcross}
\textcolor{dpvslam}{\thickcross}
}%
\hspace{\pwidth}%
\imageoverlay{RW_10.pdf}{\textcolor{orbslam3}{\thickcross}}%
\hspace{\pwidth}%
\imageoverlay{lindenhof_6.pdf}{%
\textcolor{orbslam3}{\thickcross}
\textcolor{okvis2}{\thickcross}
\textcolor{kimera}{\thickcross}
}%
\centering
\caption{%
\textbf{Visualizations of the trajectories estimated by different systems:}
\textcolor{openvins}{OpenVINS \soliddot},
\textcolor{maplab}{OpenVINS+Maplab \soliddot},
\textcolor{orbslam3}{ORB-SLAM3 \soliddot},
\textcolor{okvis2}{OKVIS2 \soliddot},
\textcolor{kimera}{Kimera VIO \soliddot},
\textcolor{dpvo}{DPVO \soliddot},
\textcolor{dpvslam}{DPV-SLAM \soliddot},
and \textcolor{mps}{Aria's SLAM \soliddot}.
Failures are shown as \thickcross\ and control points in black \soliddot.
}
\label{fig::visualization}
\end{figure*}

We can use the sparse alignment defined above to evaluate any estimated trajectory on our full set.
In the following part, we present our benchmark results and analysis for various publicly-available visual (-inertial) odometry/SLAM systems under different sensor configurations. 

\paragraph{Systems:}
We evaluate systems under monocular, monocular-inertial and multi-camera-inertial setups.
The latter relies on the two global-shutter SLAM cameras, which cannot be used in a proper stereo (rectified) mode because they have little overlap in field of view.
We try with our best efforts to cover most major popular systems, including
those based on optimization (Kimera VIO~\cite{rosinol2020kimera}, ORB-SLAM3~\cite{orbslam3}, OKVIS2~\cite{leutenegger2022okvis2}),
filtering (OpenVINS~\cite{geneva2020openvins}),
direct alignment (DSO~\cite{engel2017direct}, DM-VIO~\cite{dmvio}),
and deep learning (DPVO~\cite{teed2023deep}, DPV-SLAM~\cite{lipson2024deep}).
In addition, we evaluate the offline VI optimization in Maplab~\cite{schneider2018maplab} on top of the odometry output of OpenVINS~\cite{geneva2020openvins}.
We make sure to appropriately tune on our data the hyperparameters of each approach, often with the help of the respective authors.

\paragraph{Calibration:}
Unless stated otherwise, we use the factory calibration provided by the device.
The systems are thus responsible for refining this calibration and modeling temporal variations.
However, all systems that we consider, with the exception of OpenVINS~\cite{geneva2020openvins}, do not support online intrinsic calibration and implement only few simple camera models.
When necessary, we therefore undistort the images to pinhole cameras, using the factory calibration, and run the algorithms on the resulting images.

\subsection{Controlled experimental set}
We first evaluate the systems on our controlled experimental set
to better understand their failure patterns.

\paragraph{Setup:}
We rely on our dense pseudo-GT poses for sequences in level IV but, for sequences in level I-III, since they are short and do not include CPs, we use the output of Aria's SLAM API as pseudo-GT.
Following common practice in existing benchmarks \cite{euroc,tumvi}, we align the output trajectories with the GT poses and calculate the average of the Absolute Trajectory Error (ATE)~\cite{grupp2017evo} for each sequence.
We report a failure if the system crashes or only produces results that span less than half of the total sequence duration. 

\paragraph{Results:}
\cref{tbl:results:experimental_set} shows that all approaches work reasonably well on level I.
This is consistent with the findings of commonly-used academic datasets like EuRoC~\cite{euroc}, whose data exhibits similar controlled motions.
However, as the motion becomes more natural/egocentric, most systems start to break down by either reporting failures or suffering from drifts and large trajectory error.
In particular, the two direct methods and ORB-SLAM3 mono easily lose track due to fast and complex motion patterns.
This motivates the need for a VIO/SLAM benchmark that captures such motions in unrestricted egocentric recordings with a larger span in length and duration.

\subsection{Main benchmark}
\label{subsec:main_benchmark}
\paragraph{Setup:}
We categorize 63 sequences in our main dataset into short (up to 15 CPs, 18 sequences), medium (16-22 CPs, 10 sequences), long (more than 22 CPs, 16 sequences) and highlight specific hard challenges with low-light (9 sequences) and moving platforms (10 sequences).
We provide more statistics in the supplementary material.
Inspired by the Hilti challenge \cite{hilti2022,hilti2023}, we design a scoring function for each of the CP error (\cref{fig::scoring_sparse}).
We employ a piecewise linear scoring function, where we can confidently evaluate until \qty{5}{\centi\meter} with our survey-grade control points.

Along with the score averaged over the sequences, we also report the recall of the CP alignment error at \qty{1}{\meter} and the recall of the device position error w.r.t.\ our dense pseudo-GT poses at \qty{5}{\meter}.
Unlike the standard ATE, these three metrics can be computed even when systems estimate no or only partial trajectories.
We do not report the pose recall for Aria's SLAM, as it is used to initialize the optimization of pseudo-GT poses, and for sequences involving moving platforms, as we have limited guarantees on the accuracy of their pseudo-GT poses (see Appendix E for details).

We evaluate all methods that produce valid results on level IV of the experimental set, according to \cref{tbl:results:experimental_set}. 
Note that the sequences in the short-sequence group are overall more difficult than the level IV recordings due to the presence of non-controlled initial motion, which brings additional practical challenges to bootstrapping and IMU initialization.
We perform all evaluation in 2D to make use of all the control points.
To account for randomness, we average the metrics over 3 runs of each system.

\paragraph{Results:}
We show the main results in \cref{tbl:results:long}.
As expected, relying on multiple cameras and inertial sensors significantly benefits each VIO/SLAM system.
Overall, ORB-SLAM3 scores the best in the mono-inertial category, while OpenVINS achieves the most promising results when considering multi cameras.
The non-linear optimization of Maplab also consistently boosts the accuracy of the estimates.
However, all academic solutions suffer largely on the defined challenges and are far behind Aria's commercial SLAM system, which is non-causal and includes online calibration and a full visual-inertial bundle adjustment.
On the other hand, the evaluation of Aria's SLAM indicates that our benchmark is not saturated even for a heavily engineered system, especially in sequences that include moving platforms.
This further highlights that our dataset provides a good benchmark for all practices relevant to developing VIO/SLAM solutions for unconstrained egocentric data.

Qualitative results in \cref{fig::visualization} show that most evaluated systems face drift and failures in long recordings.
They are also more prone to failure when facing low-light conditions and moving platforms, while Aria's SLAM produces the most reasonable trajectories that best fit the control points.

\subsection{Discussions}

Our dataset opens up opportunities for more principled iterations of multi-sensor SLAM developments on top of uncontrolled egocentric recordings. 
Following our empirical studies on all the evaluated systems, 
we highlight several promising research directions to explore for addressing the unique characteristics of egocentric data: 

\begin{itemize}
    \item Online optimization of time-varying calibration, which distinguishes Aria's SLAM from existing academic baselines, adapts to the always-on nature of the wearable devices.
    \item Loop closure detection and VI bundle adjustment, to reduce odometry drift in open loop predictions.
    \item Robust outlier removal, tailored strategies for moving platforms, and better handling of tracking loss.
    \item Advanced image matching and point tracking based on machine learning models trained on large datasets.
\end{itemize}

To specifically quantify the importance of online optimization, we analyze the online calibration results from Aria's SLAM. We observe that the variation range of focal length is comparably larger in the long sequences (0.11\%) than in medium (0.09\%) and short (0.08\%) ones.
Fixing the calibration to factory calibration in our VI optimization also empirically leads to a large decrease in pose accuracy, when validated with our survey-grade control points.

\section{Conclusion}
We have introduced a new dataset and benchmark for visual-inertial odometry and SLAM for egocentric, multi-modal data captured by future head-mounted computing devices.
The dataset covers multiple challenges that no existing dataset covers: extreme low-light, crowded or moving environments like vehicles, and kilometer-long trajectories with time-varying calibration.
The dataset provides sparse, centimeter-accurate pose annotations from surveying for tens of kilometers of trajectories spanning a large city center.
This annotation is orders of magnitude more accurate than the best existing visual-inertial odometry/SLAM systems and will enable the development and benchmarking of the next generations of multi-sensor algorithms.
Our results show that much progress remains to be made, which will be supported by our public evaluation.

\section*{Acknowledgements}
We sincerely thank the authors of ORB-SLAM3, OKVIS2, Kimera-VIO, and GLOMAP for their invaluable guidance on baseline implementations. We also thank Thomas Posur and Konrad Schindler for generously providing surveying equipment and assistance. Our appreciation goes to Johannes Schönberger for his insightful feedback on COLMAP. We are also grateful to Boyang Sun and Tian Yi Lim for their insightful discussions on visual-inertial SLAM, as well as their assistance with our maplab setup. Finally, we thank Ying Zhou and Maitraya Avadhut Desai for their support with data recording and capture.

\appendix
\section*{Appendix}
\section{Dataset Statistics}
In this section, we aim to provide more detailed statistics on the proposed dataset.

\begin{figure}[tb]
  \centering
  \begin{minipage}{0.49\columnwidth}
    \centering
    \includegraphics[angle=-90,width=\linewidth]{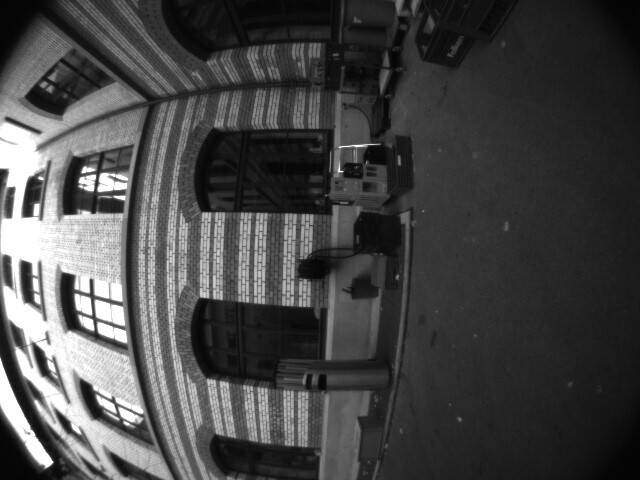}
  \end{minipage}
  \hfill
  \begin{minipage}{0.49\columnwidth}
    \centering
    \includegraphics[angle=-90,width=\linewidth]{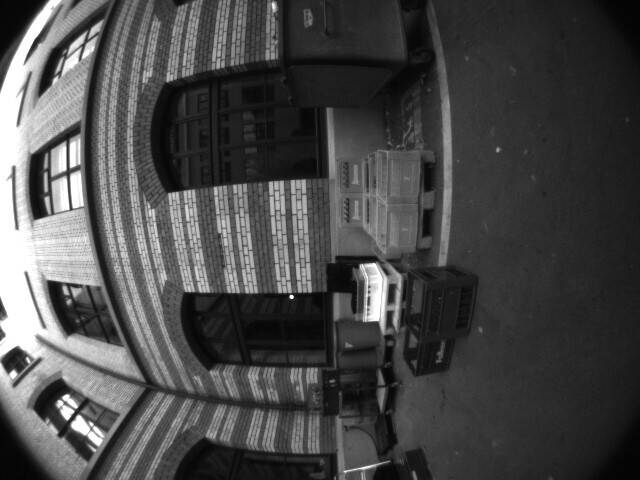}
  \end{minipage}
  \vspace{1em}

  \begin{minipage}{0.49\columnwidth}
    \centering
    \includegraphics[width=\linewidth]{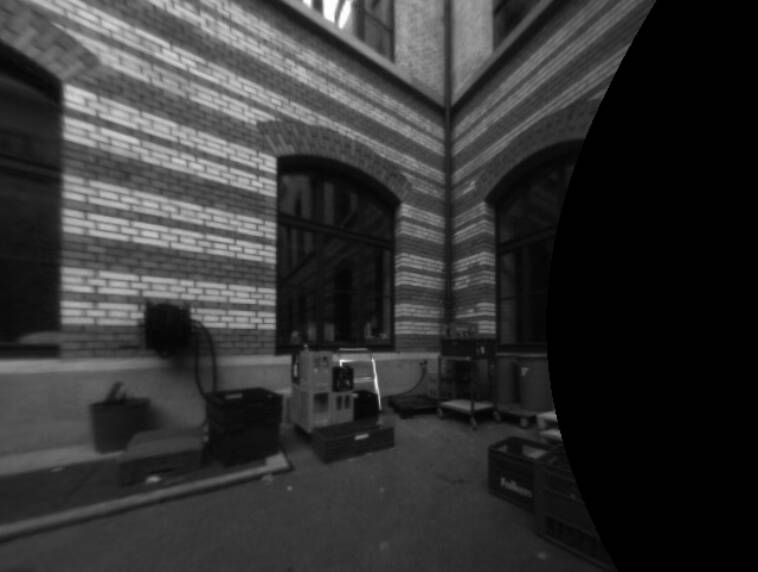}
  \end{minipage}
  \hfill
  \begin{minipage}{0.49\columnwidth}
    \centering
    \includegraphics[width=\linewidth]{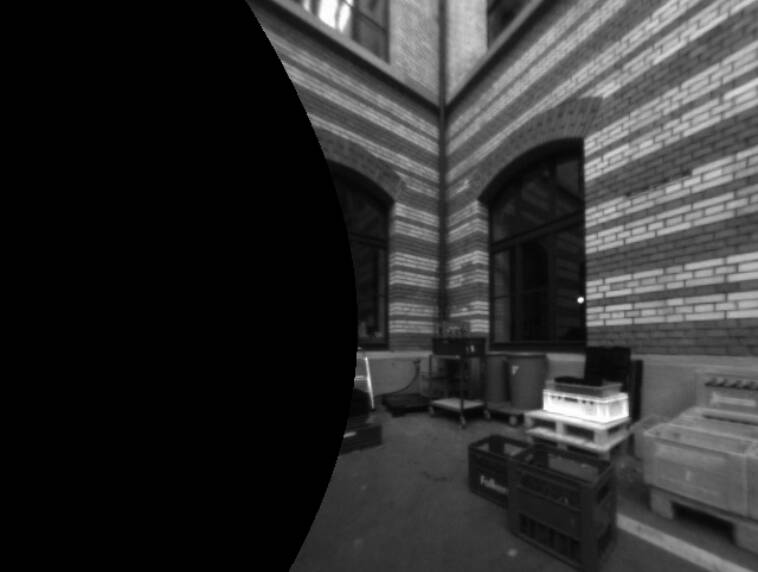}
  \end{minipage}

  \caption{\textbf{The overlap of the two SLAM cameras is severely limited, making it hard to evaluate in horizontal stereo mode.} \textbf{Top:} The original SLAM image pair, \textbf{Bottom:} The SLAM image pair after stereo rectification.}
  \label{fig:stereo_rectification}
\end{figure}

\paragraph{Sensor Configurations.}
Our sensor configurations are presented in details in Section 3. We employ a custom profile from Project Aria \cite{aria} that best fits our demand on the dataset construction. The RGB sensor (10 FPS) is rolling shutter, so we did not use it in the evaluation of this paper, but it is provided in the dataset as one of the modalities. Therefore, we use the two available SLAM cameras or the multi-camera inertial setup in our evaluation. The two SLAM cameras are on the sides of the glasses and thus do not have enough overlap to support horizontal stereo setup (as shown in \cref{fig:stereo_rectification}). 

\begin{figure}[tb]
\centering
\setlength{\pwidth}{0.32\linewidth}
\includegraphics[width=\pwidth]{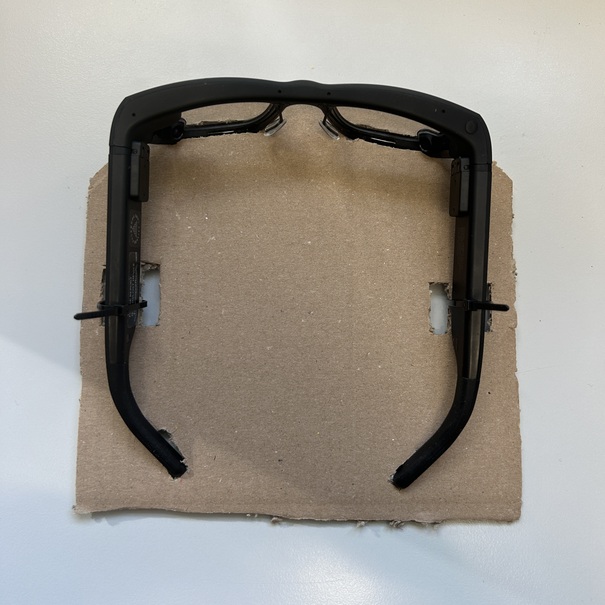}%
\hspace{0.008\linewidth}%
\includegraphics[width=\pwidth]{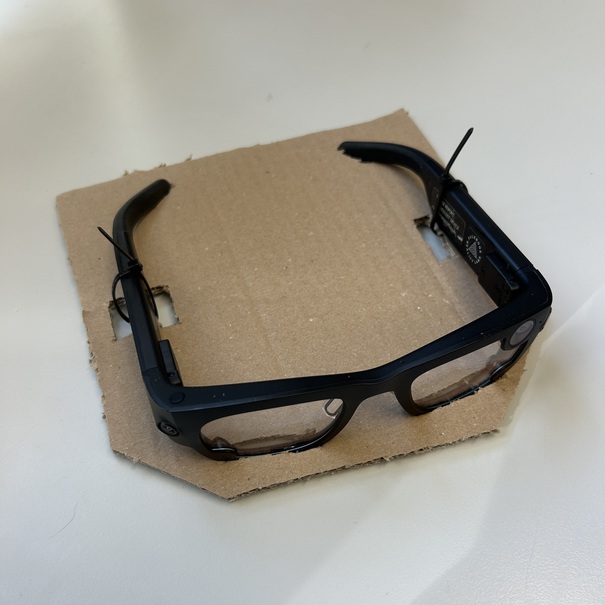}%
\hspace{0.008\linewidth}%
\includegraphics[width=\pwidth]{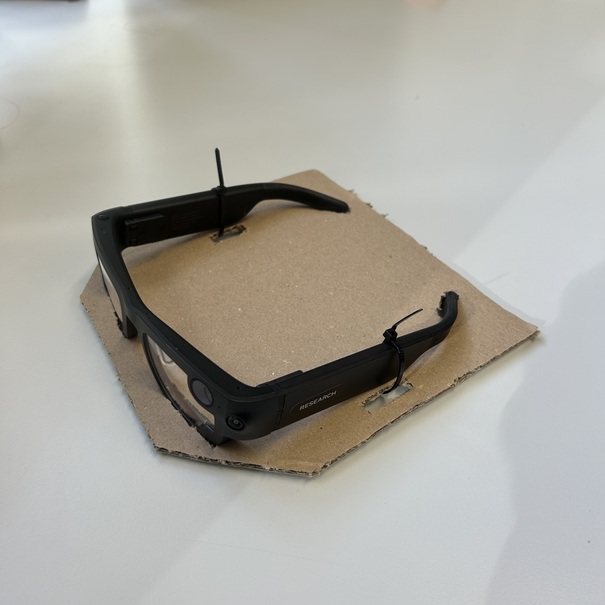}%
\centering
\caption{\textbf{The setup of the self-built platform} for the capture of level I, II, III recordings in the controlled experimental set.
}%
\label{fig::supp_platform_setup}
\end{figure}

\paragraph{Controlled Experimental Set.}
The controlled experimental set consists of sequences with four different levels of difficulties. The first three levels (I, II, III) are platform-based with handheld artificial motion, to mimic the setup of the commonly used academic datasets \cite{euroc}. Specifically, we put the Aria glasses on a self-assembled carton platform (as in Fig. \ref{fig::supp_platform_setup}). The recordings in level IV are egocentric in nature, with controlled initial motion to help mitigate issues in IMU initialization. The motion patterns gradually become more complex for the four levels, as discussed in Sec. 3 in the main paper.

\cref{fig::supp_examples_experimental_set} further shows visualizations of different motion patterns for the four levels respectively. Level II includes out-of-plane rotation while level III has fast and complex movements with significant vertical motion. In level IV, the data is recorded with head-worn glasses and exhibits natural head motion that is common in egocentric data.

\begin{figure}[tb]
\centering
\setlength{\pwidth}{0.495\linewidth}
\includegraphics[width=0.45\linewidth]{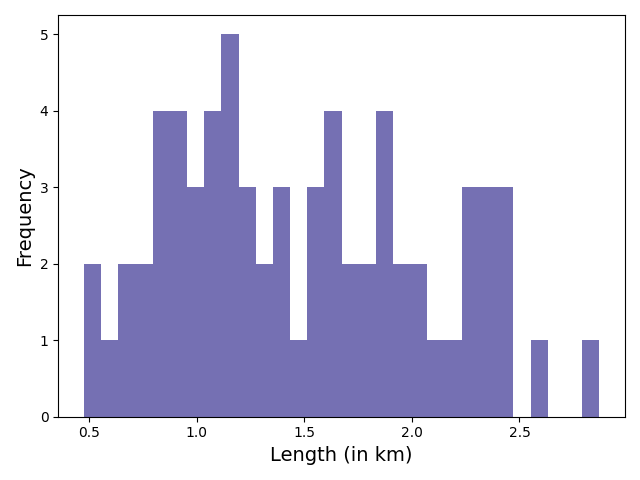}%
\hspace{0.008\linewidth}%
\includegraphics[width=0.45\linewidth]{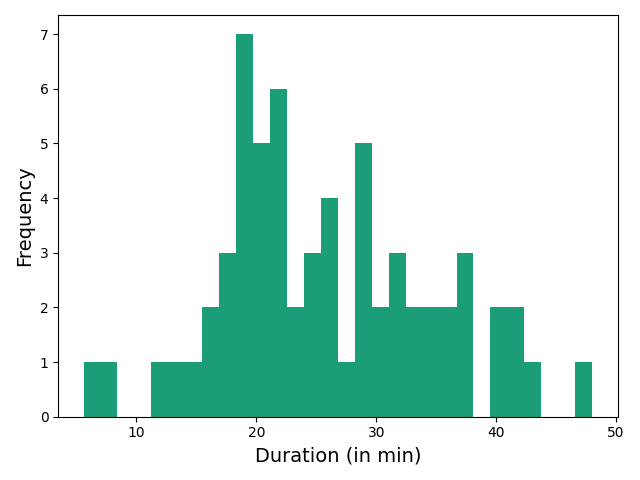}%
\centering
\caption{Histograms on the distribution of recording duration and length in the main dataset.
}
\label{fig::supp_stats_histogram}
\end{figure}

\paragraph{Main Dataset.}
The main dataset is categorized into five groups in the evaluation. We first group all recordings that cover low-light conditions or moving platforms into two specific challenge groups, and then categorize the rest of the recordings by number of covered CPs. The detailed statistics for each recording, along with the covered challenges, are listed in \cref{tbl:supp_recording_stats_part1} and \cref{tbl:supp_recording_stats_part2}. While most sequences are egocentric, we also have a few sequences where we hold the Aria glasses in hand. Depending on the trajectories, our sequences can be categorized into three types: random walking (rw), A-to-B sequence (a\_b), which connect two distant areas, and sequences that include densely mapping an area (dma). The last type of sequence enables applicability of our dataset on benchmarking multi-sequence algorithms. The sequence duration varies from 5min to 48min, covering trajectories that span kilometers. We also provide indicators on the presence of specific challenges in each sequence. \cref{fig::supp_stats_histogram} shows the histogram of duration and length for all the 63 recordings in our dataset.

\section{More Details on Score Evaluation w.r.t. Control Points}

\begin{figure*}[tb]
\centering
\scriptsize
\setlength\tabcolsep{0.7pt}
\begin{tabular}{cccccc}
\includegraphics[height=0.13\linewidth]{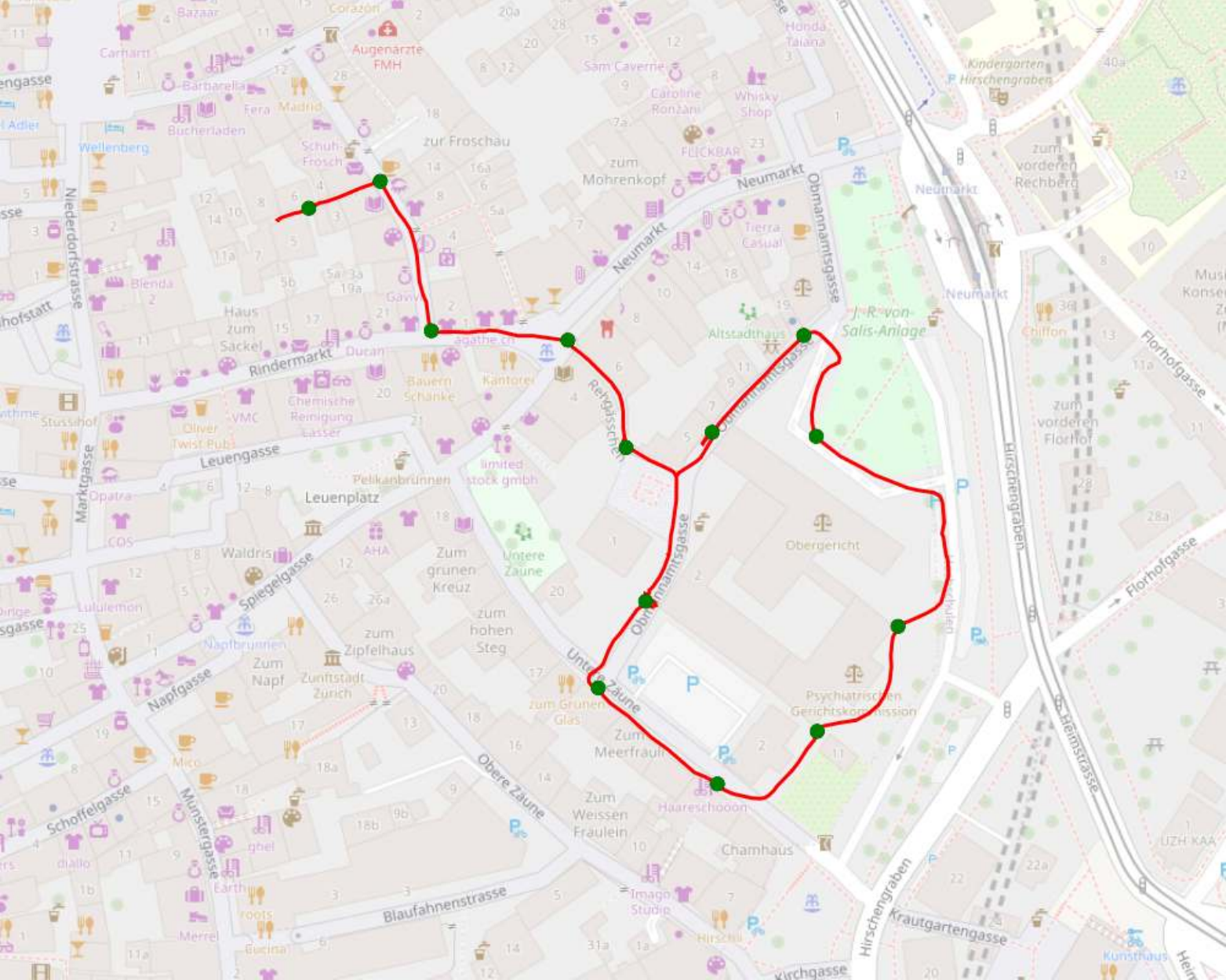} & 
\includegraphics[height=0.13\linewidth]{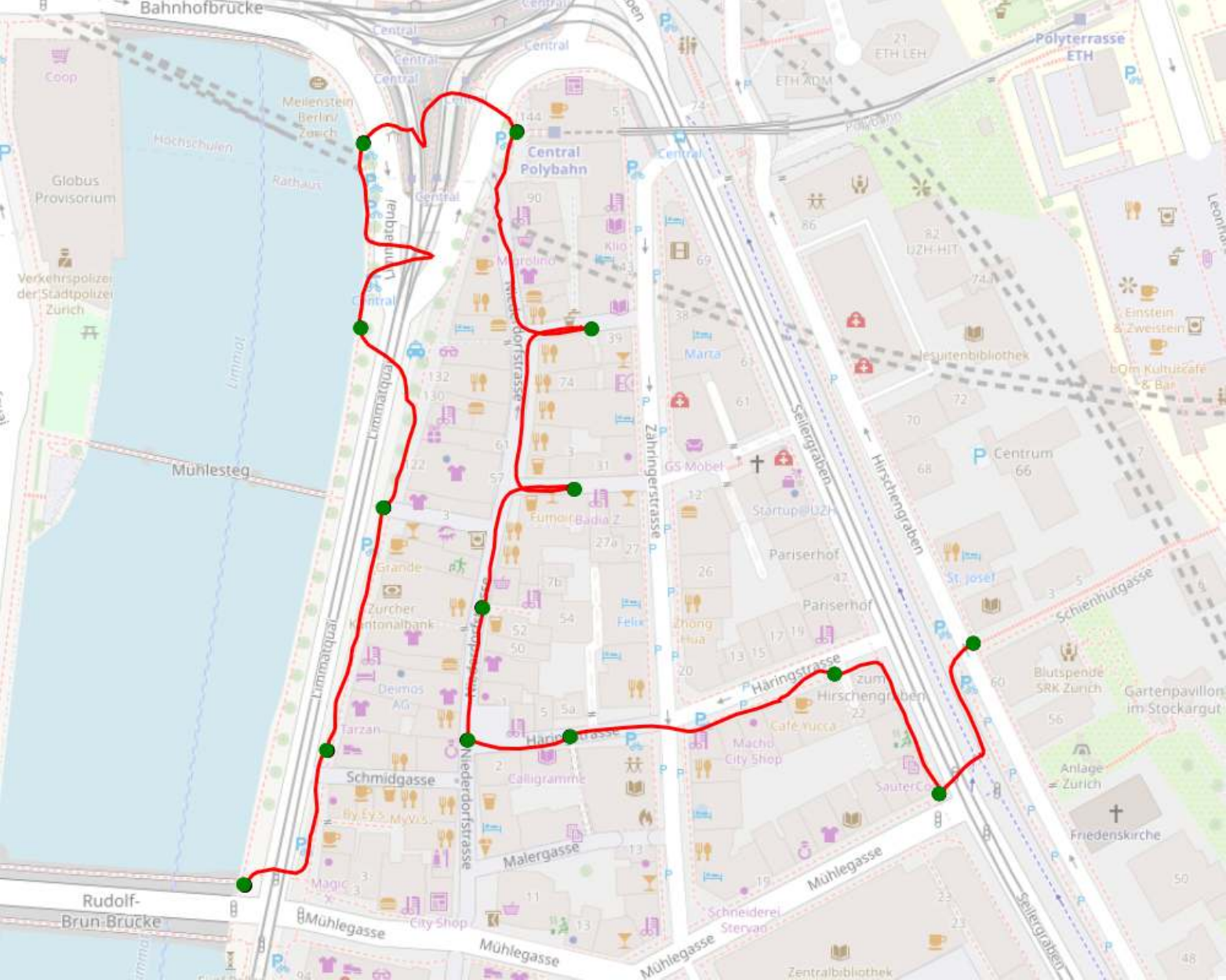} & 
\includegraphics[height=0.13\linewidth]{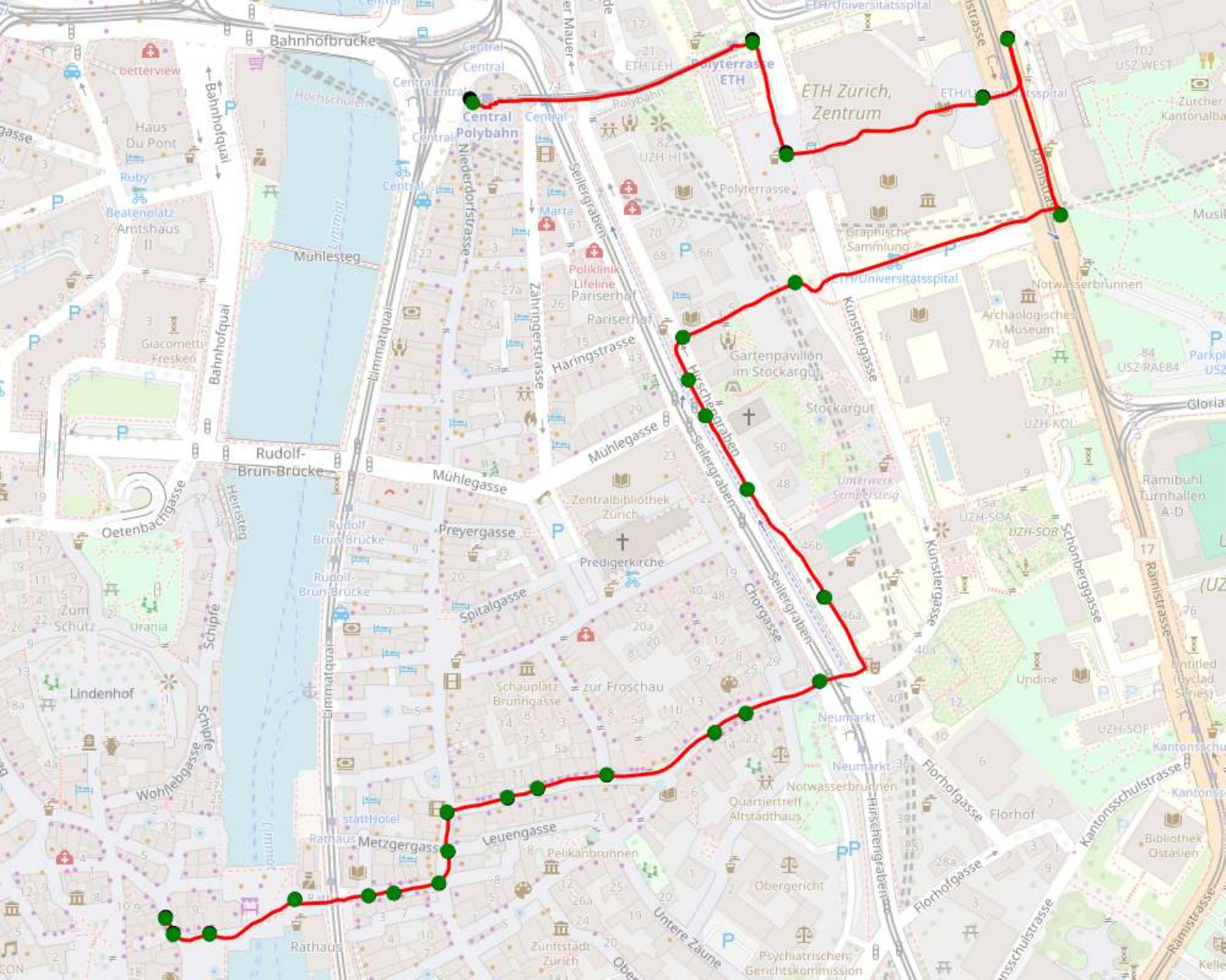} & 
\includegraphics[height=0.13\linewidth]{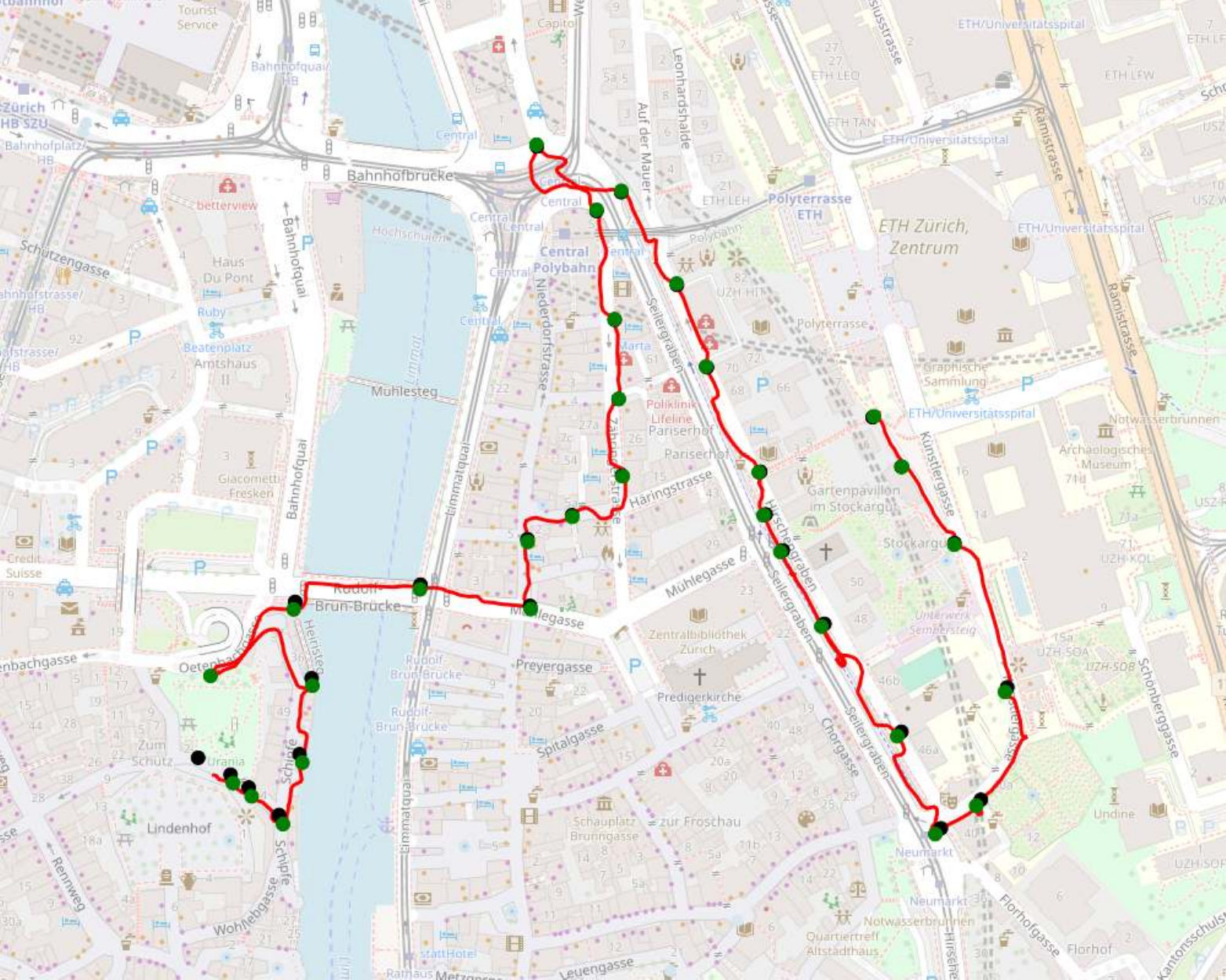} & 
\includegraphics[height=0.13\linewidth]{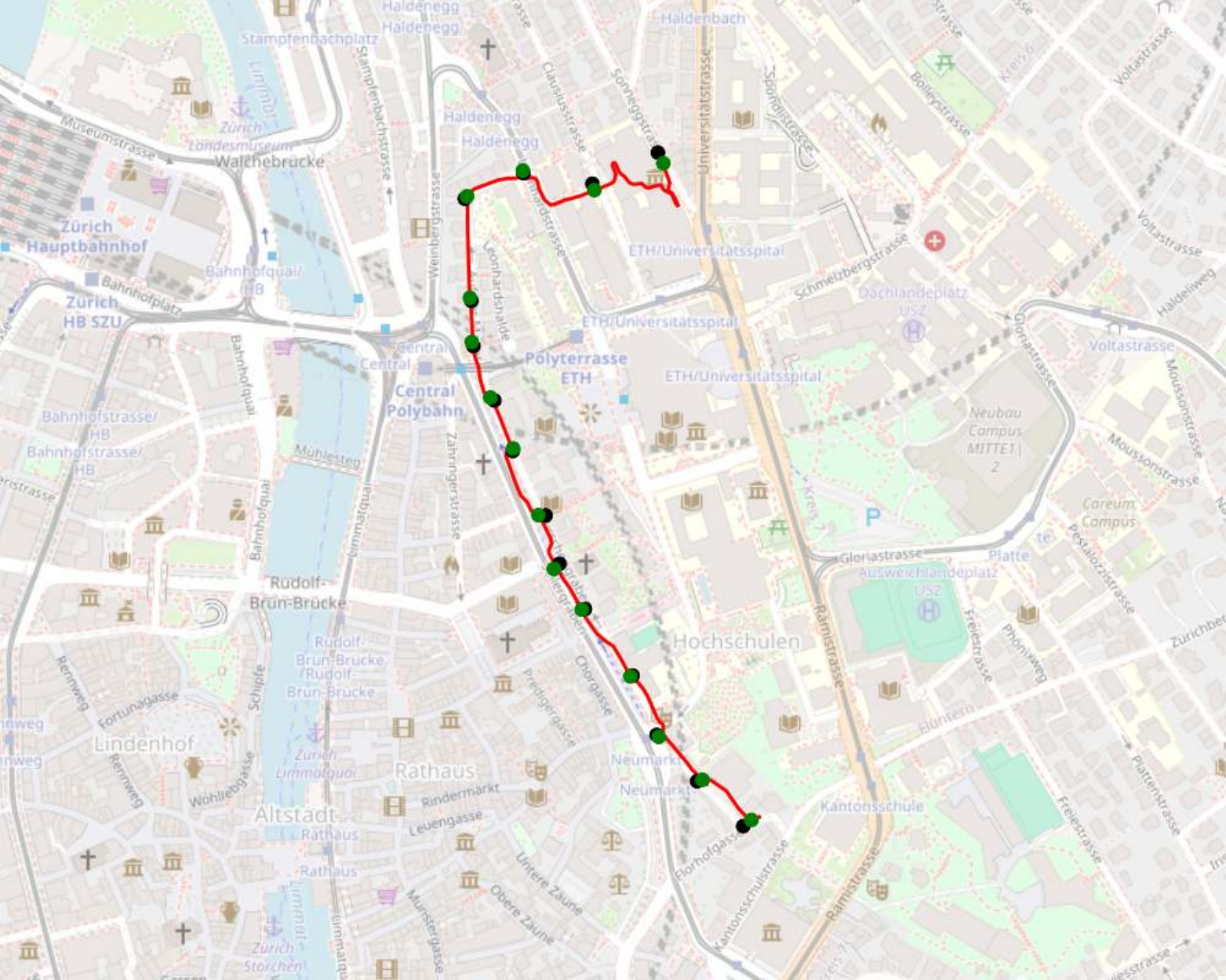} & 
\includegraphics[height=0.13\linewidth]{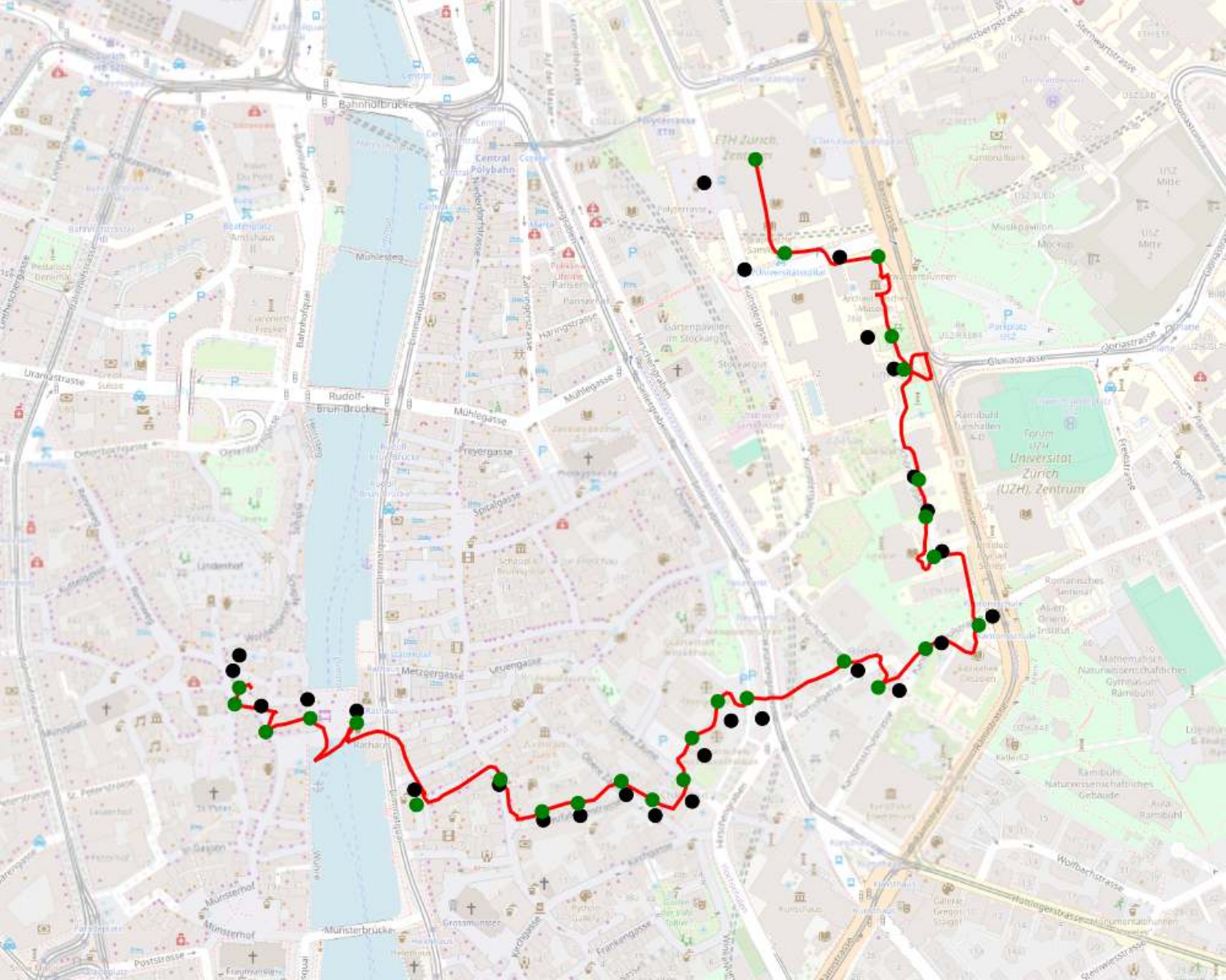} \\
score = 98.9 & score = 88.3 & score = 69.5 & score = 40.5 & score = 16.7 & score = 6.9 
\\

\end{tabular}
\centering
\caption{\textbf{Visualization of nominal trajectories with different scores.} The trajectories are transformed via sparse alignment between the triangulated points (in green) and the target control points (in black).}
\label{fig::supp_score_visualization}
\end{figure*}

\paragraph{Illustration on Different Score Levels}
We show in \cref{fig::supp_score_visualization} nominal trajectories with different levels of scores, to help better interpret the reported scores in the evaluation. 

\paragraph{Score Evaluation on 3D.}
\begin{table*}[tb]
\centering
\scriptsize
\setlength\tabcolsep{5pt}%
\begin{tabular}{l cc cc cc cc cc}
\toprule
\multirow{2}{*}{method} & \multicolumn{2}{c}{short} & \multicolumn{2}{c}{medium} & \multicolumn{2}{c}{long} & \multicolumn{2}{c}{challenge -- low-light} & \multicolumn{2}{c}{challenge -- moving platform} \\
\cmidrule(lr){2-3} \cmidrule(lr){4-5} \cmidrule(lr){6-7} \cmidrule(lr){8-9} \cmidrule(lr){10-11}
& 2D score\,$\uparrow$ & 3D score\,$\uparrow$
& 2D score\,$\uparrow$ & 3D score\,$\uparrow$
& 2D score\,$\uparrow$ & 3D score\,$\uparrow$
& 2D score\,$\uparrow$ & 3D score\,$\uparrow$
& 2D score\,$\uparrow$ & 3D score\,$\uparrow$ \\
\midrule
\mono{DPVO}                         & 9.4 & 3.8 & 5.2 & 1.4 & 1.2 & 0.3 & 3.4 & 1.0 & 2.4 & 1.0 \\
\mono{DPV-SLAM}                     & 7.5 & 3.5 & 5.2 & 2.0 & 0.4 & 0.1 & 1.9 & 1.2 & 1.7 & 0.1 \\
\midrule
\monoinertial{Kimera VIO}           & 6.3 & 4.2 & 6.6 & 4.4 & 6.3 & 3.9 & 4.2 & 2.5 & 7.1 & 2.9 \\
\monoinertial{ORB-SLAM3}            & 28.3 & 18.6 & 20.3 & 12.4 & 14.2 & 7.3 & 6.2 & 2.8 & 15.7 & 8.0  \\
\monoinertial{OpenVINS}             & 18.1 & 13.1 & 10.9 & 7.5 & 4.7 & 2.6 & 7.9 & 5.3 & 2.4 & 1.6 \\
\monoinertial{OpenVINS + Maplab}    & 22.9 & 15.8 & 13.1 & 9.6 & 5.8 & 3.6 & 9.6 & 5.2 & 3.7 & 2.6 \\
\midrule
\binoinertial{OpenVINS}             & 22.2 & 17.1 & 17.8 & 12.7 & 10.6 & 7.6 & 16.9 & 13.3 & 11.5 & 8.6 \\
\binoinertial{OpenVINS + Maplab}    & 26.0 & 19.6 & 21.3 & 14.3 & 12.6 & 7.8 & 16.5 & 12.4 & 13.0 & 9.5 \\
\binoinertial{OKVIS2}               & 24.2 & 20.5 & 13.6 & 9.9 & 3.6 & 1.7 & 15.4 & 11.0 & 4.2 & 3.4 \\

\midrule
\midrule
\binoinertial{Aria's SLAM}           & 90.7 & 87.7 & 78.5 & 73.6 & 70.8 & 65.9 & 84.2 & 82.1 & 53.6 & 46.1 \\
\bottomrule
\end{tabular}
\caption{Score evaluation on 2D and 3D.}
\label{tbl:supp_results:sparse2d3d}%
\end{table*}

Since 72.3\% (349 / 483) of the CPs we have are 2D, we evaluate the score and recall on 2D in the main paper to make use of all the CPs. For completeness, we provide the 3D score evaluation in \cref{tbl:supp_results:sparse2d3d}. Since we use the same scoring function, the 3D scores are consistently lower than the 2D ones, while the relative ranking of the evaluated systems are not largely affected. 

\section{More Results}

\subsection{Variability Analysis}
\begin{table*}[tb]
\centering
\scriptsize
\setlength\tabcolsep{5pt}%
\begin{tabular}{l c c c c c}
\toprule
method & short & medium & long & low-light & moving platform \\
\midrule
\mono{DPVO} & 9.4 $\pm$ 0.8 & 5.2 $\pm$ 1.5 & 1.2 $\pm$ 0.2 & 3.4 $\pm$ 0.7 & 2.4 $\pm$ 0.5 \\
\mono{DPV-SLAM} & 7.5 $\pm$ 0.3 & 5.2 $\pm$ 1.3 & 0.4 $\pm$ 0.6 & 1.9 $\pm$ 0.8 & 1.7 $\pm$ 0.5 \\
\monoinertial{Kimera VIO}  & 6.3 $\pm$ 0.7 & 6.6 $\pm$ 0.9 & 6.3 $\pm$ 1.0 & 4.2 $\pm$ 1.2 & 7.1 $\pm$ 1.0 \\
\monoinertial{ORB-SLAM3} & 28.3 $\pm$ 2.7 & 20.3 $\pm$ 2.5 & 14.2 $\pm$ 1.4 & 6.2 $\pm$ 3.5 & 15.7 $\pm$ 2.6 \\
\monoinertial{OpenVINS} & 18.1 $\pm$ 0.9 & 10.9 $\pm$ 1.0 & 4.7 $\pm$ 0.6 & 7.9 $\pm$ 1.1 & 2.4 $\pm$ 0.5 \\
\monoinertial{OpenVINS + Maplab} & 22.9 $\pm$ 1.3 & 13.1 $\pm$ 1.3 & 5.8 $\pm$ 0.6 & 9.6 $\pm$ 1.2 & 3.7 $\pm$ 1.0\\
\binoinertial{OpenVINS}  & 22.2 $\pm$ 0.8 & 17.8 $\pm$ 0.7 & 10.6 $\pm$ 0.8 & 16.9 $\pm$ 2.1 & 11.5 $\pm$ 1.6\\
\binoinertial{OpenVINS + Maplab}  & 26.0 $\pm$ 1.4 & 21.3 $\pm$ 1.4 & 12.6 $\pm$ 0.7 & 16.5 $\pm$ 2.0 & 13.0 $\pm$ 1.8\\
\binoinertial{OKVIS2}  & 24.2 $\pm$ 0.4 & 13.6 $\pm$ 0.8 & 3.6 $\pm$ 0.5 & 15.4 $\pm$ 1.3 & 4.2 $\pm$ 0.9 \\
\bottomrule
\end{tabular}
\caption{Variability analysis of the reported 2D scores.}
\label{tbl:supp_variability}%
\end{table*}

We report variability of the 2D scores reported in the main paper for the evaluated systems in \cref{tbl:supp_variability}. Since we run the baseline on each sequence for three times, we get three evaluated scores $x_{i1}$, $x_{i2}$, and $x_{i3}$ for each sequence $i$, on which we take the average:
\begin{equation}
    \bar{x}_i = \frac{x_{i1} + x_{i2} + x_{i3}}{3}
\end{equation}
We want to estimate the standard deviation of the average score reported in each group (over $n$ sequences):
\begin{equation}
    \bar{x} = \frac{1}{n}\sum_i^n\bar{x}_i.
\end{equation}

The unbiased estimate (with Bessel's correction) of the standard deviation of $\bar{x}$ follows:

\begin{equation}
\sigma_{\bar{x}} = \sqrt{\frac{1}{6n(n-1)} \sum_{i=1}^{n} \sum_{j=1}^3(x_{ij} - \bar{x}_i)^2}.
\end{equation}

Table \ref{tbl:supp_variability} reports the standard deviation of the 2D scores for the evaluated methods, among which ORB-SLAM3 has the largest variability in the 2D score evaluation. 

\begin{table*}[tb]
\centering
\scriptsize
\setlength\tabcolsep{5pt}%
\begin{tabular}{l cc cc cc cc cc}
\toprule
\multirow{2}{*}{method} & \multicolumn{2}{c}{short} & \multicolumn{2}{c}{medium} & \multicolumn{2}{c}{long} & \multicolumn{2}{c}{low light} & \multicolumn{2}{c}{moving platform} \\
\cmidrule(lr){2-3} \cmidrule(lr){4-5} \cmidrule(lr){6-7} \cmidrule(lr){8-9} \cmidrule(lr){10-11}
& scale & gravity
& scale & gravity
& scale & gravity
& scale & gravity
& scale & gravity
\\
\midrule
\binoinertial{OpenVINS}   & 6.38 & 3.79 & 7.52 & 5.34 & \fail & \fail & \fail & \fail & \fail & \fail \\
\binoinertial{Aria's SLAM} & 0.15 & 0.18 & 0.19 & 0.39 & 0.24 & 0.40 & 0.23 & 0.20 & \fail & \fail \\
\bottomrule
\end{tabular}
\caption{Evaluation of scale error (in percentage) and gravity error (in degree) for OpenVINS \cite{geneva2020openvins} and Aria's SLAM. We mark \fail if the method fails to output a full trajectory in any sequence from the group.}
\label{tbl:supp_results:scale_and_gravity}%
\end{table*}

\subsection{Scale and Gravity}
We further evaluate the scale and gravity of two top-performing methods in our benchmark: OpenVINS \cite{geneva2020openvins} and Aria's SLAM. Specifically, after getting the similarity transformation with sparse alignment, we calculate the scale error (in percentage) as $100|s-1|$ and the gravity error (in degree) as the angular deviation between the rotation in the transformation and the ideal vertical direction. Results are shown in \cref{tbl:supp_results:scale_and_gravity}. To identify if Aria's SLAM suffers from negative scale drift, we further calculate the average value of the scale during sparse alignment across all 63 sequences in four categories: short, medium, long, and low light. Aria's SLAM has an average scale of 1.00222, which represents a negative scale drift of 0.22\%.

\section{More Details on Accuracy Validation}
\label{sec::supp_accuracy_validation}

\begin{figure}[tb]
\centering
\setlength{\pwidth}{0.495\linewidth}
\includegraphics[width=\pwidth]{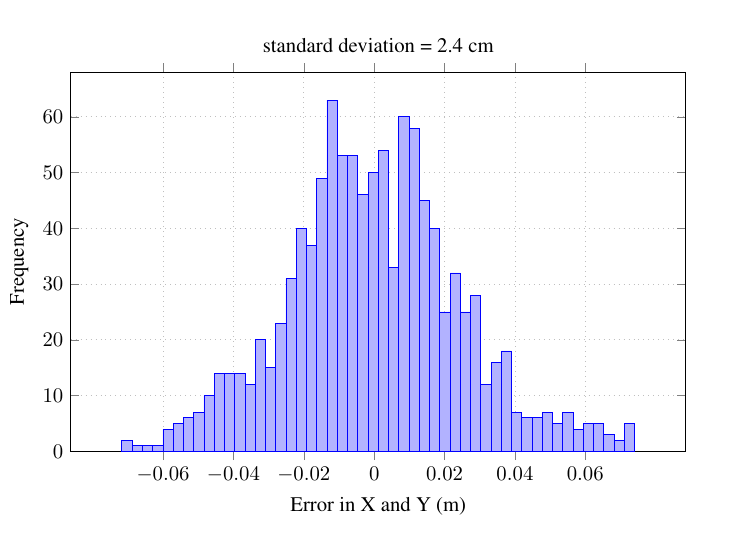}%
\hspace{0.008\linewidth}%
\includegraphics[width=\pwidth]{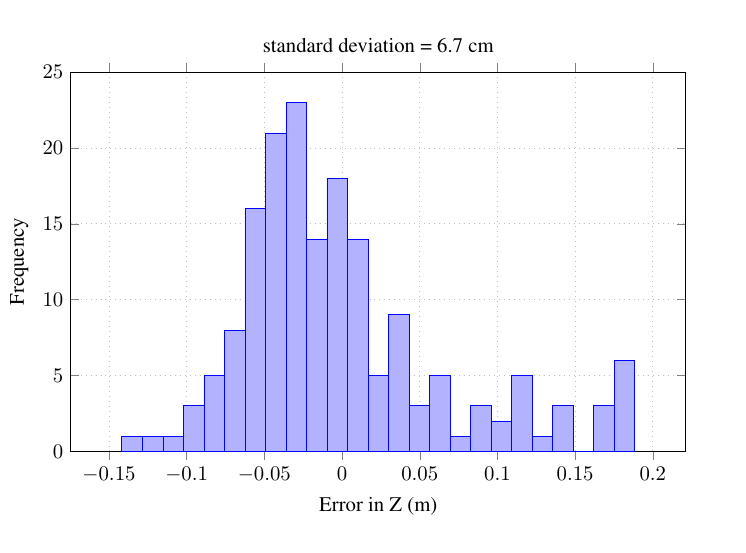}%
\centering
\caption{Distribution of the horizontal (left) and vertical (right) errors between our GNSS-RTK measurements and the public CP data.}
\label{fig::gnss_rtk}
\end{figure}

\subsection{Surveying Statistics}
\label{sec::supp_gnss_rtk}
To validate the accuracy of public data of control points, we additionally measure each of the CP three times with GNSS-RTK, on different days. Our measurements have an uncertainty of $\sim$1.5cm horizontally and 3cm vertically according to the calibration of our surveying instrument. \cref{fig::gnss_rtk} show the distribution of the errors between our measurements and the public control points. Horizontally, the error closely follows a Gaussian distribution with a standard deviation of 2.4cm. Safely assuming that our measurement is independent to the public CP data, we manage to get a similar estimate to the $\sim$1cm uncertainty claimed by the public CP data. This validates the reliability of our main benchmark on evaluating 2D scores with sparse alignment. Since the vertical measurement is often missing in the public CP data, the data points for the vertical errors are sparser. Nonetheless, the distribution of the vertical errors indicate that the z dimension of the control points is centimeter accurate and is sufficient to support the construction of our pseudo ground truth.

\subsection{Validation of Our Visual-Inertial Optimization}
Since Aria's SLAM is a closed-source system, we develop our custom visual-inertial optimization framework to facilitate the construction of dense pseudo-GT poses.
To validate the visual optimization, we add Gaussian noise to both the factory calibrations and the camera poses. Our bundle adjustment is capable of recovering the original focal length and camera poses.
To further validate the inertial optimization, we perform oracle experiments on a stationary recording, i.e., a sequence with negligible device motion. Our inertial optimization on the bias terms is able to compensate noise added on top of the rectified IMU measurements.

\begin{figure}[tb]
\centering
\scriptsize
\setlength\tabcolsep{1pt}
\begin{tabular}{cccccc}
\multicolumn{6}{c}{Level I: platform-based, controlled motion} \\
\includegraphics[height=0.16\linewidth,angle=-90]{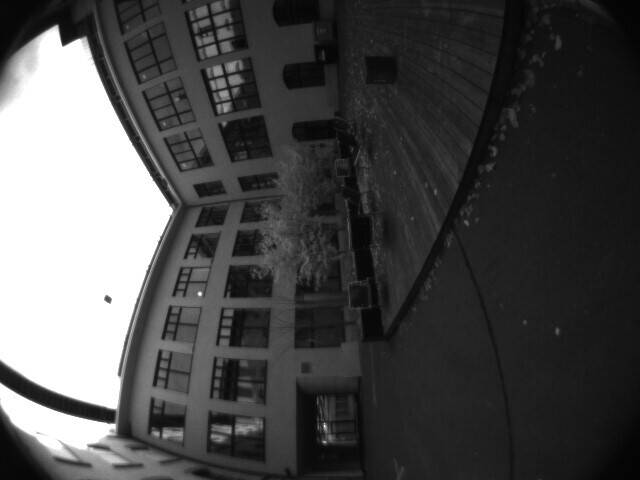} & 
\includegraphics[height=0.16\linewidth,angle=-90]{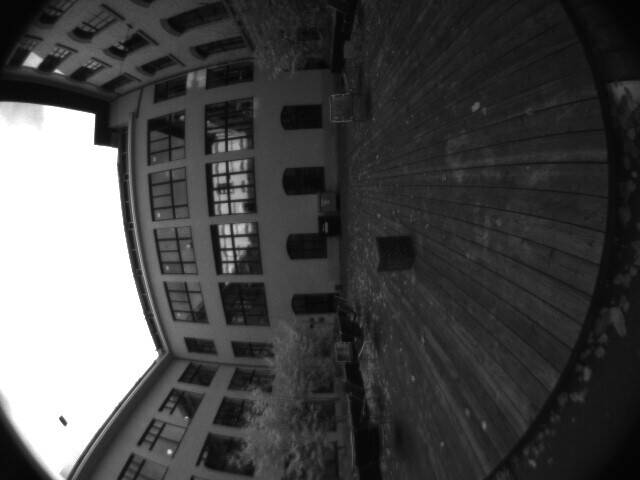} & 
\includegraphics[height=0.16\linewidth,angle=-90]{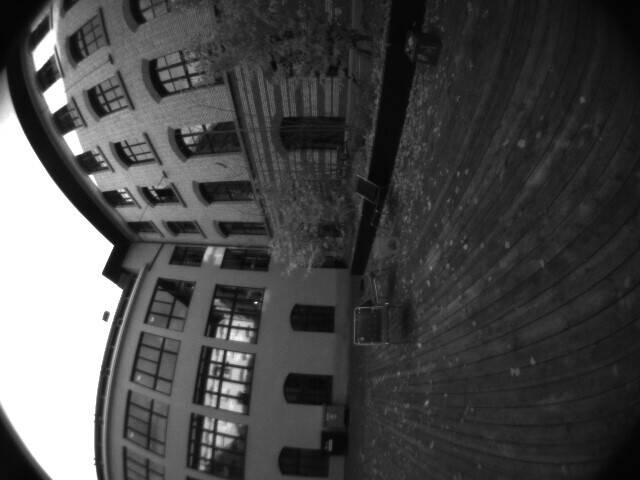} & 
\includegraphics[height=0.16\linewidth,angle=-90]{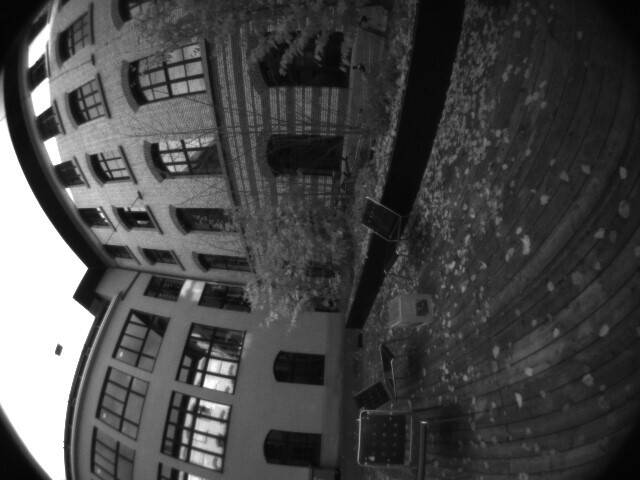} & 
\includegraphics[height=0.16\linewidth,angle=-90]{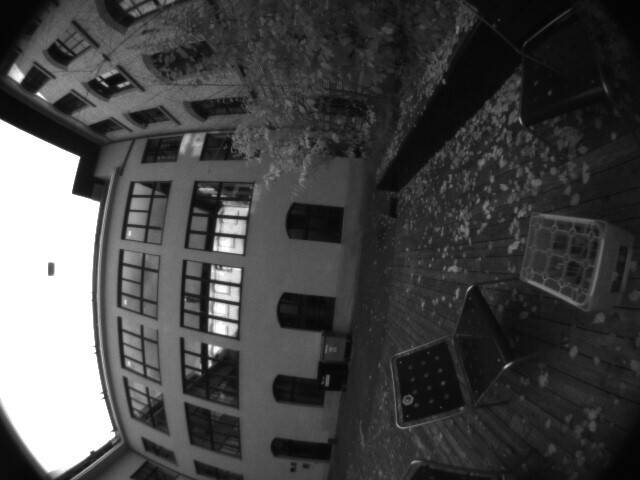} & 
\includegraphics[height=0.16\linewidth,angle=-90]{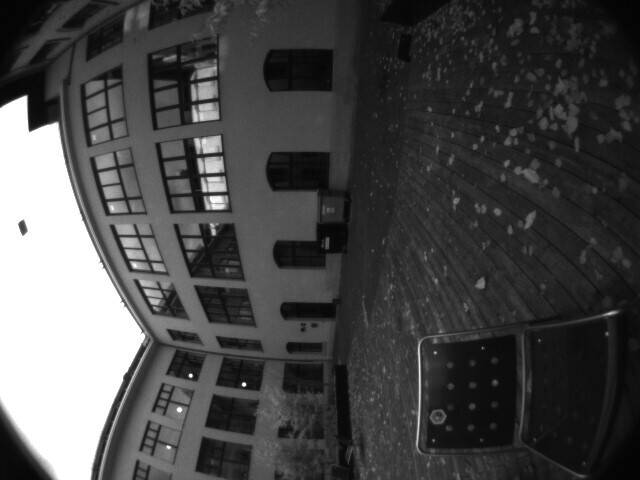} \\
\\
\multicolumn{6}{c}{Level II: platform-based, out-of-plane rotation} \\
\includegraphics[height=0.16\linewidth,angle=-90]{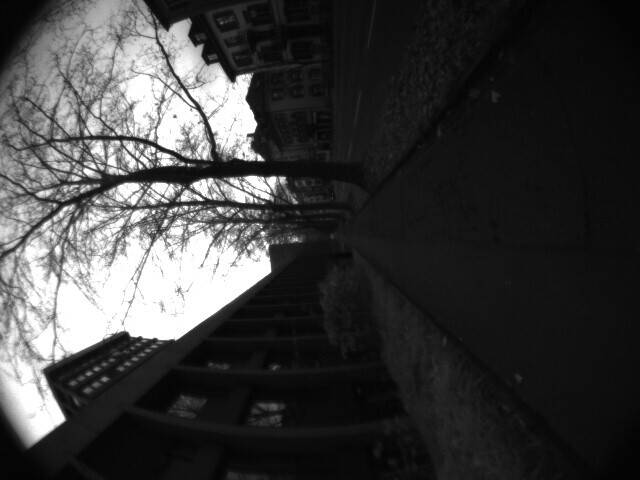} & 
\includegraphics[height=0.16\linewidth,angle=-90]{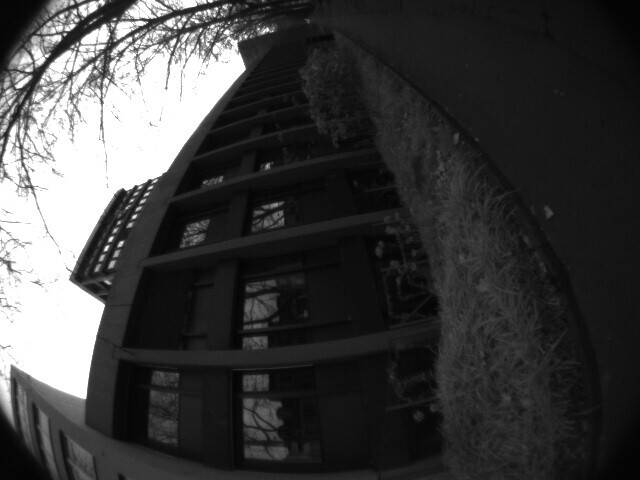} & 
\includegraphics[height=0.16\linewidth,angle=-90]{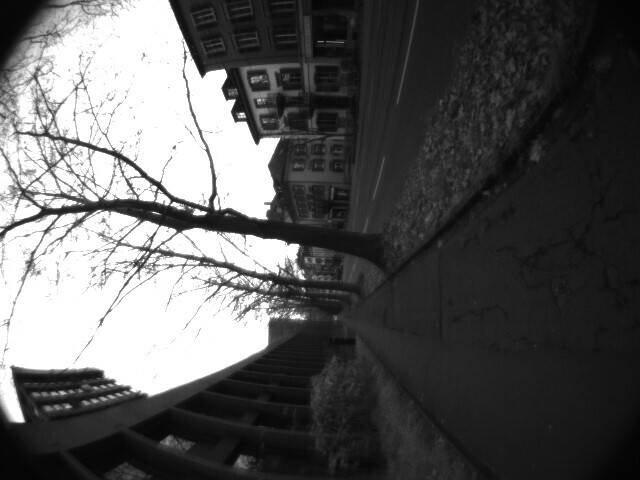} & 
\includegraphics[height=0.16\linewidth,angle=-90]{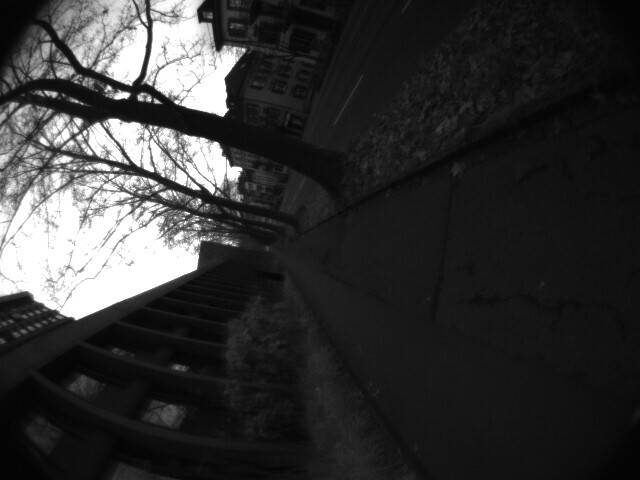} & 
\includegraphics[height=0.16\linewidth,angle=-90]{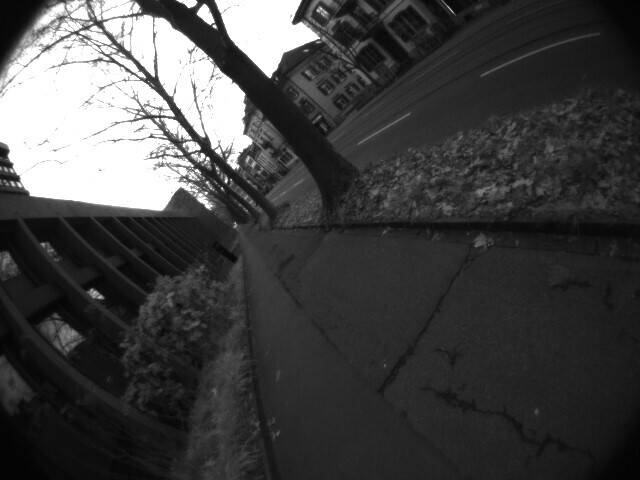} & 
\includegraphics[height=0.16\linewidth,angle=-90]{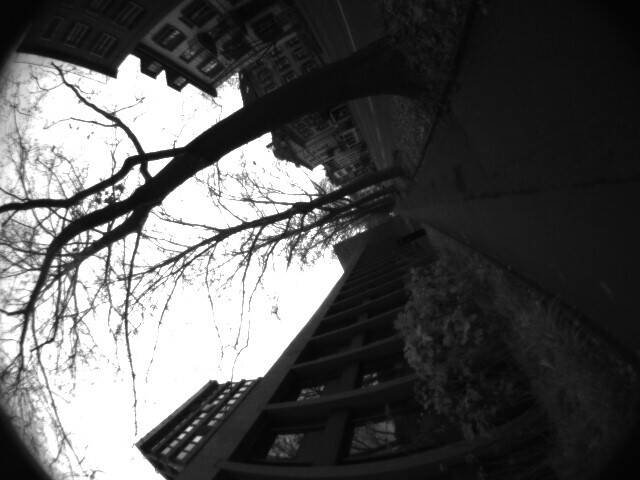} \\
\\
\multicolumn{6}{c}{Level III: platform-based, fast and complex motion} \\
\includegraphics[height=0.16\linewidth,angle=-90]{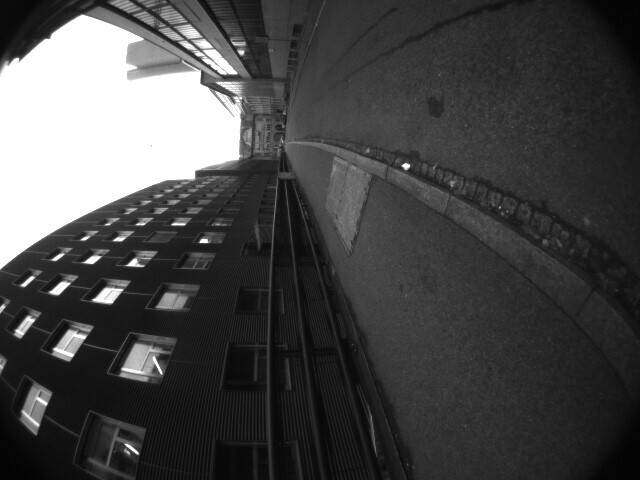} & 
\includegraphics[height=0.16\linewidth,angle=-90]{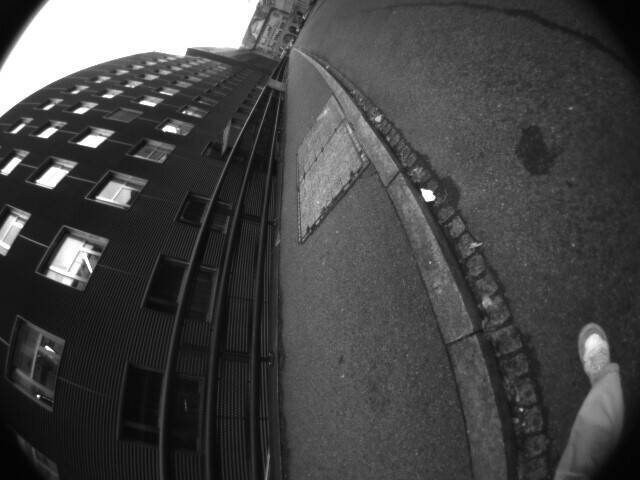} & 
\includegraphics[height=0.16\linewidth,angle=-90]{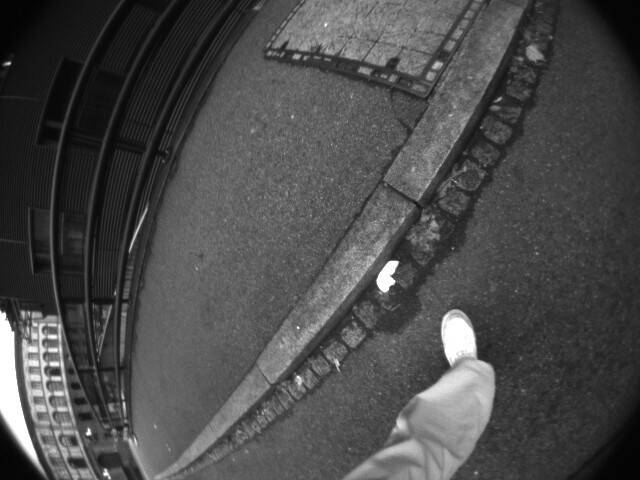} & 
\includegraphics[height=0.16\linewidth,angle=-90]{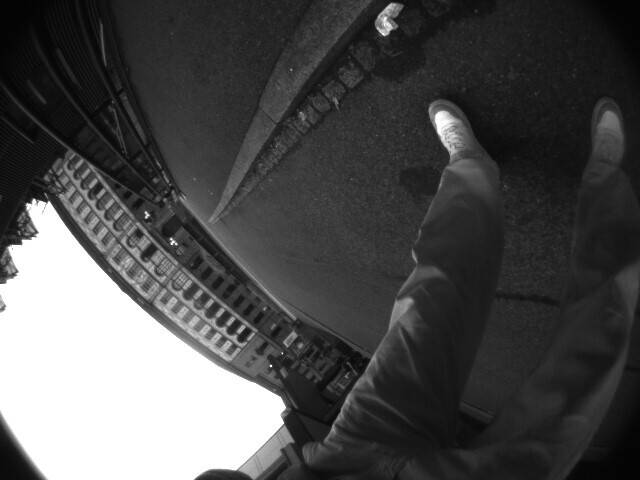} & 
\includegraphics[height=0.16\linewidth,angle=-90]{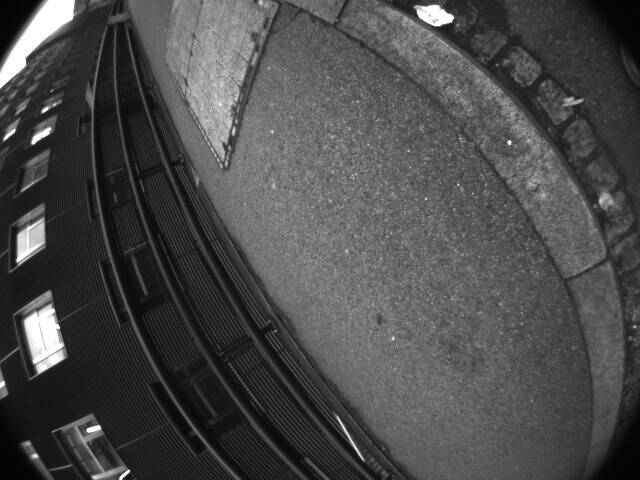} & 
\includegraphics[height=0.16\linewidth,angle=-90]{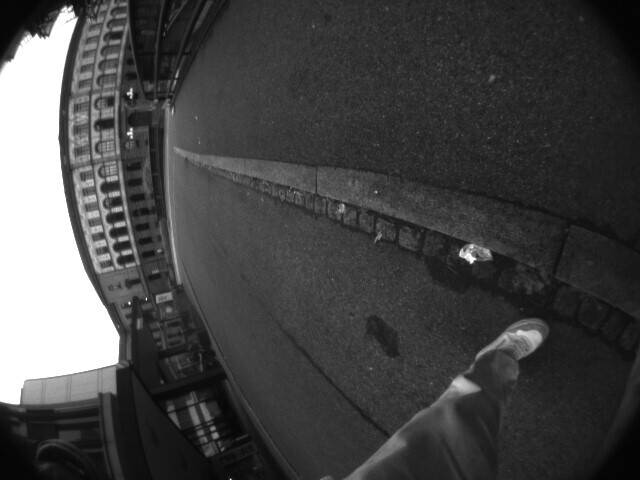} \\
\\
\multicolumn{6}{c}{Level IV: egocentric recordings} \\
\includegraphics[height=0.16\linewidth,angle=-90]{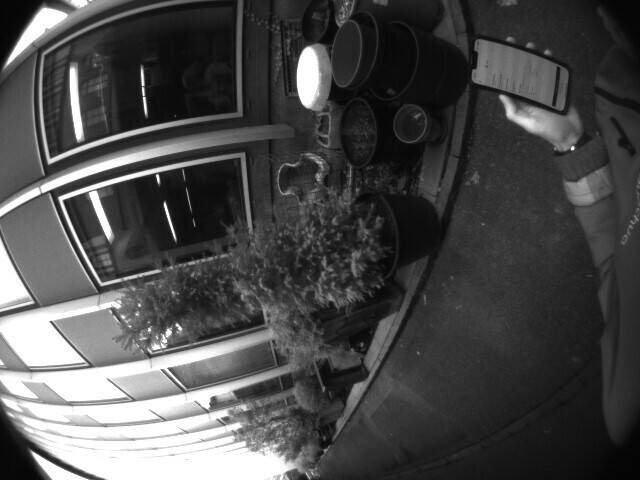} & 
\includegraphics[height=0.16\linewidth,angle=-90]{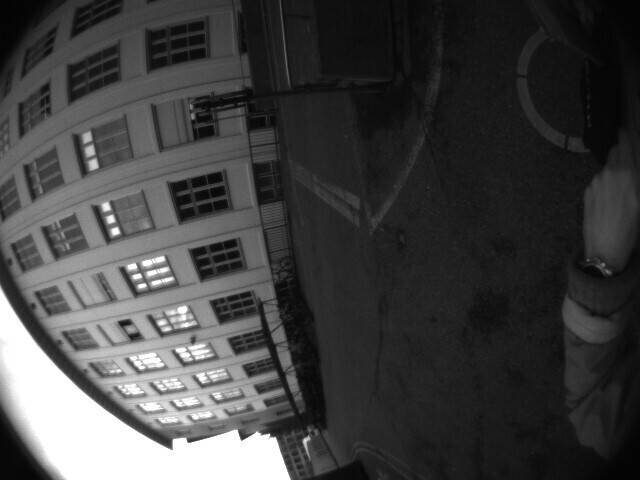} & 
\includegraphics[height=0.16\linewidth,angle=-90]{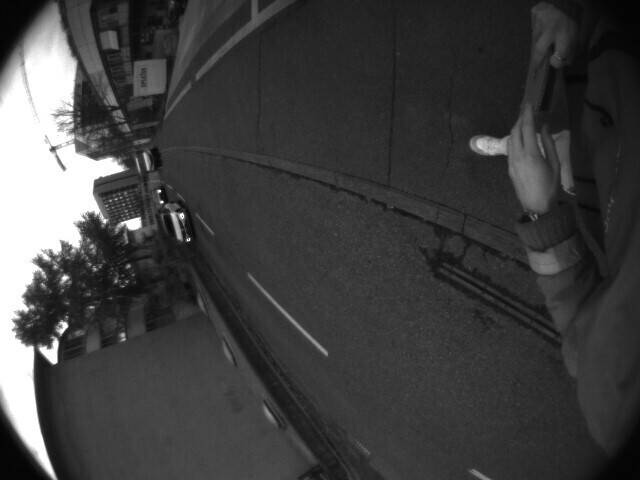} & 
\includegraphics[height=0.16\linewidth,angle=-90]{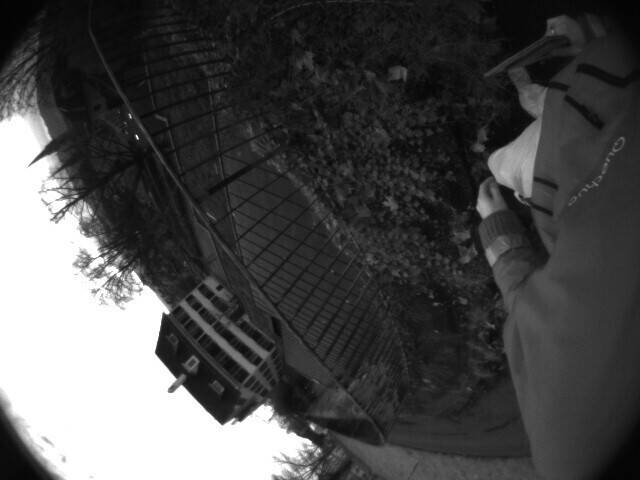} & 
\includegraphics[height=0.16\linewidth,angle=-90]{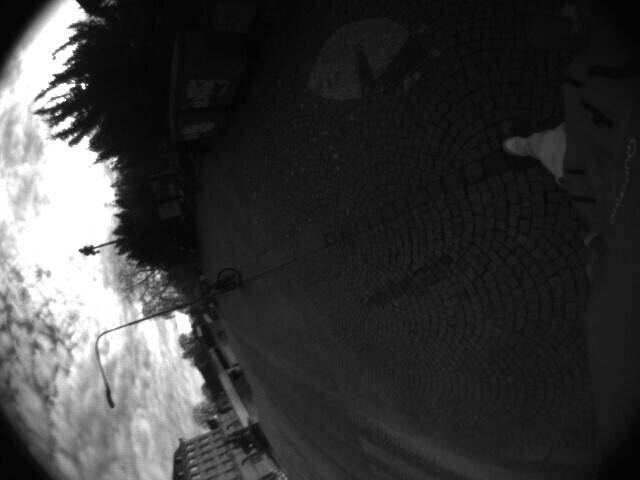} & 
\includegraphics[height=0.16\linewidth,angle=-90]{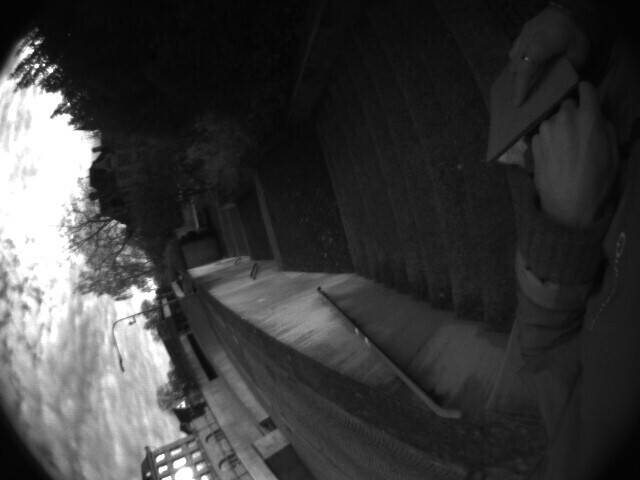} \\
\end{tabular}
\centering
\caption{Qualitative visualizations of the nominal motion patterns in the four levels of controlled experimental set.}
\label{fig::supp_examples_experimental_set}
\end{figure}

\begin{table}[tb]
\centering
\scriptsize
\setlength\tabcolsep{3pt} %
\begin{tabular}{l cccc}
\toprule
\multicolumn{5}{c}{\textbf{Level I (3 sequences)}} \\
\midrule
Sequence ID & motion & \# CPs & duration (in min) & length (in km) \\
\midrule
R\_01\_easy & platform-based & N/A & 2.41 & 0.16\\
R\_02\_easy & platform-based & N/A & 2.48 & 0.18\\
R\_03\_easy & platform-based & N/A & 2.64 & 0.20\\
\midrule
Average & - & N/A & 2.51 & 0.18 \\
\midrule
\midrule
\multicolumn{5}{c}{\textbf{Level II (4 sequences)}} \\
\midrule
Sequence ID & motion & \# CPs & duration (in min) & length (in km) \\
\midrule
R\_04\_medium & platform-based & N/A & 4.38 & 0.32\\
R\_05\_medium & platform-based & N/A & 5.13 & 0.47\\
R\_06\_medium & platform-based & N/A & 6.51 & 0.62\\
R\_07\_medium & platform-based & N/A & 6.67 & 0.60\\
\midrule
Average & - & N/A & 5.67 & 0.50 \\
\midrule
\midrule
\multicolumn{5}{c}{\textbf{Level III (3 sequences)}} \\
\midrule
Sequence ID & motion & \# CPs & duration (in min) & length (in km) \\
\midrule
R\_08\_hard & platform-based & N/A & 10.27 & 0.75\\
R\_09\_hard & platform-based & N/A & 13.01 & 0.94\\
R\_10\_hard & platform-based & N/A & 15.60 & 1.34\\
\midrule
Average & - & N/A & 12.96 & 1.01 \\
\midrule
\midrule
\multicolumn{5}{c}{\textbf{Level IV (3 sequences)}} \\
\midrule
Sequence ID & motion & \# CPs & duration (in min) & length (in km) \\
\midrule
R\_11\_5cp & egocentric & 5 & 7.96 & 0.48\\
R\_12\_10cp & egocentric & 10 & 16.90 & 1.01\\
R\_13\_15cp & egocentric & 15 & 23.44 & 1.17\\
\midrule
Average & - & 10 & 16.10 & 0.89 \\
\bottomrule
\end{tabular}
\caption{Detailed statistics for the controlled experimental set.
}
\label{tbl:supp_experimental_set_stats}
\end{table}

\subsection{Covariance Estimation}
To know when this pseudo-GT is reliable, we compute uncertainties on the device poses as the inverse of the Hessian matrix of the least-squares optimization (Laplace's approximation). We calculate all the 6x6 on-manifold pose covariance from our joint optimization on a test sequence with 12 control points available. The median value of the positional uncertainty across sequences is 20.0 centimeters. Although our dense pseudo ground truth poses is not as accurate as our survey-grade control points, they are sufficiently accurate to measure keyframe errors larger than 50.0 centimeters for trajectories that span kilometers.

\section{Inertial-only optimization}

\label{sec::supp_imu_only}

Traveling in a moving platform poses unique challenges for visual-inertial odometry and SLAM due to the inconsistency between the visual signals and the actual motion. When the camera is moving with the vehicle, the visual features inside the vehicle are potentially misleading and only give constraints to the relative motion between the camera and the vehicle. This often results in tracking failures of visual-inertial systems, including Aria's SLAM API. 

\begin{figure}[tb]
\centering
\includegraphics[width=0.45\linewidth, height=100pt]{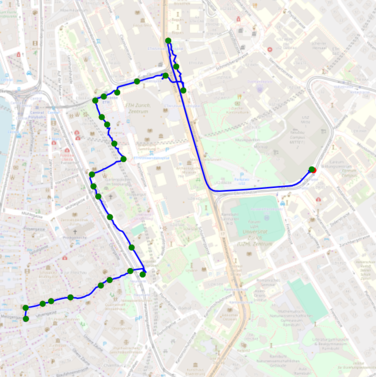}%
\hspace{0.0016pt}
\includegraphics[width=0.45\linewidth, height=100pt]{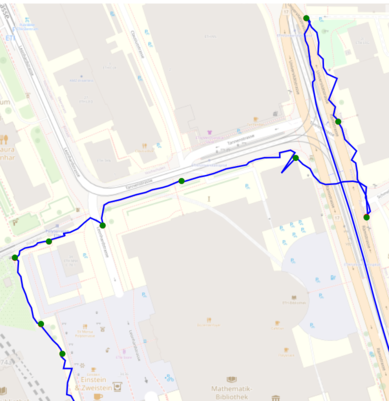}%
\centering
\caption{\textbf{Result of inertial-only optimization on a moving platform.} While the visual constraints can potentially mislead the motion estimation, one can achieve reasonable results \textbf{(left)} by optimizing only with inertial preintegration factors \cite{imupreintegration} and control points. However, the inertial-only optimization results suffer from local flickering \textbf{(right)} and are thus not sufficiently accurate to be used as dense pseudo GT poses for per-keyframe evaluation.
}%
\label{fig::imu_only_optimization}
\end{figure}

As discussed in the main paper, in the moving platform section, we may rely only on the inertial and CP information for our joint optimization, while dropping the visual constraints. As shown in Fig. \ref{fig::imu_only_optimization}, the inertial-only optimization achieves reasonable trajectory prediction that aligns with the movement of the vehicle. However, due to lack of accurate per-frame measurements in the optimization, the resulting poses suffer from local flickering and is thus not accurate enough to serve as the pseudo ground truth in the evaluation. 

\begin{table}[tb]
\centering
\scriptsize
\setlength\tabcolsep{3pt}%
\begin{tabular}{l c c c c c c}
\toprule
\multirow{2}{*}{Method} & \multicolumn{1}{c}{\shortstack{outdoor \\ walking}} & \multicolumn{1}{c}{\shortstack{exposure \\ change}} & \multicolumn{1}{c}{\shortstack{low \\ light}} & \multicolumn{1}{c}{\shortstack{fast \\ motion}} & \multicolumn{1}{c}{\shortstack{dynamic \\ scenes}} \\
\cmidrule(lr){2-6}
& R@50cm\,$\uparrow$ & R@50cm\,$\uparrow$ & R@50cm\,$\uparrow$ & R@50cm\,$\uparrow$ & R@50cm\,$\uparrow$ \\
\midrule
\mono{COLMAP} & 28.3 & 23.9 & 7.5 & 46.3 & 80.0 \\
\mono{GLOMAP} & 64.1 & 42.9 & 15.6 & 78.5 & 62.7 \\
\mono{ORB-SLAM3} & 31.3 & 21.8 & 14.7 & 40.5 & 31.7\\
\midrule
\monoinertial{ORB-SLAM3} & 41.7 & 40.9 & 17.5 & 61.0 & 89.2 \\
\monoinertial{OpenVINS} & 54.7 & 55.1 & 22.2 & 74.6 & 34.3 \\
\bottomrule
\end{tabular}
\caption{Evaluation results on selected short snippets w.r.t. our dense pseudo ground-truth poses.}
\label{tbl:results:snippets}
\end{table}

\section{A Study on Short Snippets}

To evaluate against methods that are comparably heavy and that cannot scale well to the long sequences, we select shorter 2-min segments from the sequences. In particular, we focus on parts where Aria's SLAM is the most accurate and the dense ground truth is sufficiently accurate, while also covering unique egocentric challenges in the dataset. This results in a total of 27 sequence categorized in five groups: outdoor walking (9 sequences), exposure change (6 sequences), low light (3 sequences), fast motion (6 sequences), and dynamic scenes (3 sequences). We evaluate two widely recognized SfM methods: COLMAP \cite{schoenberger2016sfm} and GLOMAP \cite{pan2024global}, and include the top-performing VIO/SLAM methods on our benchmark: ORB-SLAM3 (monocular and monocular-inertial) \cite{orbslam3} and OpenVINS (monocular-inertial) \cite{geneva2020openvins}. The SfM pipelines are unable to produce a decent output on our full‑length sequences, so we only run them on the selected snippets. Similar to the practice for the full-sequence evaluation, we feed in the factory calibration for each of tested method and apply undistortion beforehand if necessary. 

\begin{table*}[tb]
\centering
\scriptsize
\setlength\tabcolsep{1pt}%
\begin{tabular}{l c ccc c ccc c ccc c ccc c ccc}
\toprule
\multirow{2}{*}{method} & \multirow{2}{*}{causal} & \multicolumn{3}{c}{short} & & \multicolumn{3}{c}{medium} & &\multicolumn{3}{c}{long} & & \multicolumn{3}{c}{challenge -- low-light} & & \multicolumn{3}{c}{challenge -- moving platform} \\
\cmidrule(lr){3-5} \cmidrule(lr){7-9} \cmidrule(lr){11-13} \cmidrule(lr){15-17} \cmidrule(lr){19-21}
& & score\,$\uparrow$ & CP@1m\,$\uparrow$ & R@5m\,$\uparrow$
& & score\,$\uparrow$ & CP@1m\,$\uparrow$ & R@5m\,$\uparrow$
& & score\,$\uparrow$ & CP@1m\,$\uparrow$ & R@5m\,$\uparrow$
& & score\,$\uparrow$ & CP@1m\,$\uparrow$ & R@5m\,$\uparrow$
& & score\,$\uparrow$ & CP@1m\,$\uparrow$ & R@5m\,$\uparrow$\\
\midrule
\mono{DPVO}                         & \checkmark & 0.0 & 0.0 & 0.7 & & 0.0 & 0.0 & 0.0 & & 0.0 & 0.0 & 0.4 & & 0.9 & 0.0 & 2.7 & & 1.5 & 0.0 & \na \\
\mono{DPV-SLAM}                     & x          & 0.0 & 0.0 & 1.2 & & 1.3 & 0.0 & 2.1 & & 0.0 & 0.0 & 0.5 & & 1.7 & 0.0 & 3.2 & & 1.3 & 0.0 & \na \\
\midrule
\monoinertial{Kimera VIO}           & \checkmark & 7.8 & 0.0 & 13.2 & & 1.7 & 2.7 & 2.0 & & 4.9 & 2.0 & 8.7 & & 17.7 & 3.1 & 44.1 & & 10.0 & 6.3 & \na \\ 
\monoinertial{ORB-SLAM3}            & x          & 11.2 & 0.0 & 27.1 & & 0.0 & 0.0 & 0.0 & & 4.2 & 0.0 & 7.8 & & 0.0 & 0.0 & 0.0 & & 19.8 & 2.7 & \na \\
\monoinertial{OpenVINS}             & \checkmark & 25.2 & 7.4 & 66.5 & & 5.8 & 0.0 & 12.5 & & 4.2 & 0.0 & 12.0 & & 0.0 & 0.0 & 0.0 & & 6.7  & 2.7  & \na \\
\monoinertial{OpenVINS + Maplab}    & x          & 27.1 & 7.6 & 69.3 & & 6.0 & 0.1 & 15.1 & & 4.7 & 0.0 & 12.9 & & 0.0 & 0.0 & 0.0 & & 7.3  & 2.8  & \na \\
\midrule
\binoinertial{OpenVINS}             & \checkmark & 33.2 & 8.3 & 86.6 & & 12.4 & 2.8 & 23.5 & & 5.5 & 0.0 & 11.9 & & 9.9  & 3.3 & 22.8 & & 9.2 & 0.0 & \na \\
\binoinertial{OpenVINS + Maplab}    & x          & 35.3 & 9.1 & 88.1 & & 12.8 & 3.0 & 25.7 & & 5.7 & 0.0 & 12.1 & & 10.2 & 3.5 & 23.1 & & 9.4 & 0.0 & \na \\
\binoinertial{OKVIS2}               & x          & 36.7 & 11.2 & 78.6 & & 9.9 & 0.0 & 22.2 & & 12.3 & 0.0 & 38.8 & & 1.1 & 0.0 & 2.1 & & 20.3 & 3.6 & \na \\

\midrule
\midrule
\binoinertial{Aria's SLAM}           & x         & 96.8 & 100.0 & \na & & 82.3 & 95.0 & \na & & 91.5 & 100.0 & \na & & 80.6 & 96.8 & \na & & 39.8 & 46.4 & \na \\
\bottomrule
\end{tabular}
\caption{\textbf{Evaluation on the additional set.} The 10 additional sequences were captured for the data release as described in \cref{supp:data_release}.}
\label{tbl:results:data_release_long}%
\end{table*}

\begin{table}[tb]
\centering
\scriptsize
\setlength\tabcolsep{3pt} %
\begin{tabular}{l c c c c c l}
  \toprule
  \multicolumn{7}{c}{\textbf{Additional set}} \\
  \midrule
  Sequence ID
    & motion
    & type
    & \# CPs
    & \makecell{duration\\(in min)}
    & \makecell{length\\(in km)}
    & challenge \\
  \midrule
Sequence 1-19 & egocentric & rw  & 14 & 15.31 & 1.53 & short \\
Sequence 1-20 & egocentric & rw  & 13 & 16.90 & 1.72 & short\\
Sequence 2-11 & egocentric & dma & 18 & 19.80 & 2.05 & medium\\
Sequence 2-12 & egocentric & dma & 20 & 27.95 & 3.07 & medium\\
Sequence 3-17 & egocentric & a\_b & 27 & 29.90 & 3.09 & long \\
Sequence 3-18 & egocentric & a\_b & 24 & 28.25 & 2.74 & long \\
Sequence 4-10 & egocentric & dma & 16 & 23.70 & 2.23 & low light \\
Sequence 4-11 & egocentric & dma & 15 & 18.80 & 1.79 & low light \\
Sequence 5-11 & egocentric & dma & 14 & 18.76 & 1.76 & moving platform\\
Sequence 5-12 & egocentric & dma & 18 & 22.23 & - & moving platform \\
\bottomrule
\end{tabular}
\caption{\textbf{Detailed per-sequence statistics for the additional set} (for the data release as described in \cref{supp:data_release}).}
\label{tbl:data_release_stats}
\end{table}

Table \ref{tbl:results:snippets} presents the recall evaluation at 50 centimeters using our dense pseudo ground-truth poses. The results indicate that while structure-from-motion (SfM) methods are not specifically designed for video sequences, they achieve higher accuracy compared to monocular SLAM approaches. This can be attributed to their offline nature, which allows for large-scale bundle adjustment. However, visual-inertial systems demonstrate superior performance over both SfM and visual odometry/SLAM methods in challenging conditions, such as facing exposure variations or low-light environments, where visual cues are less reliable.

\section{Data Release}
\label{supp:data_release}

Our training set comprises 13 sequences of the controlled experimental set and 10 additional sequences (two for each of the five main dataset challenge categories). The statistics and benchmarking for the 10 additional sequences are given in  \cref{tbl:results:data_release_long} and \cref{tbl:data_release_stats}. For every training sequence we release the raw data, factory calibrations, sparse and pseudo‐dense ground‐truth. The test set consists of all 63 sequences from the main dataset, for which we release only the raw data and factory calibrations. 

\section{More Visualizations}
We provide more qualitative examples of the recordings in Figures \ref{fig:supp_examples_full_set_1}, \ref{fig:supp_examples_full_set_2}, and \ref{fig:supp_examples_full_set_3} covering challenges that are unique to egocentric data.

\newcommand{\yes}{\checkmark}%
\newcommand{\no}{x}%

\begin{table*}[tb]
\centering
\scriptsize
\setlength\tabcolsep{5pt} %
\begin{tabular}{l ccccc cccc}
\toprule
\multicolumn{10}{c}{\textbf{Group 1: Short (18 sequences)}} \\
\midrule
\multirow{2}{*}{Sequence ID} & \multirow{2}{*}{motion} &  \multirow{2}{*}{type} & \multirow{2}{*}{\# CPs} & \multirow{2}{*}{duration (in min)} & \multirow{2}{*}{length (in km)} & low & moving & indoor-outdoor & dynamic \\
& & & & & & light & platform & transition & scenes \\
\midrule
Sequence 1-1 & handheld & dma & 12 & 15.97 & 1.16 & \no & \no & \no & \no \\
Sequence 1-2 & egocentric & dma & 15 & 21.95 & 1.43 & \no & \no & \no & \no \\
Sequence 1-3 & egocentric & dma & 13 & 17.95 & 1.05 & \no & \no & \yes & \no \\
Sequence 1-4 & egocentric & dma & 15 & 23.77 & 1.63 & \no & \no & \no & \no \\
Sequence 1-5 & egocentric & dma & 12 & 15.10 & 0.59 & \no & \no & \no & \yes \\
Sequence 1-6 & egocentric & dma & 11 & 18.52 & 0.94 & \no & \no & \no & \yes \\
Sequence 1-7 & egocentric & dma & 14 & 19.15 & 1.00 & \no & \no & \no & \no \\
Sequence 1-8 & egocentric & dma & 14 & 17.92 & 0.85 & \no & \no & \no & \no \\
Sequence 1-9 & egocentric & dma & 13 & 16.92 & 0.81 & \no & \no & \no & \no \\
Sequence 1-10 & egocentric & dma & 15 & 19.87 & 0.90 & \no & \no & \no & \yes \\
Sequence 1-11 & egocentric & rw & 14 & 12.48 & 0.78 & \no & \no & \no & \no \\
Sequence 1-12 & egocentric & rw & 13 & 19.00 & 1.04 & \no & \no & \no & \no \\
Sequence 1-13 & egocentric & rw & 14 & 21.38 & 1.16 & \no & \no & \no & \yes \\
Sequence 1-14 & egocentric & rw & 15 & 20.03 & 1.19 & \no & \no & \no & \no \\
Sequence 1-15 & egocentric & rw & 14 & 18.67 & 1.01 & \no & \no & \no & \yes \\
Sequence 1-16 & egocentric & rw & 14 & 19.15 & 1.14 & \no & \no & \no & \yes \\
Sequence 1-17 & egocentric & rw & 15 & 18.38 & 1.09 & \no & \no & \no & \no \\
Sequence 1-18 & handheld & dma & 5 & 5.60 & 0.48 & \no & \no & \no & \no \\
\midrule
Average & - & - & 13.2 & 17.88 & 1.01 & - & - & - & - \\
\midrule
\midrule
\multicolumn{10}{c}{\textbf{Group 2: Medium (10 sequences)}} \\
\midrule
\multirow{2}{*}{Sequence ID} & \multirow{2}{*}{motion} &  \multirow{2}{*}{type} & \multirow{2}{*}{\# CPs} & \multirow{2}{*}{duration (in min)} & \multirow{2}{*}{length (in km)} & low & moving & indoor-outdoor & dynamic \\
& & & & & & light & platform & transition & scenes \\
\midrule
Sequence 2-1 & handheld & a\_b & 21 & 28.42 & 1.65 & \no & \no & \yes & \no \\
Sequence 2-2 & handheld & a\_b & 22 & 28.27 & 1.74 & \no & \no & \yes & \no \\
Sequence 2-3 & egocentric & dma & 17 & 22.18 & 1.31 & \no & \no & \yes & \no \\
Sequence 2-4 & egocentric & dma & 19 & 29.23 & 1.88 & \no & \no & \yes & \no \\
Sequence 2-5 & egocentric & dma & 20 & 28.33 & 1.87 & \no & \no & \yes & \yes \\
Sequence 2-6 & egocentric & dma & 16 & 24.22 & 1.62 & \no & \no & \no & \no \\
Sequence 2-7 & egocentric & dma & 16 & 22.25 & 1.53 & \no & \no & \yes & \no \\
Sequence 2-8 & egocentric & dma & 16 & 13.35 & 0.69 & \no & \no & \no & \no \\
Sequence 2-9 & egocentric & dma & 18 & 24.87 & 1.42 & \no & \no & \no & \yes \\
Sequence 2-10 & egocentric & rw & 18 & 21.15 & 1.27 & \no & \no & \no & \yes \\
\midrule
Average & - & - & 18.3 & 25.23 & 1.46 & - & - & - & - \\
\midrule
\midrule
\multicolumn{10}{c}{\textbf{Group 3: Long (16 sequences)}} \\
\midrule
\multirow{2}{*}{Sequence ID} & \multirow{2}{*}{motion} &  \multirow{2}{*}{type} & \multirow{2}{*}{\# CPs} & \multirow{2}{*}{duration (in min)} & \multirow{2}{*}{length (in km)} & low & moving & indoor-outdoor & dynamic \\
& & & & & & light & platform & transition & scenes \\
\midrule
Sequence 3-1 & handheld & a\_b & 27 & 33.40 & 1.84 & \no & \no & \no & \no \\
Sequence 3-2 & handheld & a\_b & 26 & 28.53 & 1.67 & \no & \no & \no & \no \\
Sequence 3-3 & handheld & a\_b & 26 & 33.87 & 1.79 & \no & \no & \no & \no \\
Sequence 3-4 & egocentric & a\_b & 27 & 31.47 & 1.85 & \no & \no & \no & \yes \\
Sequence 3-5 & egocentric & a\_b & 27 & 29.27 & 1.79 & \no & \no & \yes & \yes \\
Sequence 3-6 & egocentric & a\_b & 27 & 29.78 & 1.71 & \no & \no & \no & \no \\
Sequence 3-7 & egocentric & a\_b & 25 & 30.62 & 1.35 & \no & \no & \no & \no \\
Sequence 3-8 & egocentric & a\_b & 30 & 42.83 & 2.60 & \no & \no & \no & \yes \\
Sequence 3-9 & egocentric & a\_b & 26 & 36.42 & 2.13 & \no & \no & \no & \no \\
Sequence 3-10 & egocentric & a\_b & 28 & 48.00 & 2.87 & \no & \no & \yes & \yes \\
Sequence 3-11 & egocentric & a\_b & 27 & 40.63 & 2.35 & \no & \no & \no & \no \\
Sequence 3-12 & egocentric & a\_b & 28 & 37.27 & 2.40 & \no & \no & \no & \no \\
Sequence 3-13 & egocentric & a\_b & 26 & 35.70 & 2.35 & \no & \no & \no & \no \\
Sequence 3-14 & egocentric & a\_b & 26 & 32.05 & 1.98 & \no & \no & \no & \no \\
Sequence 3-15 & egocentric & a\_b & 26 & 37.37 & 2.30 & \no & \no & \no & \yes \\
Sequence 3-16 & egocentric & a\_b & 27 & 37.52 & 2.42 & \no & \no & \no & \no \\
\midrule
Average & - & - & 26.8 & 40.30 & 1.99 & - & - & - & - \\
\bottomrule
\end{tabular}
\caption{\textbf{Detailed per-sequence statistics for the main dataset} (short, medium, long).}%
\label{tbl:supp_recording_stats_part1}%
\end{table*}

\begin{table*}[tb]
\centering
\scriptsize
\setlength\tabcolsep{5pt} %
\begin{tabular}{l ccccc cccc}
\toprule
\multicolumn{10}{c}{\textbf{Group 4: Challenge - low light (9 sequences)}} \\
\midrule
\multirow{2}{*}{Sequence ID} & \multirow{2}{*}{motion} &  \multirow{2}{*}{type} & \multirow{2}{*}{\# CPs} & \multirow{2}{*}{duration (in min)} & \multirow{2}{*}{length (in km)} & low & moving & indoor-outdoor & dynamic \\
& & & & & & light & platform & transition & scenes \\
\midrule
Sequence 4-1 & egocentric & a\_b & 30 & 34.93 & 1.95 & \yes & \no & \yes & \no \\
Sequence 4-2 & egocentric & dma & 16 & 26.75 & 1.64 & \yes & \no & \yes & \no \\
Sequence 4-3 & egocentric & dma & 14 & 23.07 & 1.29 & \yes & \no & \no & \no \\
Sequence 4-4 & egocentric & dma & 16 & 26.20 & 1.58 & \yes & \no & \no & \no \\
Sequence 4-5 & egocentric & dma & 15 & 25.50 & 1.53 & \yes & \no & \no & \no \\
Sequence 4-6 & egocentric & dma & 15 & 18.68 & 0.90 & \yes & \no & \no & \no \\
Sequence 4-7 & egocentric & dma & 16 & 20.70 & 0.95 & \yes & \no & \no & \no \\
Sequence 4-8 & egocentric & dma & 14 & 22.02 & 1.21 & \yes & \no & \no & \no \\
Sequence 4-9 & egocentric & dma & 13 & 15.77 & 0.87 & \yes & \no & \no & \no \\
\midrule
Average & - & - & 16.5 & 23.41 & 1.32 & - & - & - & - \\
\midrule
\midrule
\multicolumn{10}{c}{\textbf{Group 5: Challenge - moving platform (10 sequences)}} \\
\midrule
\multirow{2}{*}{Sequence ID} & \multirow{2}{*}{motion} &  \multirow{2}{*}{type} & \multirow{2}{*}{\# CPs} & \multirow{2}{*}{duration (in min)} & \multirow{2}{*}{length (in km)} & low & moving & indoor-outdoor & dynamic \\
& & & & & & light & platform & transition & scenes \\
\midrule
Sequence 5-1 & handheld & a\_b & 25 & 39.93 & 2.41 & \no & \yes & \no & \no \\
Sequence 5-2 & handheld & a\_b & 27 & 32.32 & 2.33 & \no & \yes & \no & \no \\
Sequence 5-3 & handheld & a\_b & 27 & 41.05 & 2.22 & \no & \yes & \yes & \no \\
Sequence 5-4 & handheld & a\_b & 22 & 27.92 & 2.17 & \no & \yes & \yes & \no \\
Sequence 5-5 & egocentric & a\_b & 27 & 35.20 & $\sim$ 2.17 & \no & \yes & \yes & \no \\
Sequence 5-6 & egocentric & a\_b & 29 & 41.43 & $\sim$ 2.23 & \yes & \yes & \yes & \no \\
Sequence 5-7 & egocentric & dma & 15 & 20.62 & $\sim$ 1.10 & \no & \yes & \no & \yes \\
Sequence 5-8 & egocentric & dma & 15 & 21.12 & 1.08 & \no & \yes & \no & \yes \\
Sequence 5-9 & egocentric & dma & 15 & 24.65 & 2.25 & \no & \yes & \no & \no \\
Sequence 5-10 & egocentric & dma & 16 & 26.60 & 2.04 & \no & \yes & \no & \no \\
\midrule
Average & - & - & 21.8 & 29.93 & $\sim$ 2.00 & - & - & - & - \\
\bottomrule
\end{tabular}
\caption{\textbf{Detailed per-sequence statistics for the main dataset} (challenge - low light, challenge - moving platform).
}
\label{tbl:supp_recording_stats_part2}
\end{table*}

\begin{figure*}[tb]
\centering
\setlength\tabcolsep{1pt}
\begin{tabular}{cccccccc}
\multicolumn{8}{c}{\textbf{dynamic scenes}} \\ [-7pt]
\includegraphics[height=0.108\linewidth,angle=-90]{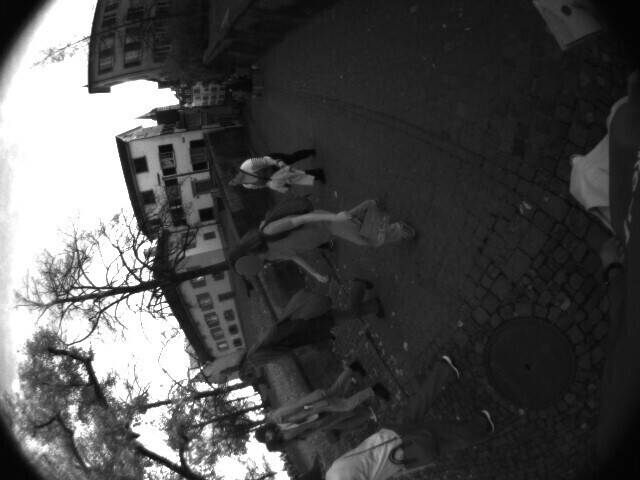} & 
\includegraphics[height=0.108\linewidth,angle=-90]{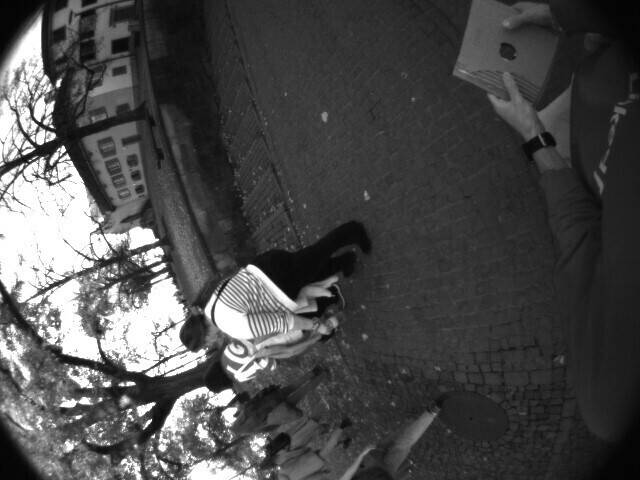} & 
\includegraphics[height=0.108\linewidth,angle=-90]{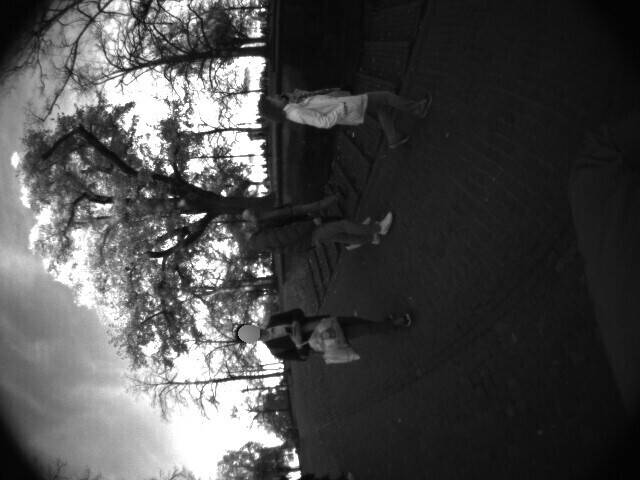} & 
\includegraphics[height=0.108\linewidth,angle=-90]{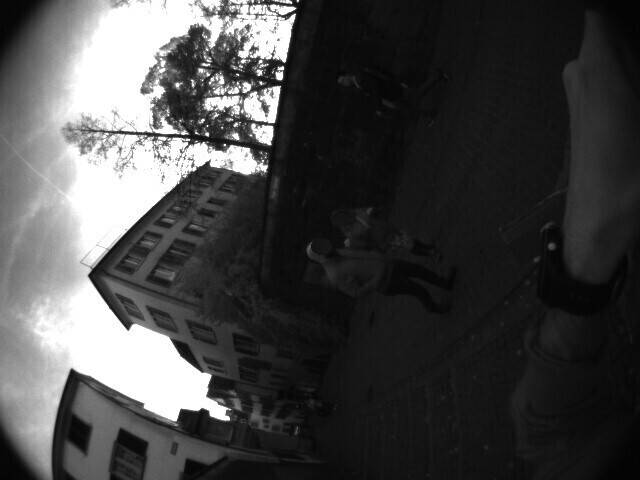} & 
\includegraphics[height=0.108\linewidth,angle=-90]{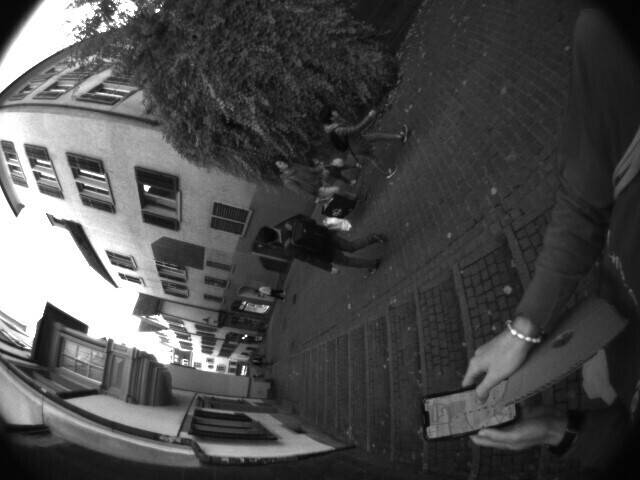} & 
\includegraphics[height=0.108\linewidth,angle=-90]{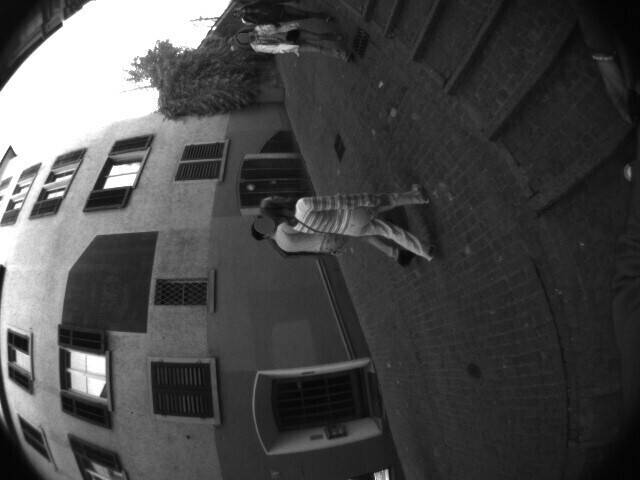} & 
\includegraphics[height=0.108\linewidth,angle=-90]{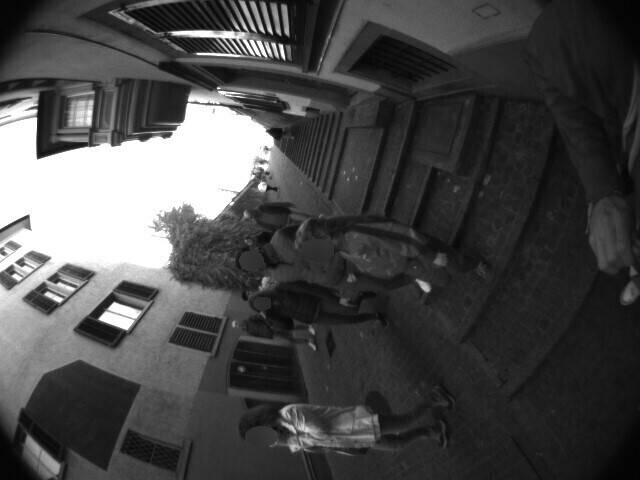} & 
\includegraphics[height=0.108\linewidth,angle=-90]{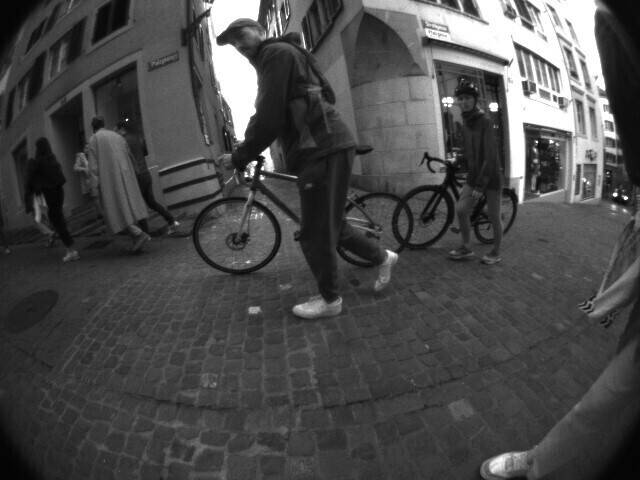} \\ [-5pt]

\includegraphics[height=0.108\linewidth,angle=-90]{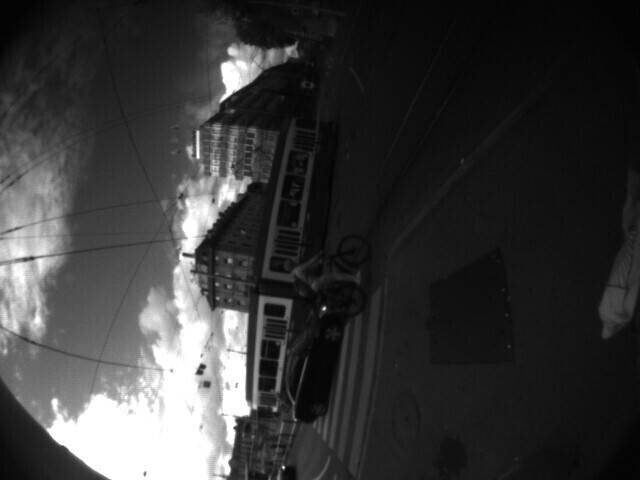} & 
\includegraphics[height=0.108\linewidth,angle=-90]{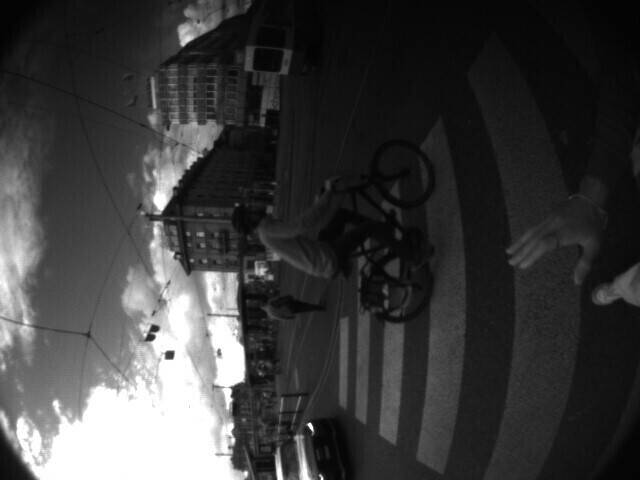} & 
\includegraphics[height=0.108\linewidth,angle=-90]{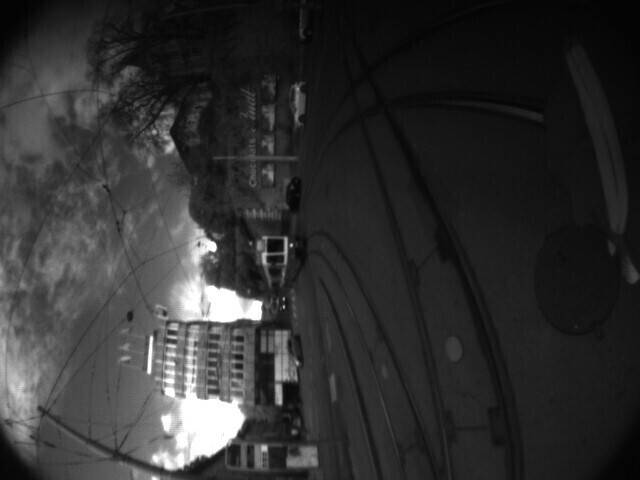} & 
\includegraphics[height=0.108\linewidth,angle=-90]{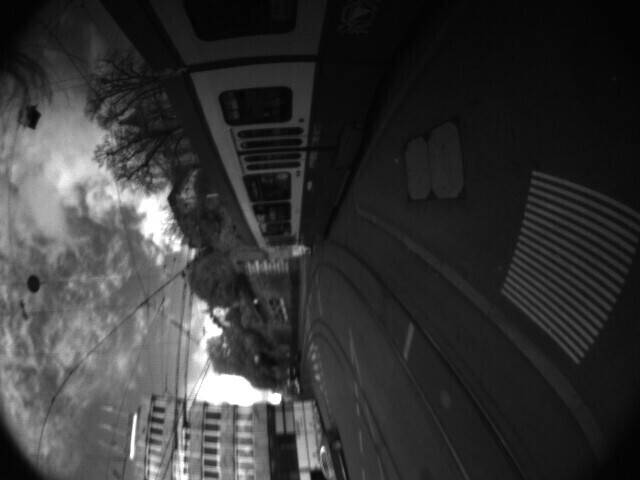} & 
\includegraphics[height=0.108\linewidth,angle=-90]{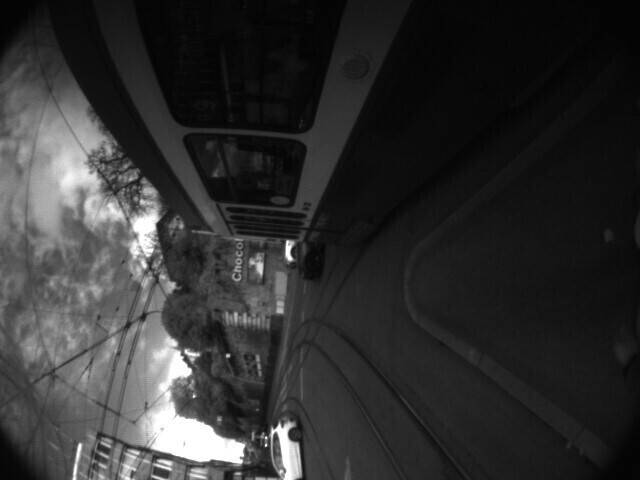} & 
\includegraphics[height=0.108\linewidth,angle=-90]{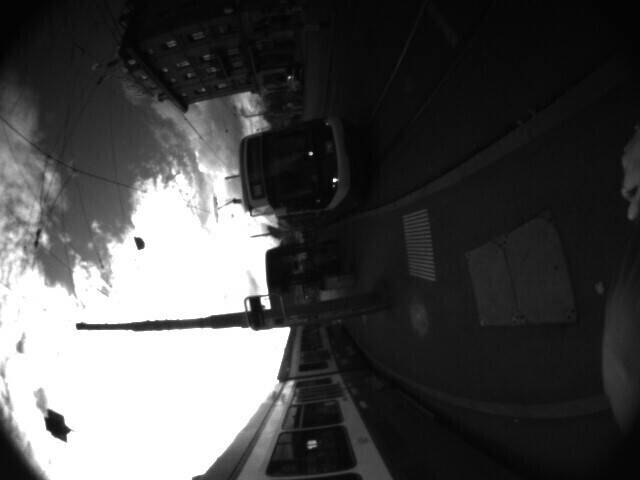} & 
\includegraphics[height=0.108\linewidth,angle=-90]{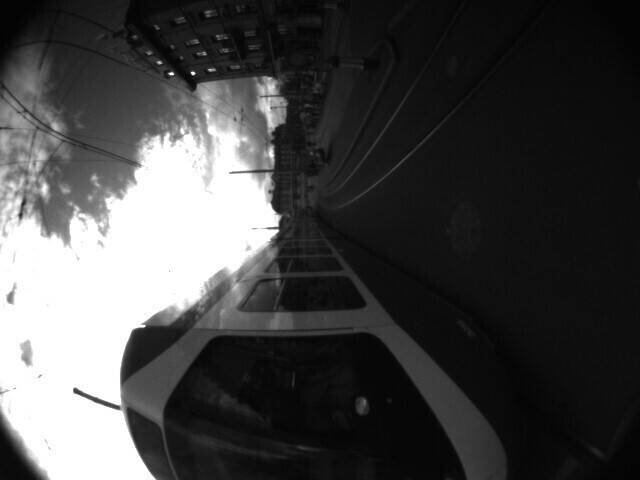} & 
\includegraphics[height=0.108\linewidth,angle=-90]{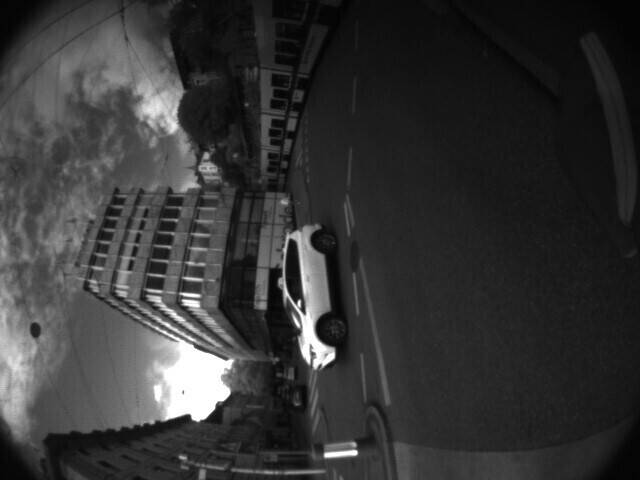} \\ [-5pt]

\includegraphics[height=0.108\linewidth,angle=-90]{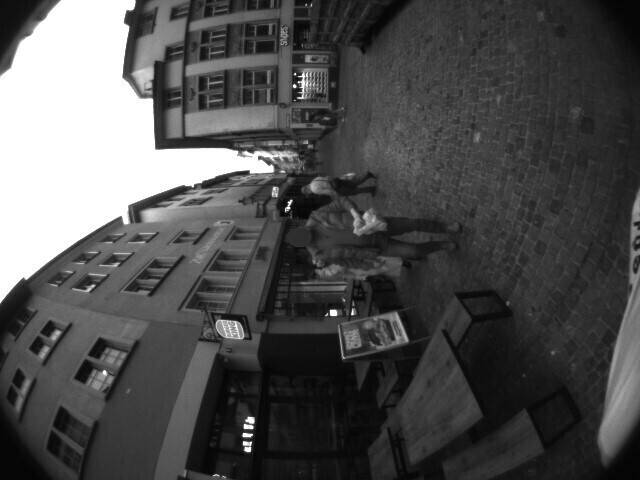} & 
\includegraphics[height=0.108\linewidth,angle=-90]{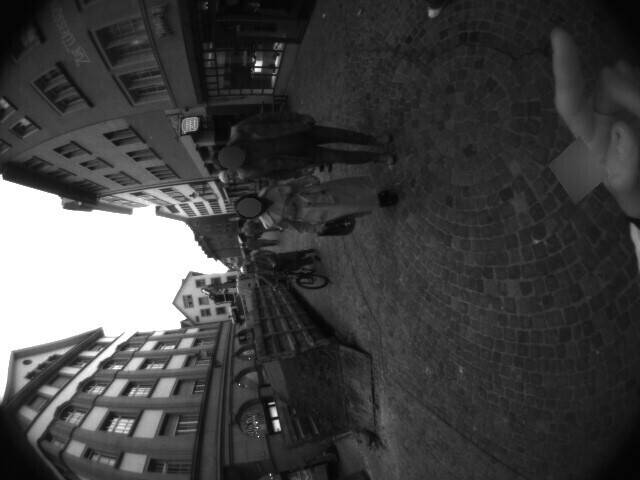} & 
\includegraphics[height=0.108\linewidth,angle=-90]{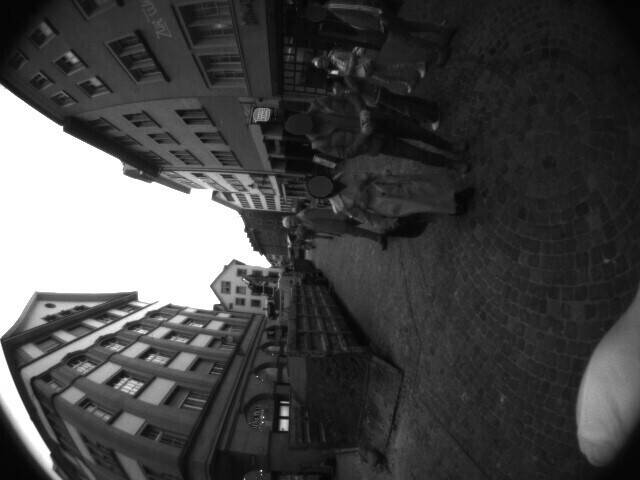} & 
\includegraphics[height=0.108\linewidth,angle=-90]{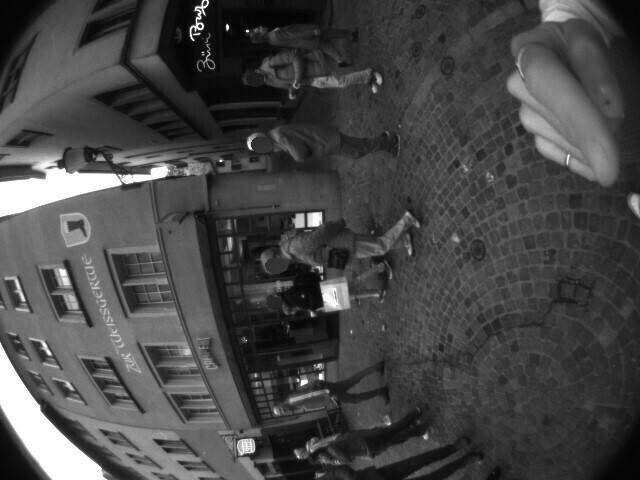} & 
\includegraphics[height=0.108\linewidth,angle=-90]{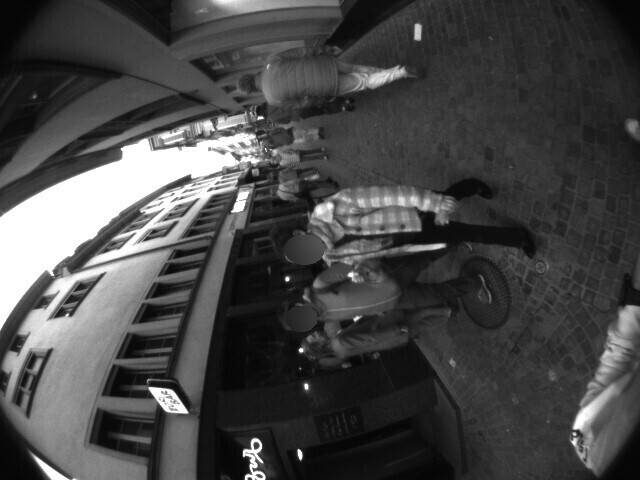} & 
\includegraphics[height=0.108\linewidth,angle=-90]{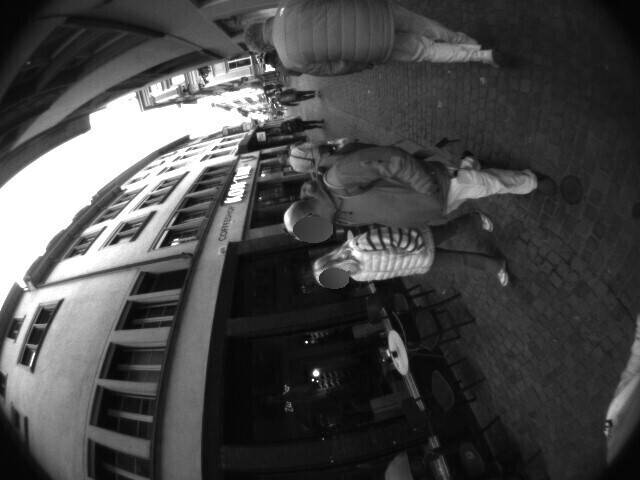} & 
\includegraphics[height=0.108\linewidth,angle=-90]{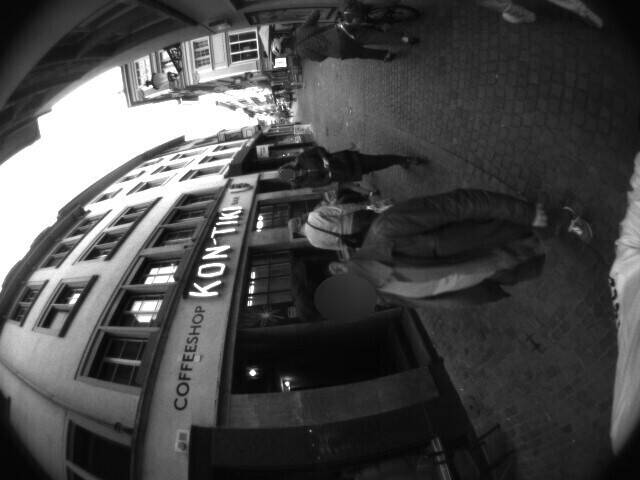} & 
\includegraphics[height=0.108\linewidth,angle=-90]{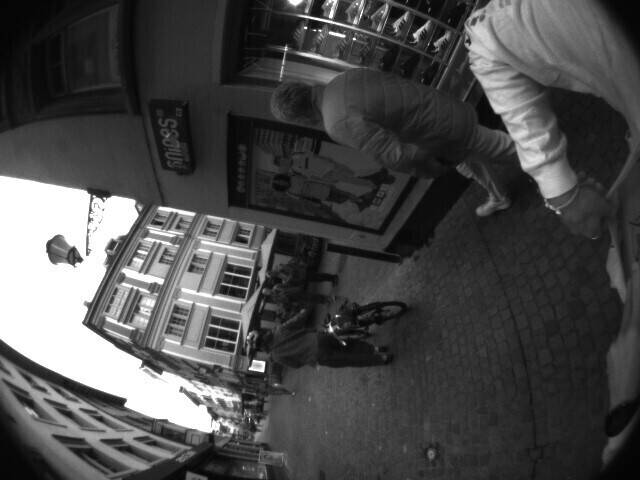} \\ [-5pt]

\includegraphics[height=0.108\linewidth,angle=-90]{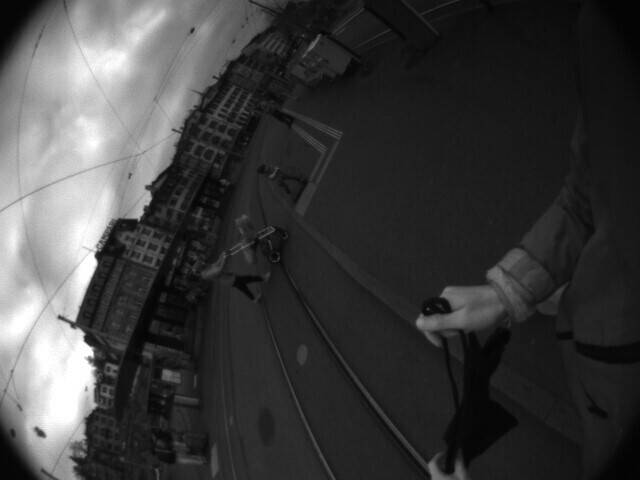} & 
\includegraphics[height=0.108\linewidth,angle=-90]{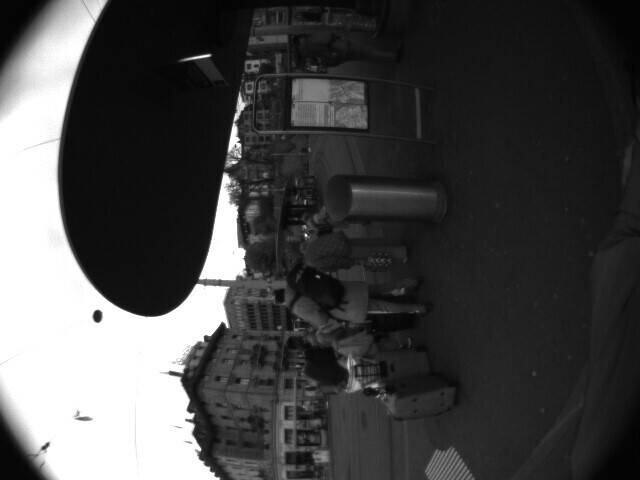} & 
\includegraphics[height=0.108\linewidth,angle=-90]{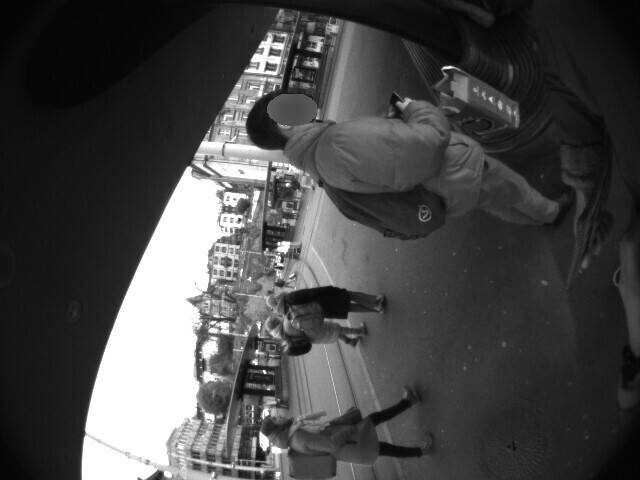} & 
\includegraphics[height=0.108\linewidth,angle=-90]{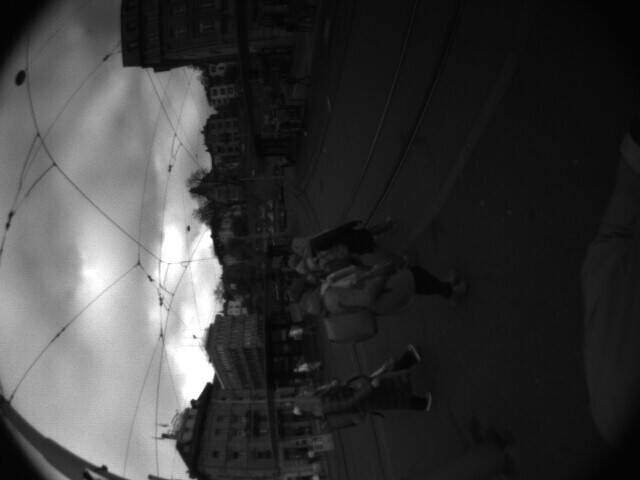} & 
\includegraphics[height=0.108\linewidth,angle=-90]{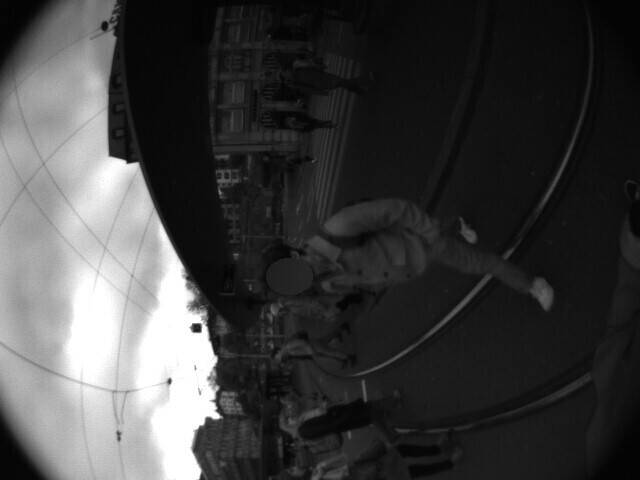} & 
\includegraphics[height=0.108\linewidth,angle=-90]{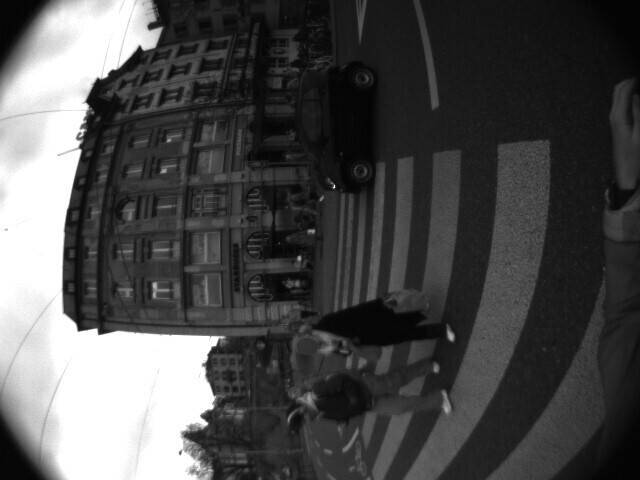} & 
\includegraphics[height=0.108\linewidth,angle=-90]{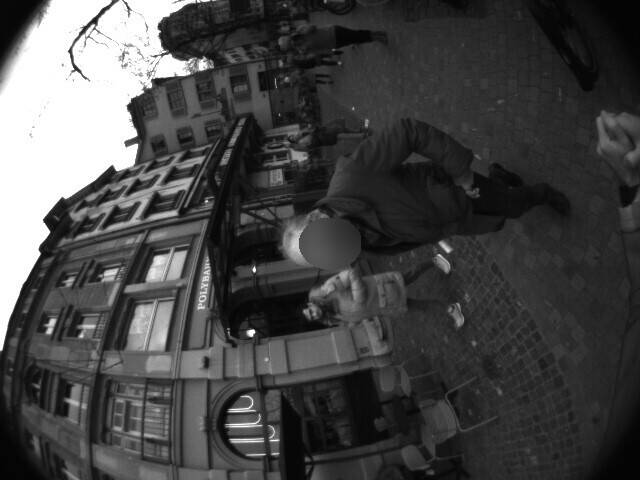} & 
\includegraphics[height=0.108\linewidth,angle=-90]{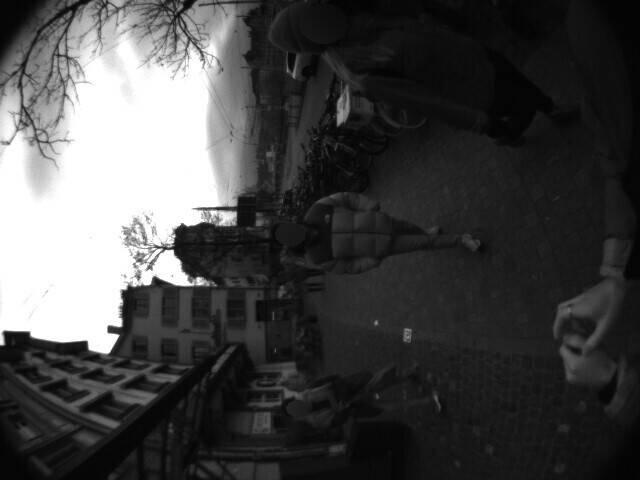} \\
\\
\multicolumn{8}{c}{\textbf{exposure changes}} \\ [-7pt]
\includegraphics[height=0.108\linewidth,angle=-90]{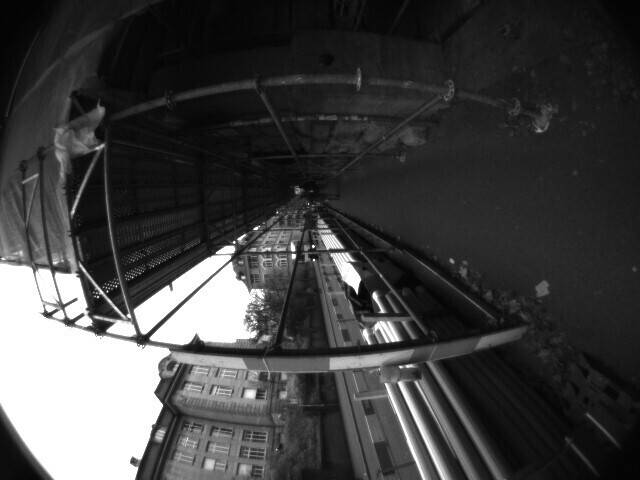} & 
\includegraphics[height=0.108\linewidth,angle=-90]{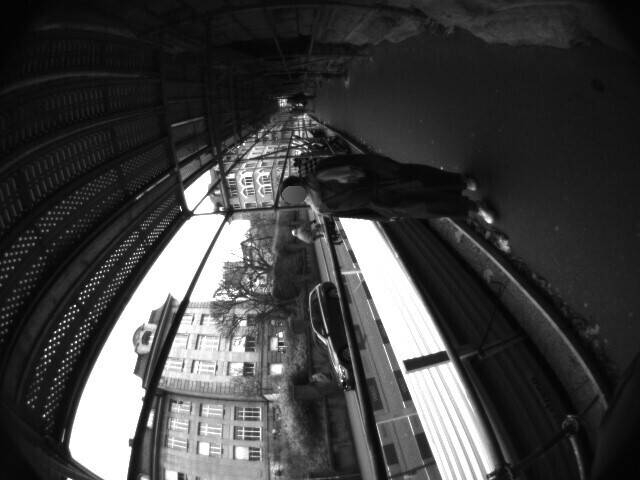} & 
\includegraphics[height=0.108\linewidth,angle=-90]{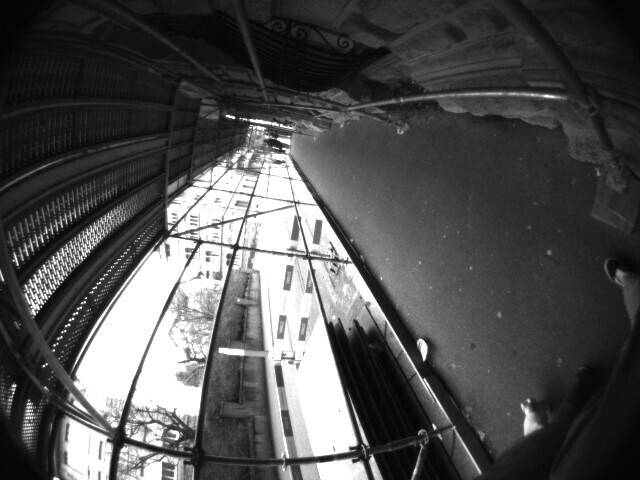} & 
\includegraphics[height=0.108\linewidth,angle=-90]{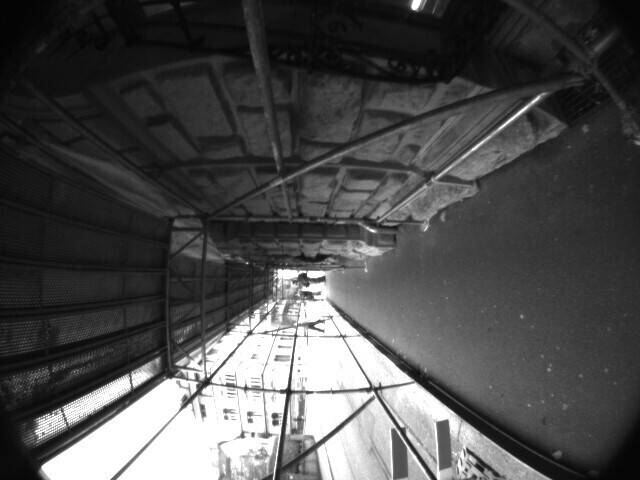} & 
\includegraphics[height=0.108\linewidth,angle=-90]{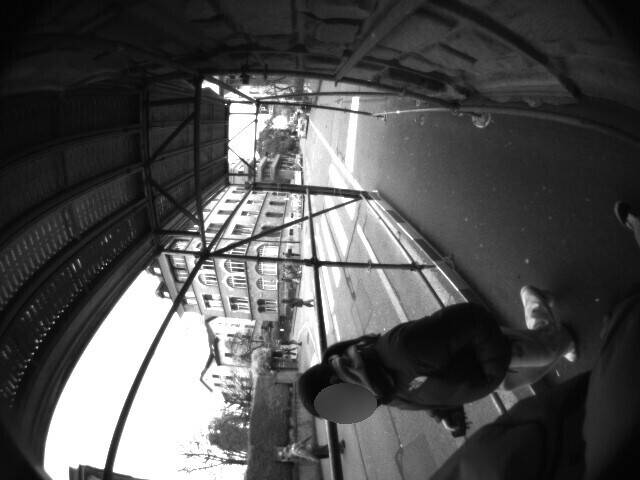} & 
\includegraphics[height=0.108\linewidth,angle=-90]{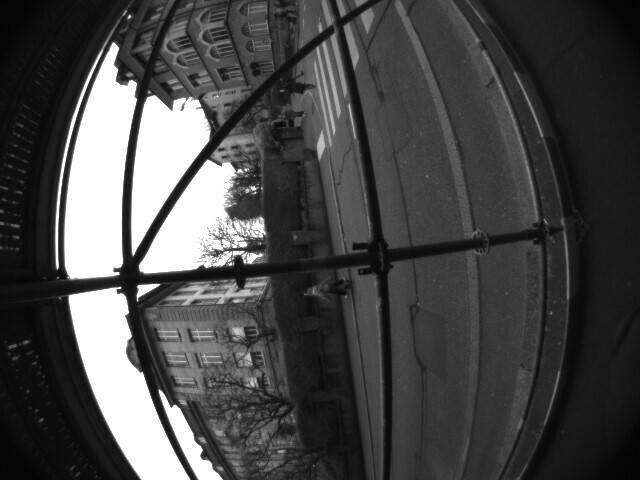} &
\includegraphics[height=0.108\linewidth,angle=-90]{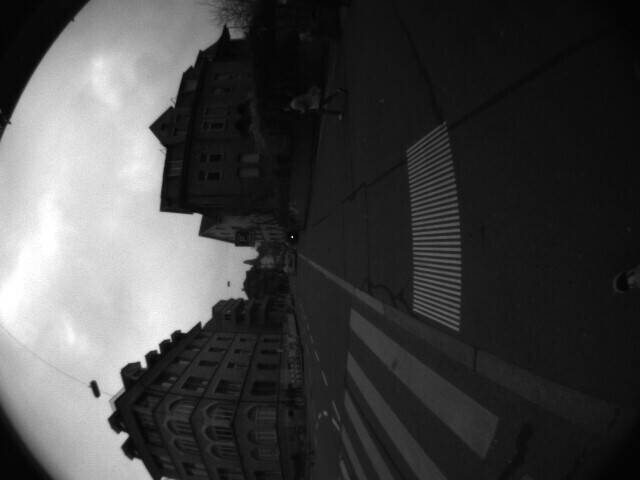} & 
\includegraphics[height=0.108\linewidth,angle=-90]{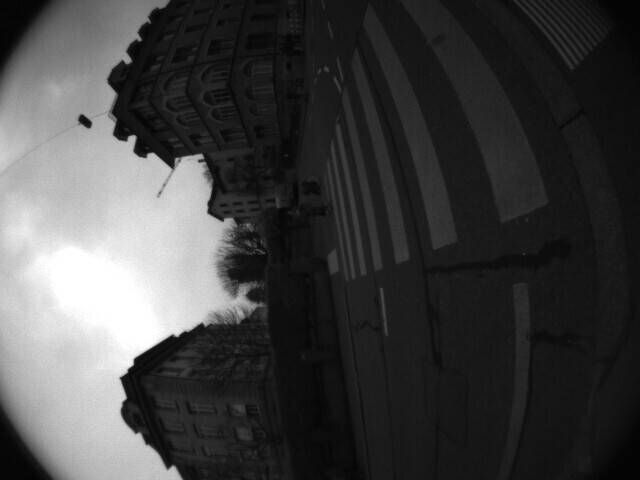} \\ [-5pt] 

\includegraphics[height=0.108\linewidth,angle=-90]{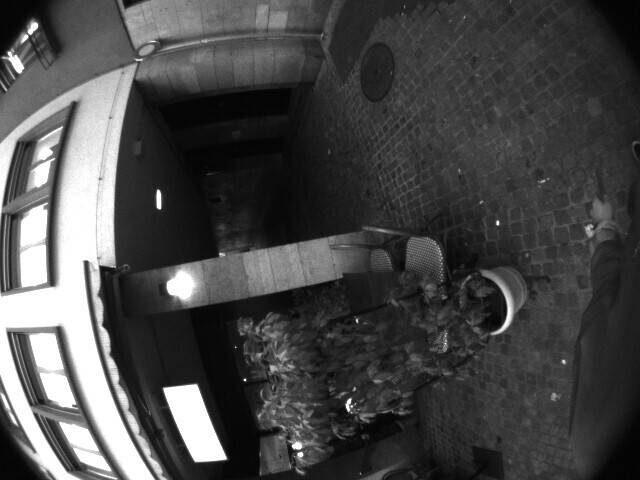} & 
\includegraphics[height=0.108\linewidth,angle=-90]{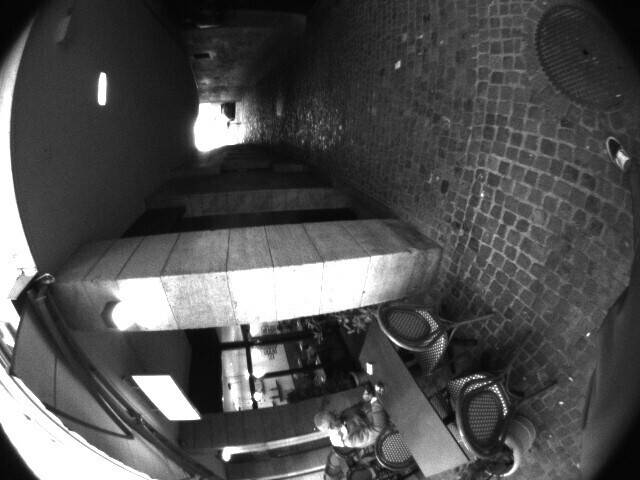} & 
\includegraphics[height=0.108\linewidth,angle=-90]{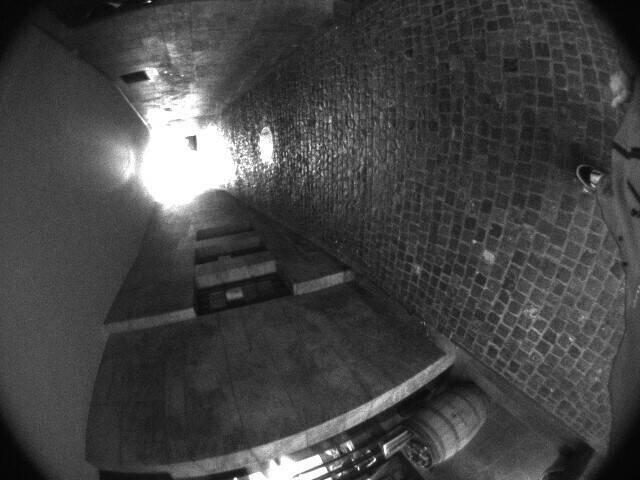} & 
\includegraphics[height=0.108\linewidth,angle=-90]{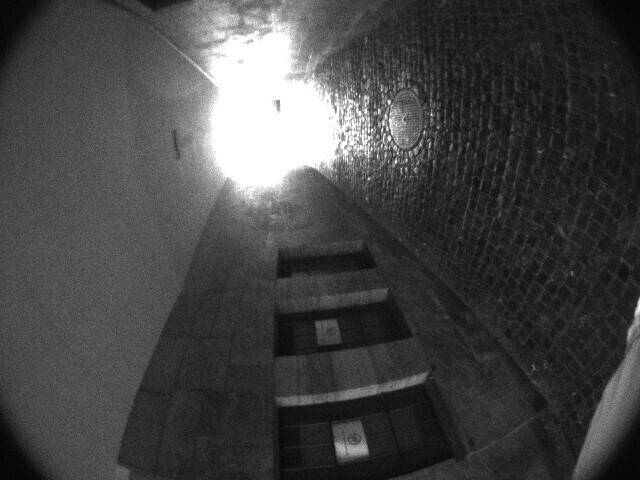} & 
\includegraphics[height=0.108\linewidth,angle=-90]{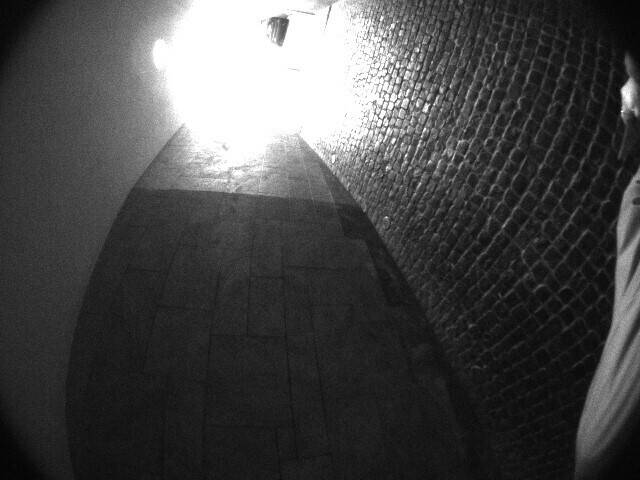} & 
\includegraphics[height=0.108\linewidth,angle=-90]{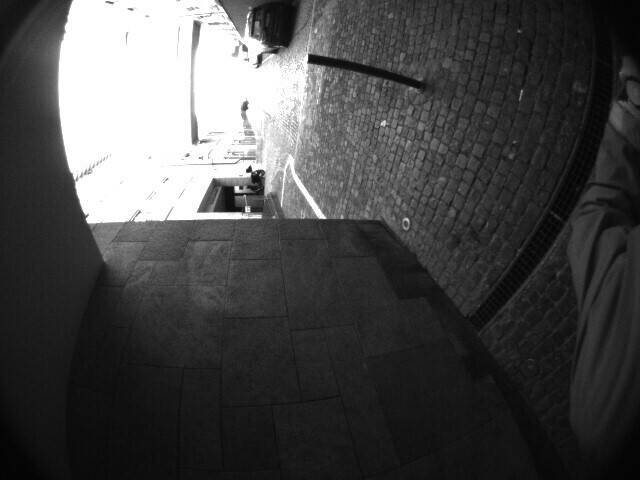} & 
\includegraphics[height=0.108\linewidth,angle=-90]{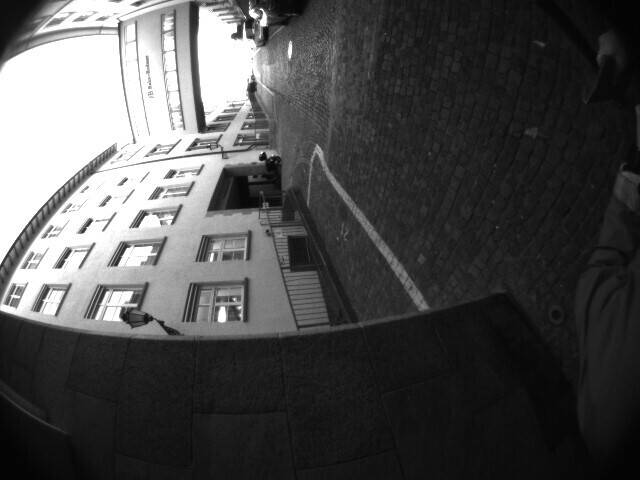} & 
\includegraphics[height=0.108\linewidth,angle=-90]{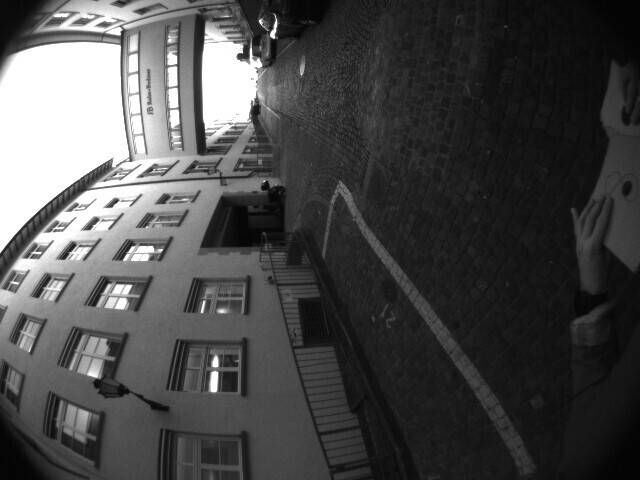} \\ [-5pt]

\includegraphics[height=0.108\linewidth,angle=-90]{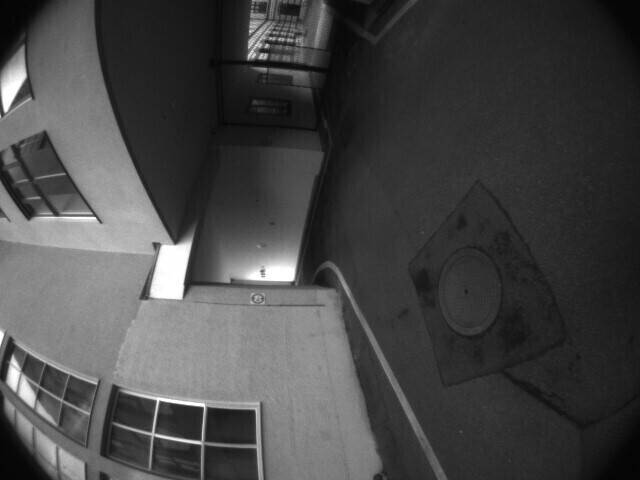} & 
\includegraphics[height=0.108\linewidth,angle=-90]{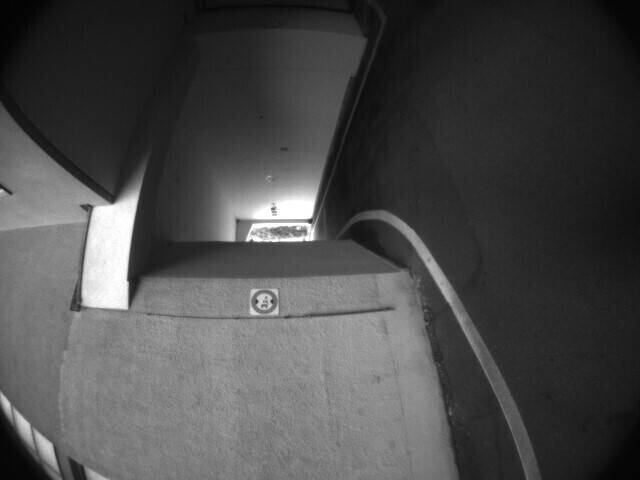} & 
\includegraphics[height=0.108\linewidth,angle=-90]{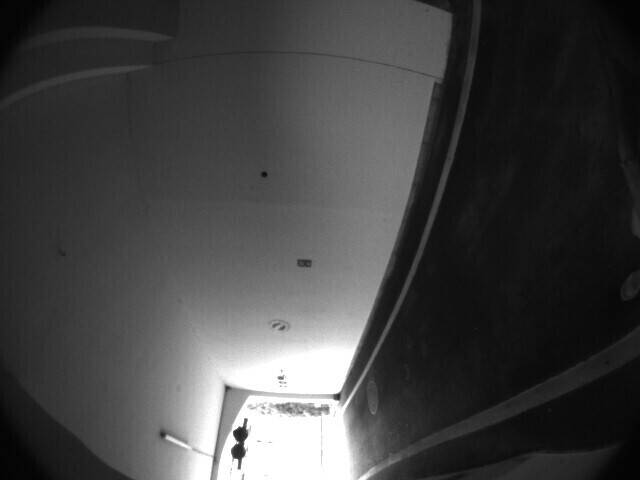} & 
\includegraphics[height=0.108\linewidth,angle=-90]{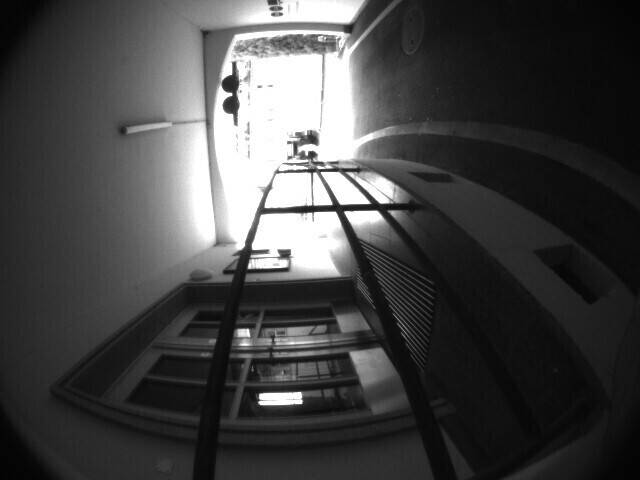} & 
\includegraphics[height=0.108\linewidth,angle=-90]{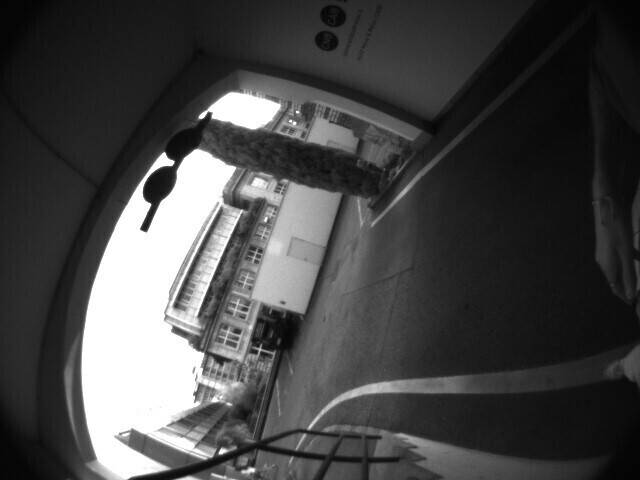} & 
\includegraphics[height=0.108\linewidth,angle=-90]{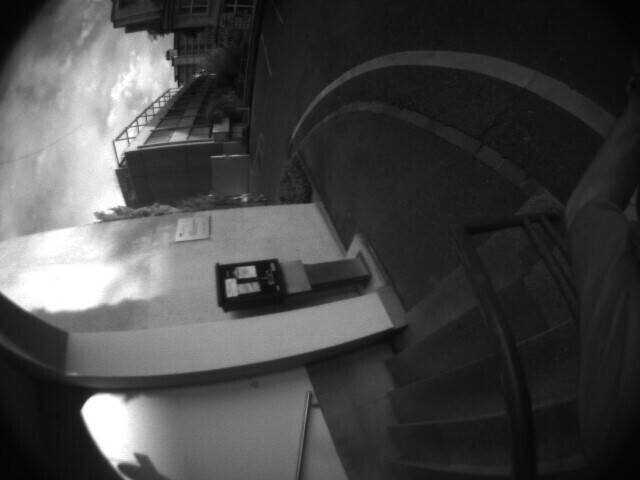} & 
\includegraphics[height=0.108\linewidth,angle=-90]{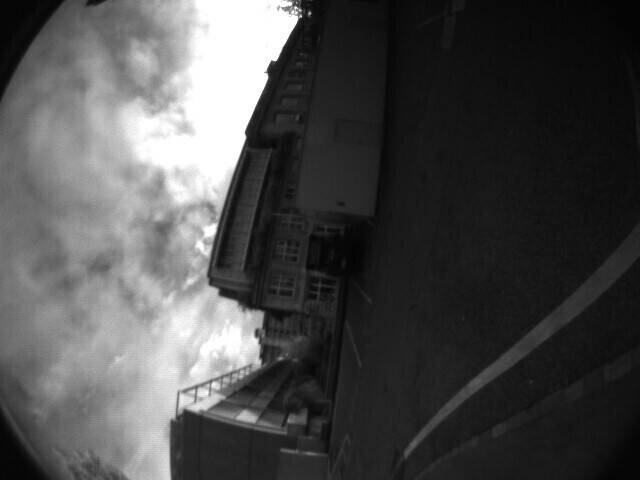} & 
\includegraphics[height=0.108\linewidth,angle=-90]{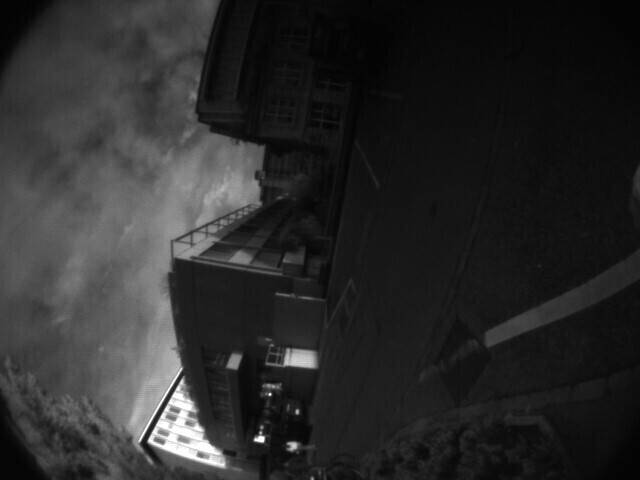} \\ [-5pt]

\includegraphics[height=0.108\linewidth,angle=-90]{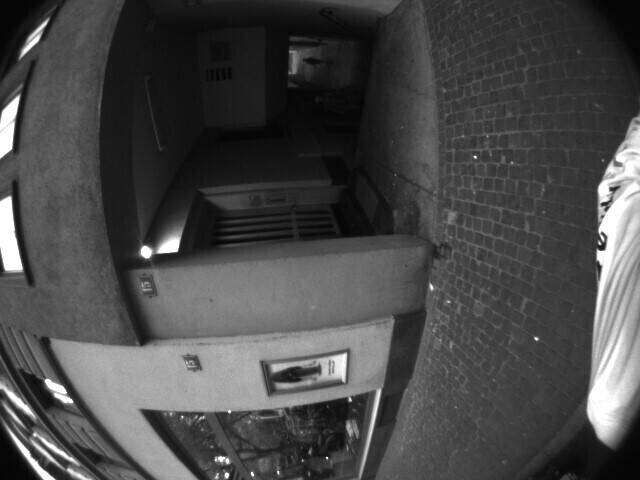} & 
\includegraphics[height=0.108\linewidth,angle=-90]{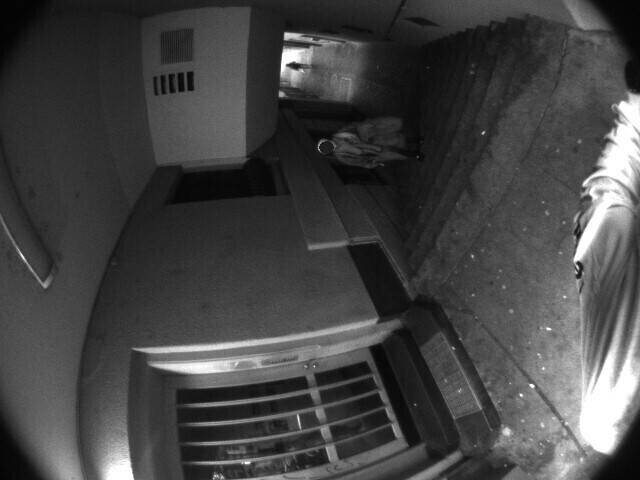} & 
\includegraphics[height=0.108\linewidth,angle=-90]{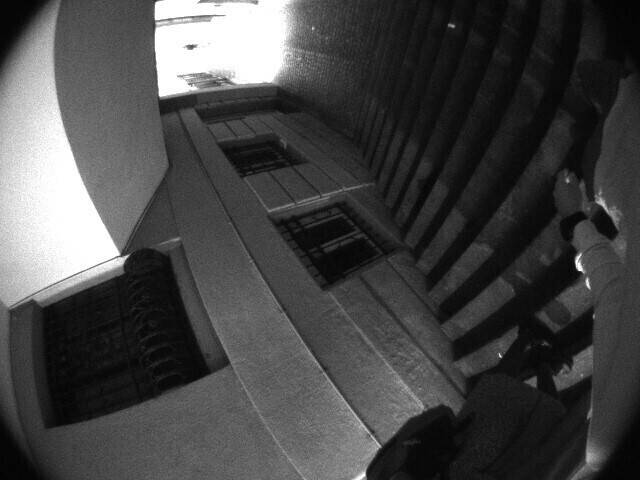} & 
\includegraphics[height=0.108\linewidth,angle=-90]{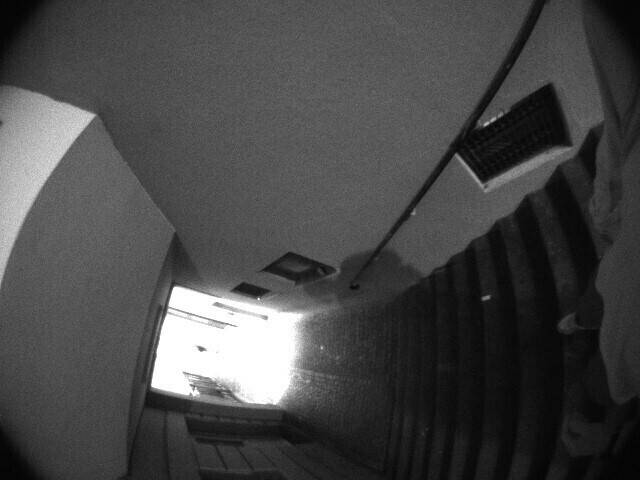} & 
\includegraphics[height=0.108\linewidth,angle=-90]{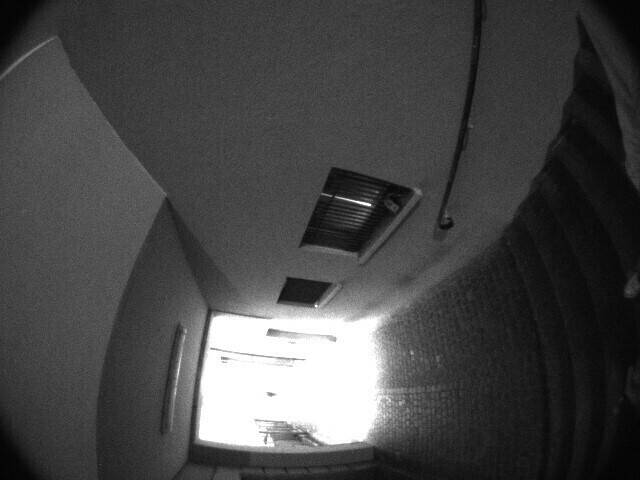} & 
\includegraphics[height=0.108\linewidth,angle=-90]{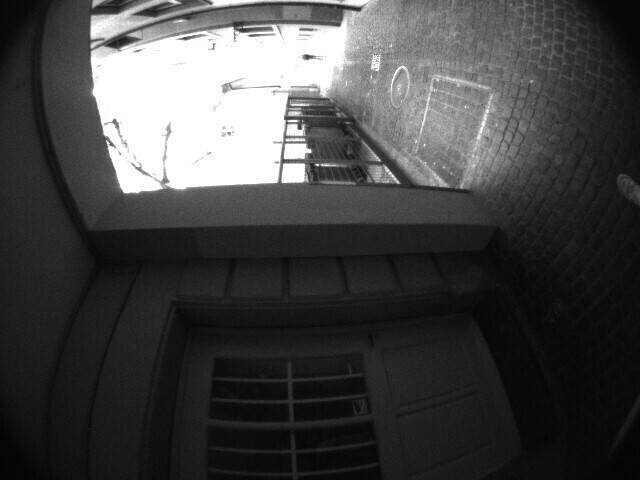} & 
\includegraphics[height=0.108\linewidth,angle=-90]{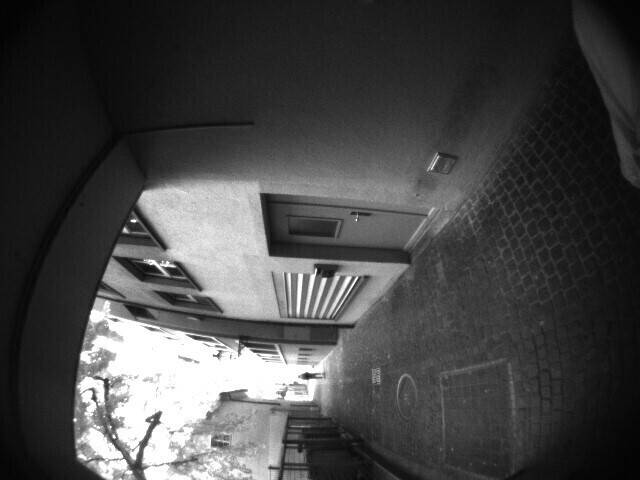} & 
\includegraphics[height=0.108\linewidth,angle=-90]{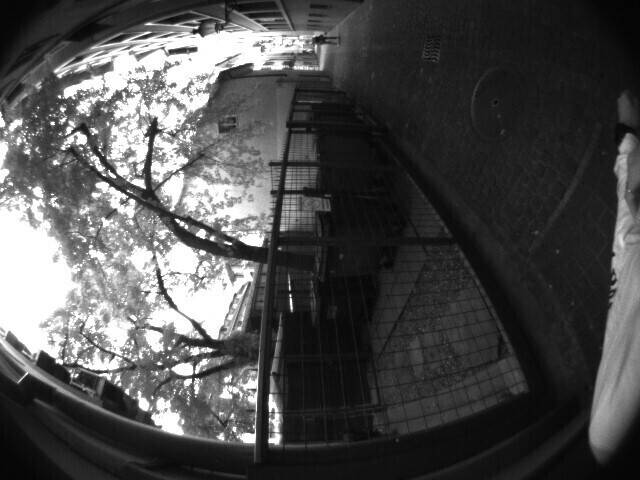} \\
\end{tabular}
\centering
\caption{\textbf{Visualizations of the egocentric recordings in our dataset.}}
\label{fig:supp_examples_full_set_1}%
\end{figure*}

\begin{figure*}[tb]
\centering
\setlength\tabcolsep{1pt}
\begin{tabular}{cccccccc}
\multicolumn{8}{c}{\textbf{low light}} \\ [-5pt]
\includegraphics[height=0.108\linewidth,angle=-90]{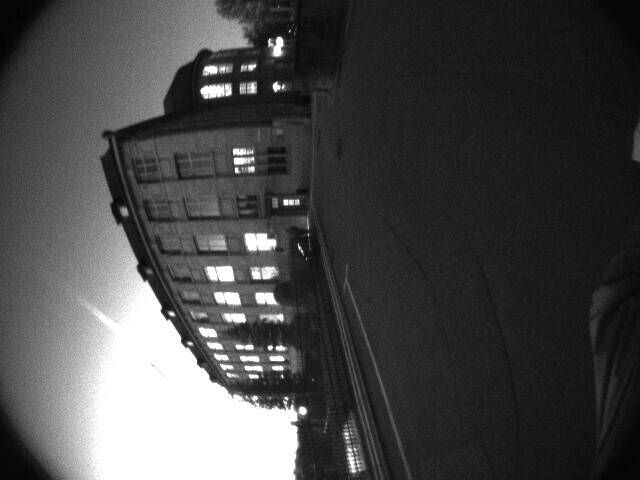} & 
\includegraphics[height=0.108\linewidth,angle=-90]{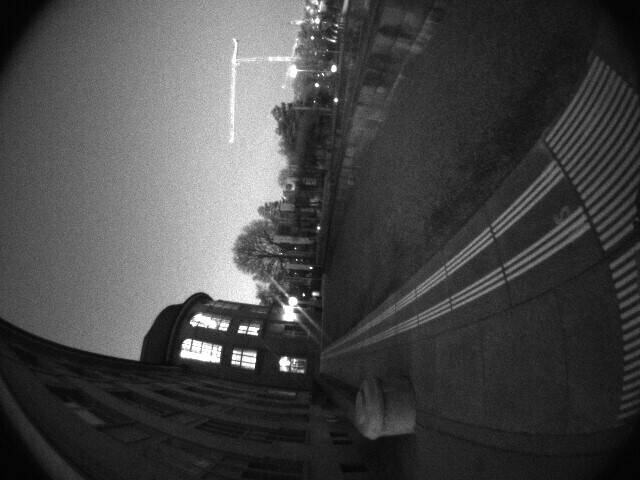} & 
\includegraphics[height=0.108\linewidth,angle=-90]{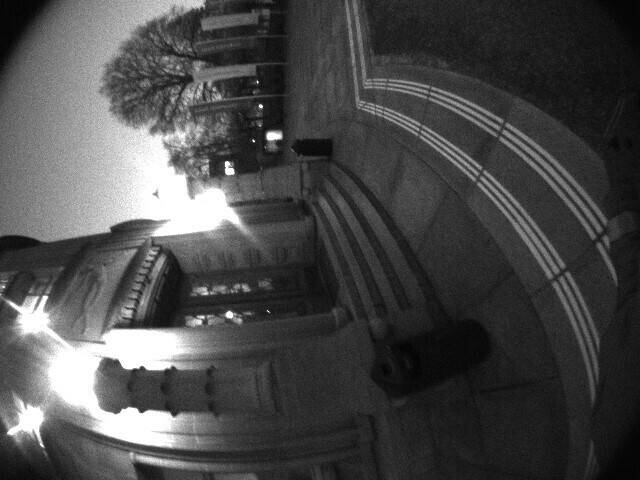} & 
\includegraphics[height=0.108\linewidth,angle=-90]{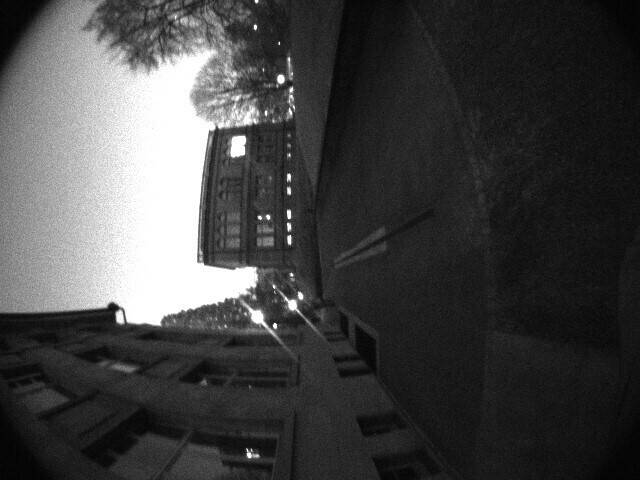} & 
\includegraphics[height=0.108\linewidth,angle=-90]{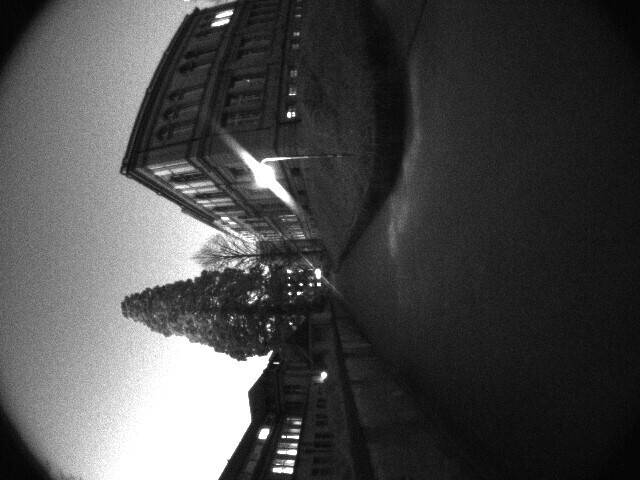} & 
\includegraphics[height=0.108\linewidth,angle=-90]{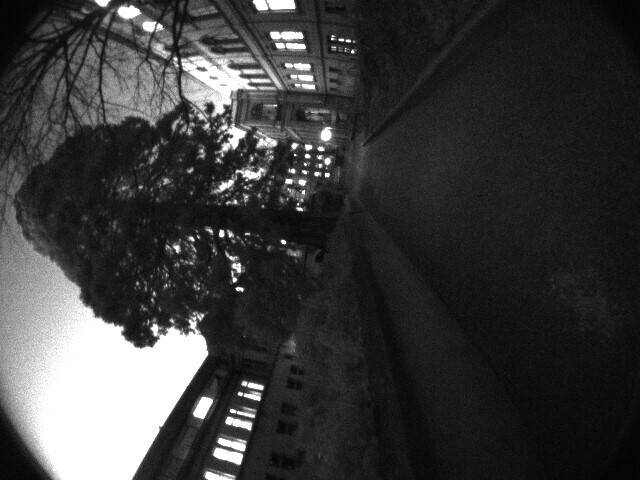} & 
\includegraphics[height=0.108\linewidth,angle=-90]{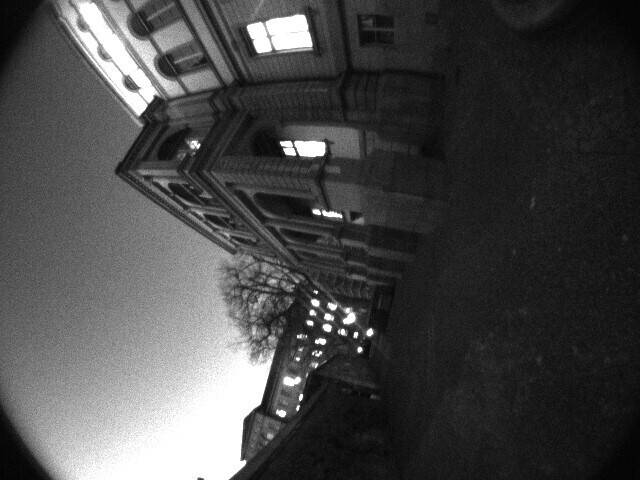} & 
\includegraphics[height=0.108\linewidth,angle=-90]{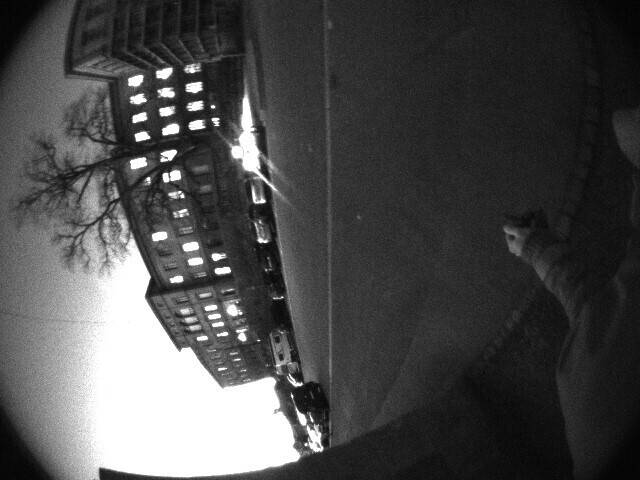} \\ [-5pt]

\includegraphics[height=0.108\linewidth,angle=-90]{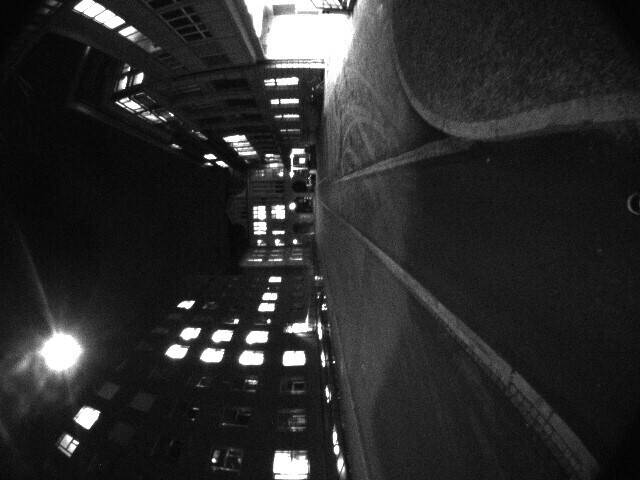} & 
\includegraphics[height=0.108\linewidth,angle=-90]{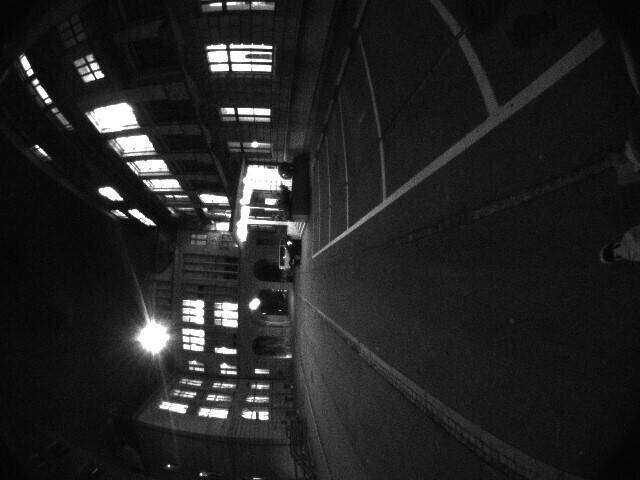} & 
\includegraphics[height=0.108\linewidth,angle=-90]{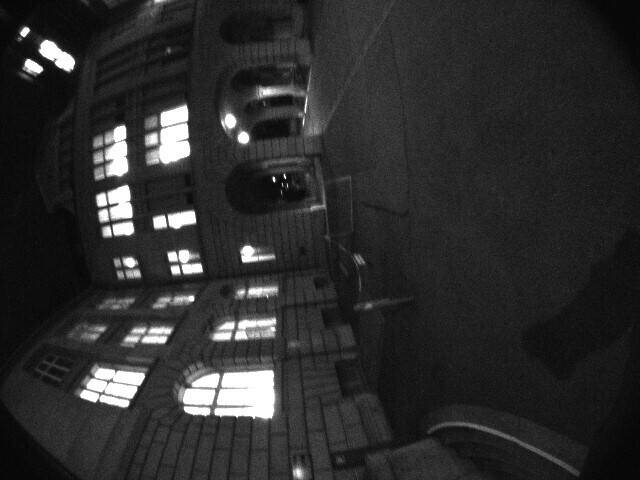} & 
\includegraphics[height=0.108\linewidth,angle=-90]{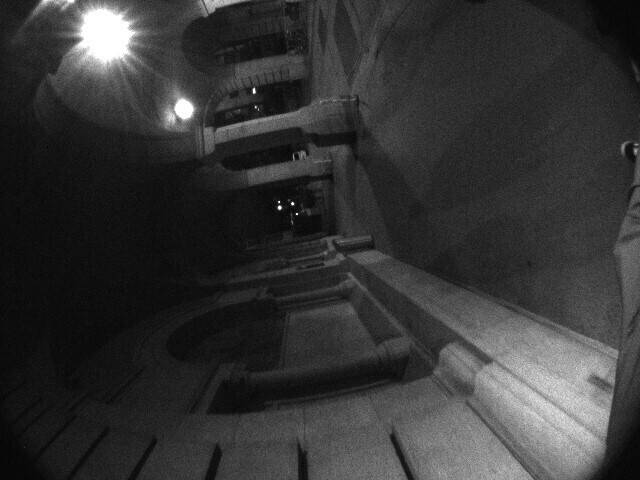} & 
\includegraphics[height=0.108\linewidth,angle=-90]{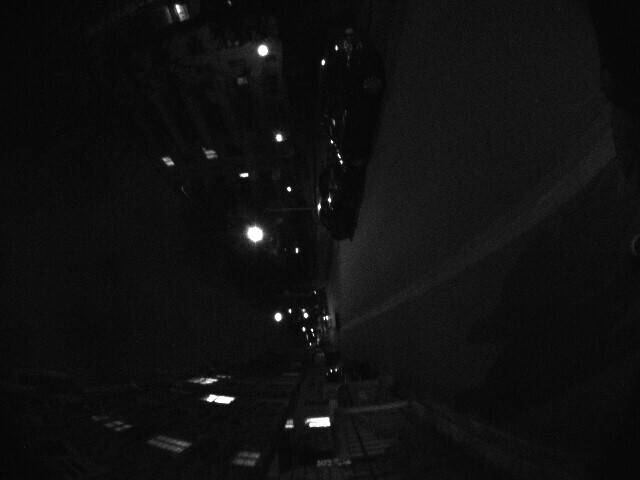} & 
\includegraphics[height=0.108\linewidth,angle=-90]{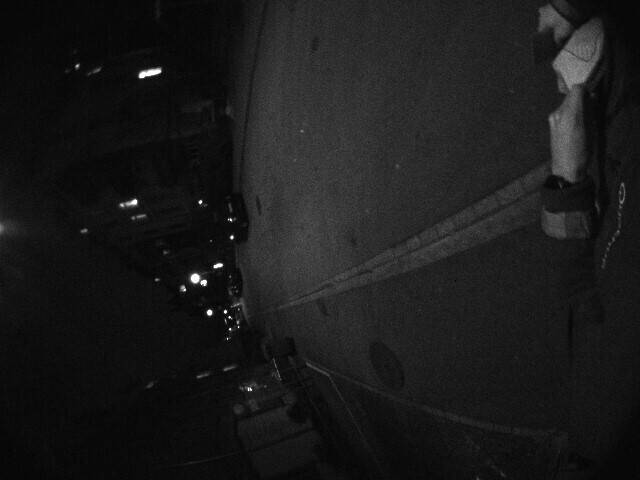} & 
\includegraphics[height=0.108\linewidth,angle=-90]{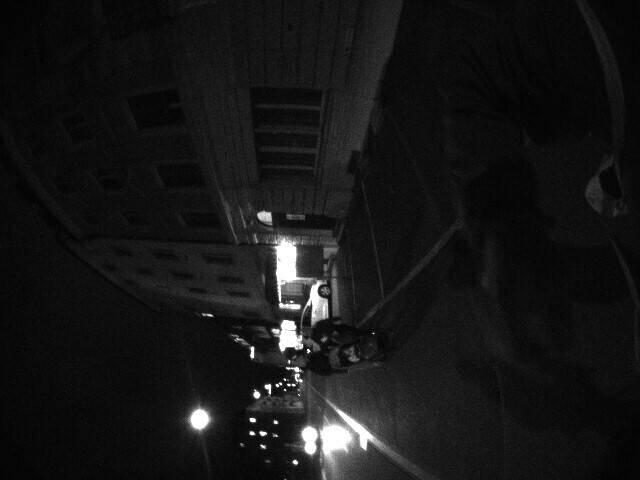} & 
\includegraphics[height=0.108\linewidth,angle=-90]{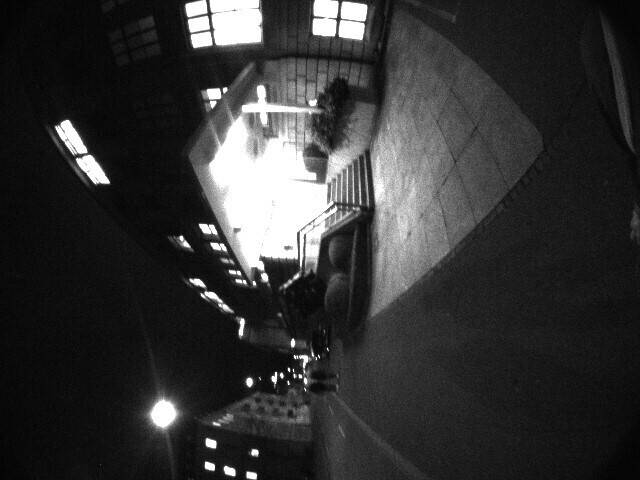} \\ [-5pt]

\includegraphics[height=0.108\linewidth,angle=-90]{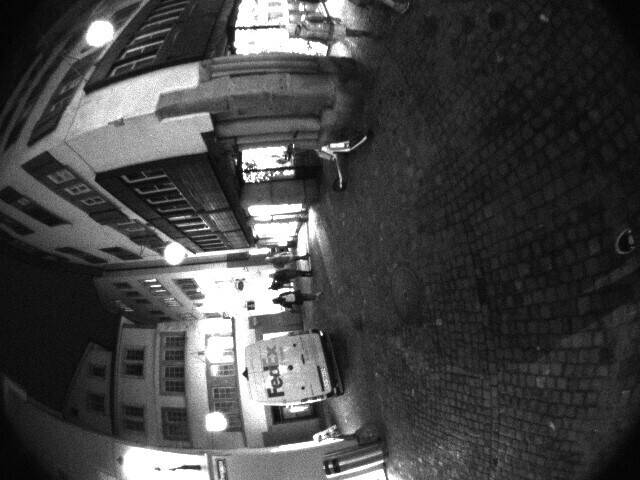} & 
\includegraphics[height=0.108\linewidth,angle=-90]{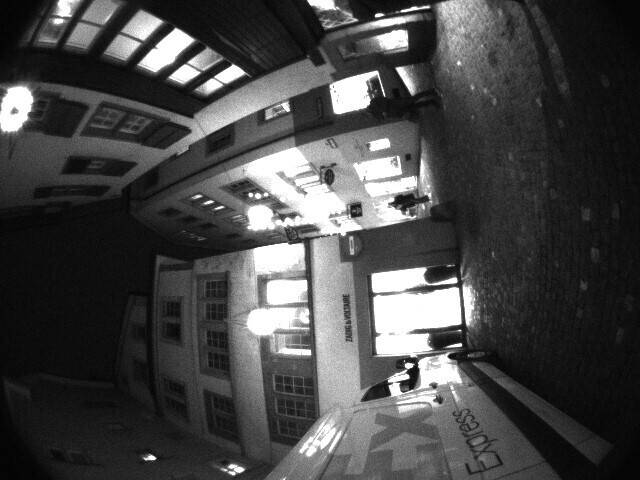} & 
\includegraphics[height=0.108\linewidth,angle=-90]{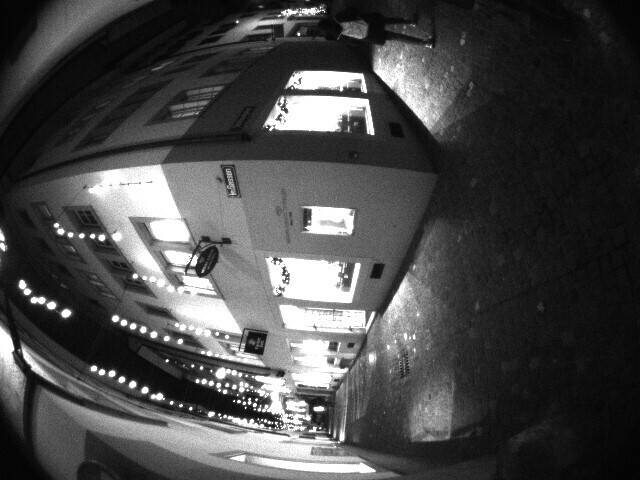} & 
\includegraphics[height=0.108\linewidth,angle=-90]{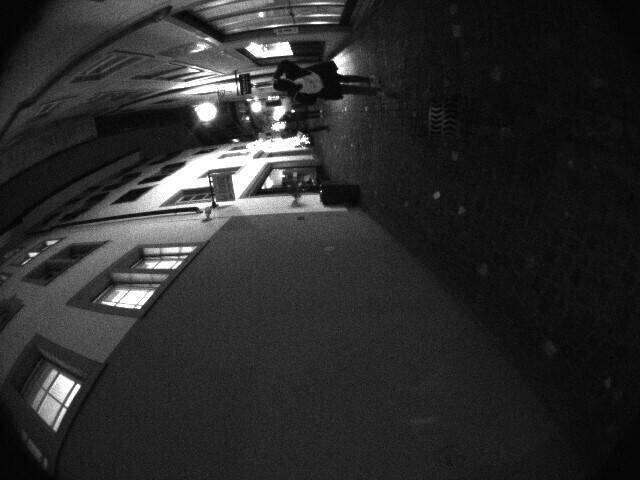} & 
\includegraphics[height=0.108\linewidth,angle=-90]{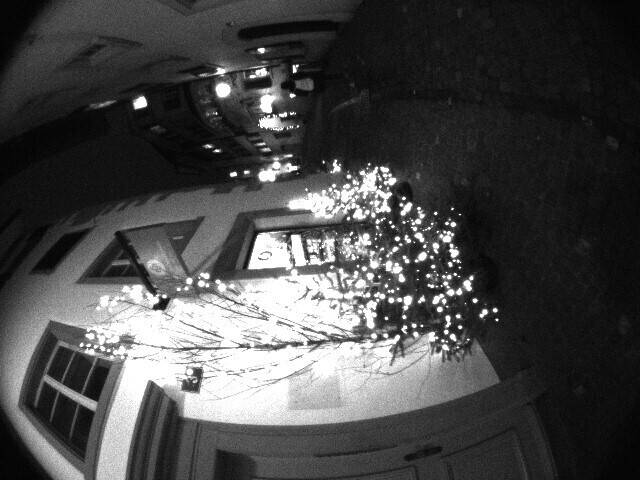} & 
\includegraphics[height=0.108\linewidth,angle=-90]{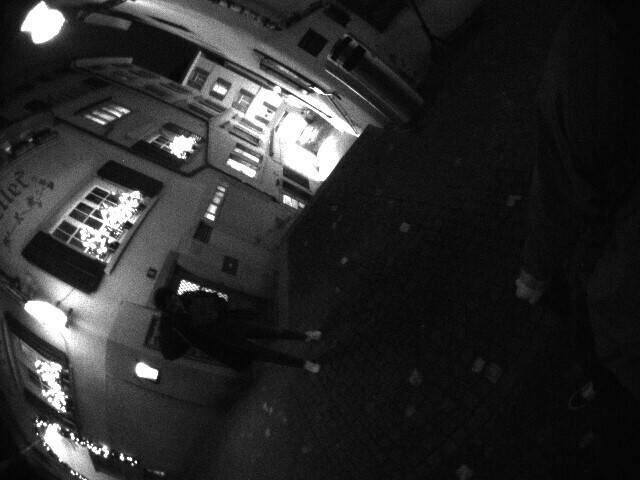} & 
\includegraphics[height=0.108\linewidth,angle=-90]{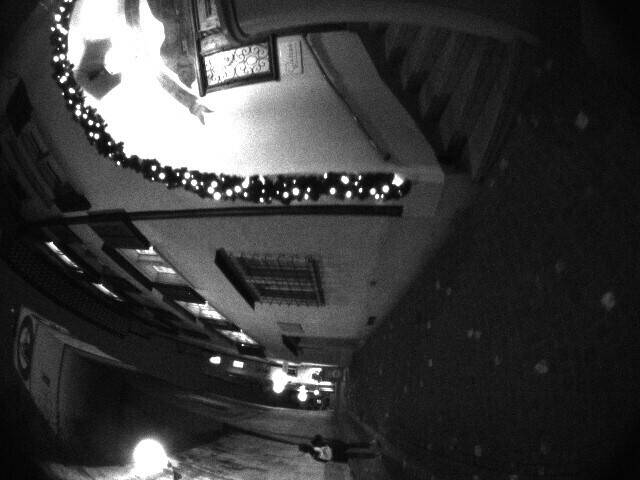} & 
\includegraphics[height=0.108\linewidth,angle=-90]{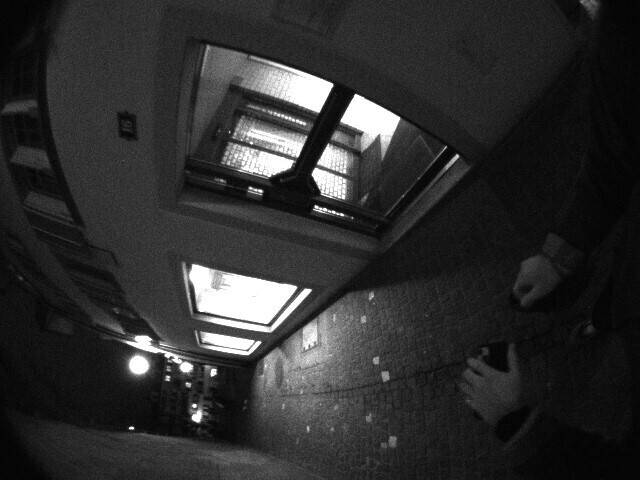} \\ [-5pt]

\includegraphics[height=0.108\linewidth,angle=-90]{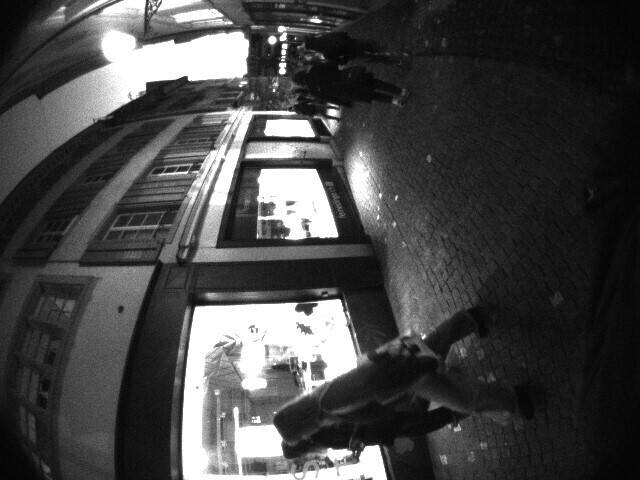} & 
\includegraphics[height=0.108\linewidth,angle=-90]{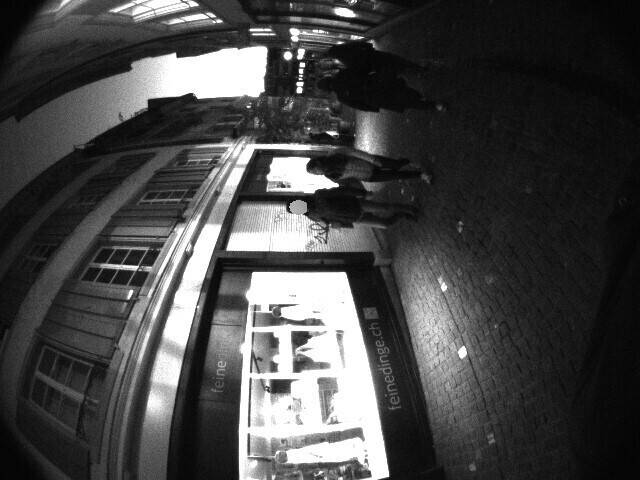} & 
\includegraphics[height=0.108\linewidth,angle=-90]{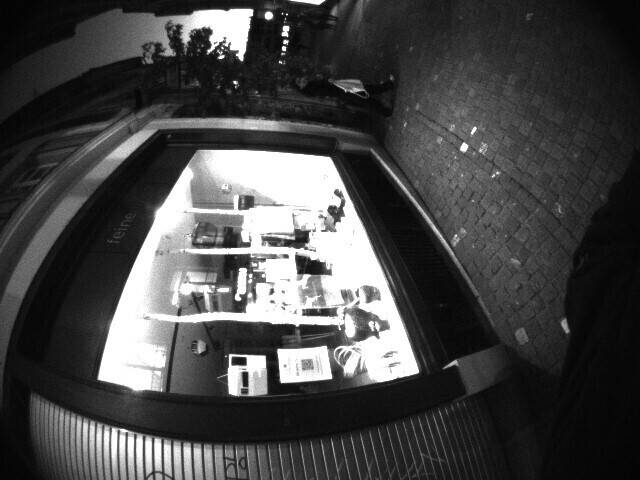} & 
\includegraphics[height=0.108\linewidth,angle=-90]{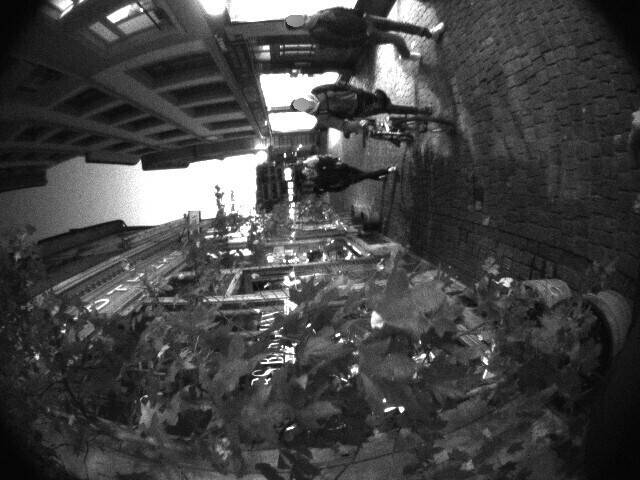} & 
\includegraphics[height=0.108\linewidth,angle=-90]{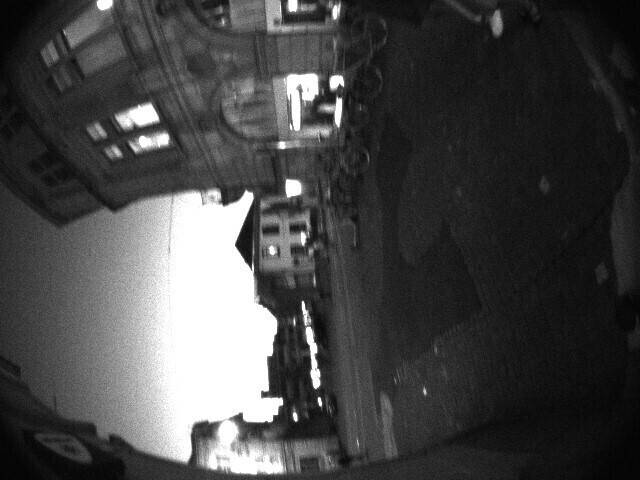} & 
\includegraphics[height=0.108\linewidth,angle=-90]{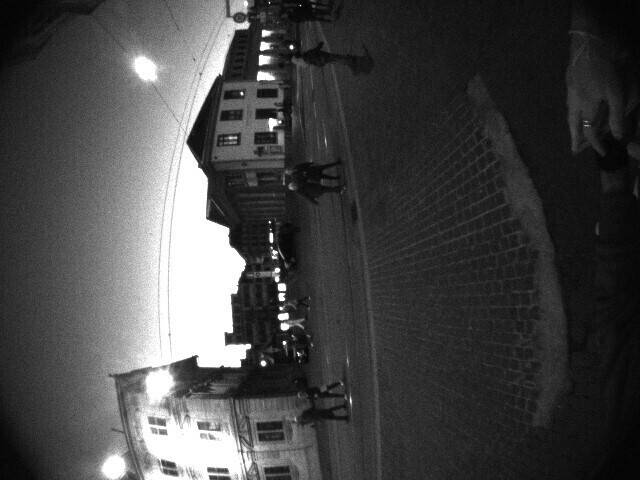} & 
\includegraphics[height=0.108\linewidth,angle=-90]{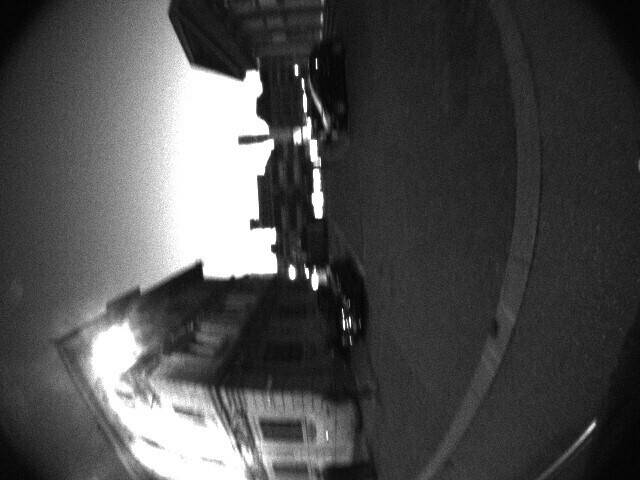} & 
\includegraphics[height=0.108\linewidth,angle=-90]{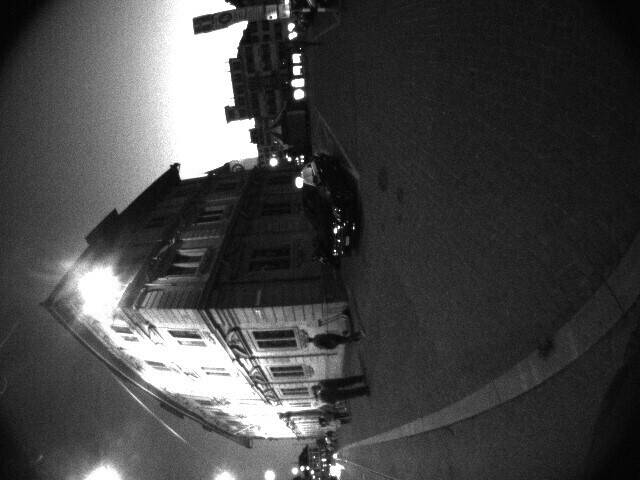} \\
\\
\multicolumn{8}{c}{\textbf{moving platforms}} \\ [-5pt]
\includegraphics[height=0.108\linewidth,angle=-90]{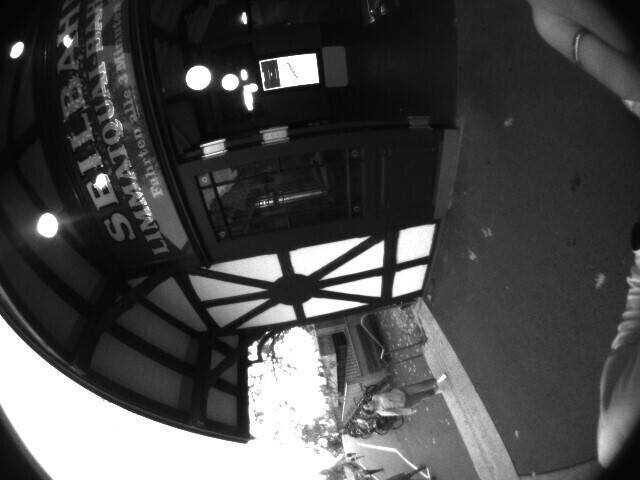} & 
\includegraphics[height=0.108\linewidth,angle=-90]{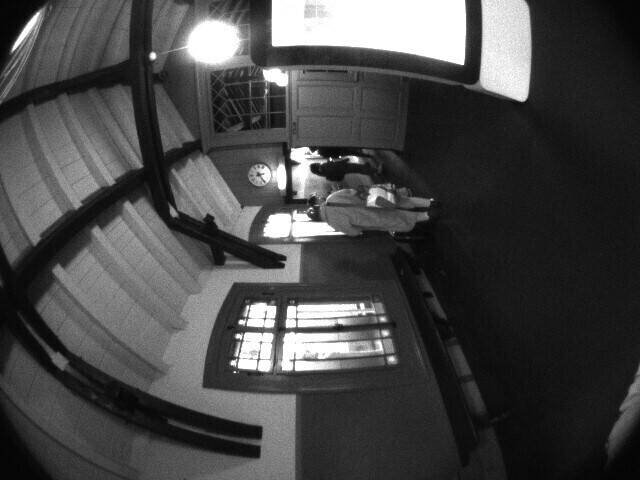} & 
\includegraphics[height=0.108\linewidth,angle=-90]{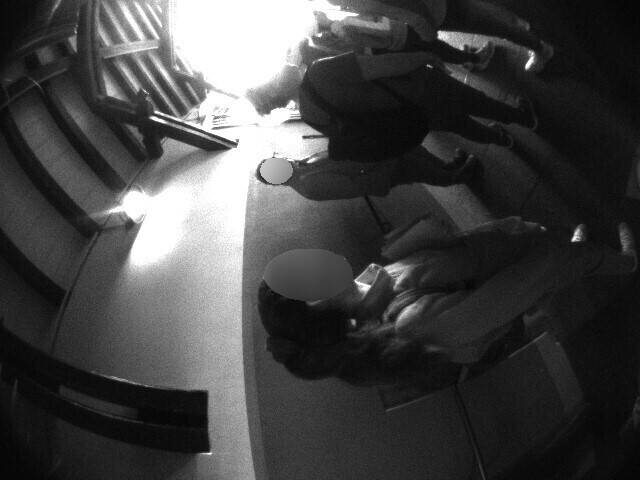} & 
\includegraphics[height=0.108\linewidth,angle=-90]{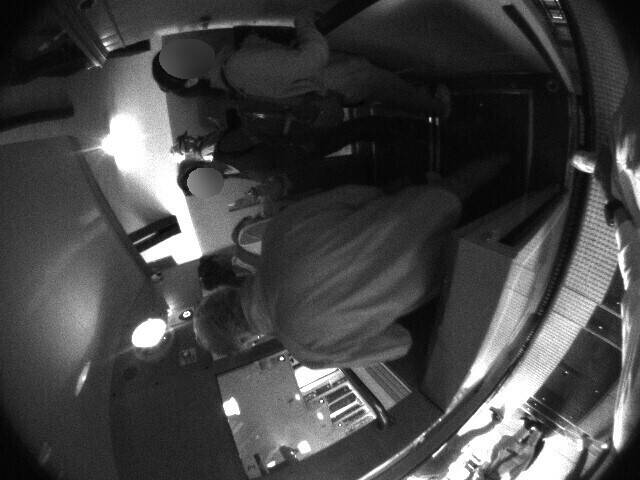} & 
\includegraphics[height=0.108\linewidth,angle=-90]{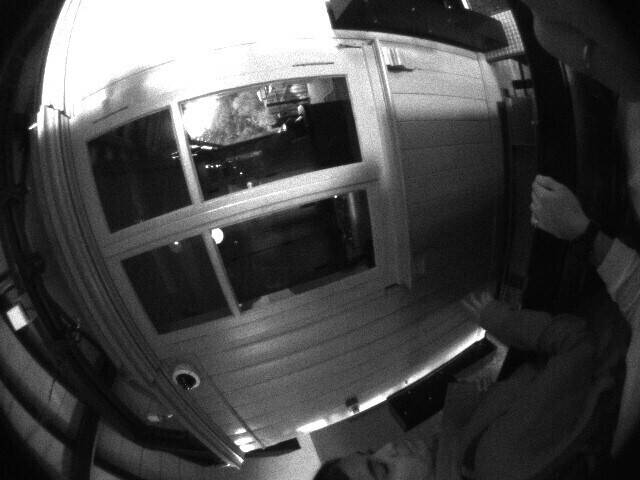} & 
\includegraphics[height=0.108\linewidth,angle=-90]{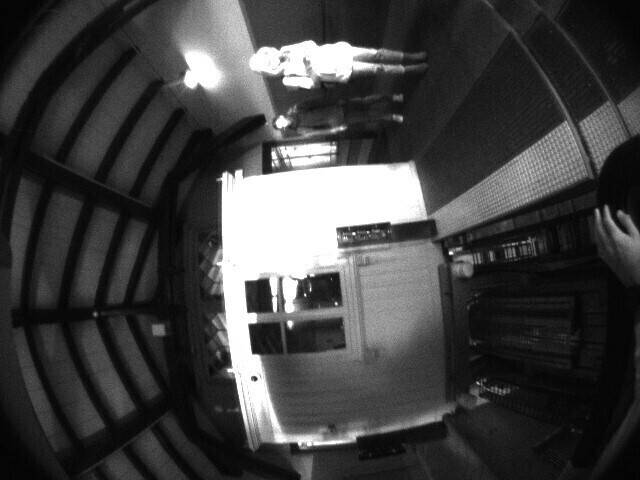} & 
\includegraphics[height=0.108\linewidth,angle=-90]{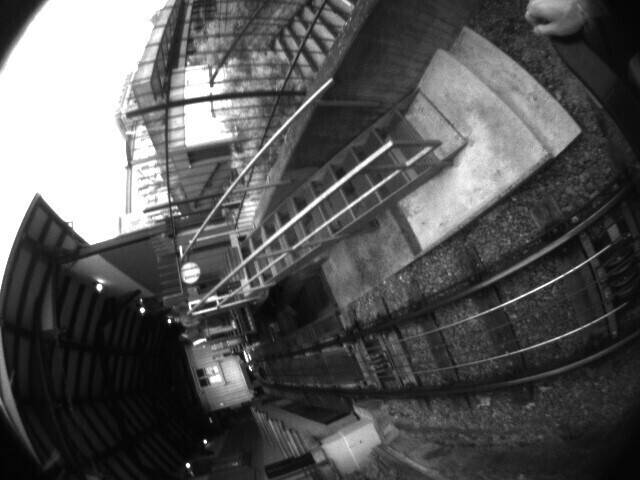} & 
\includegraphics[height=0.108\linewidth,angle=-90]{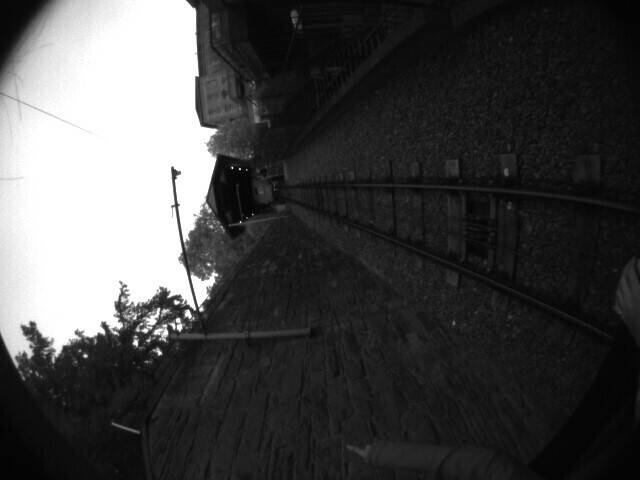} \\ [-5pt]

\includegraphics[height=0.108\linewidth,angle=-90]{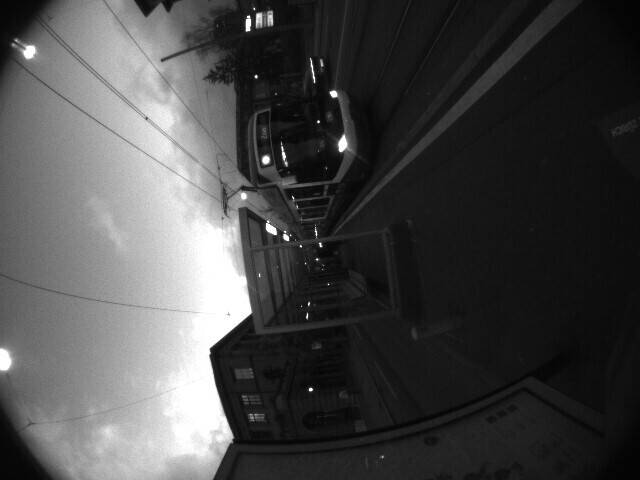} & 
\includegraphics[height=0.108\linewidth,angle=-90]{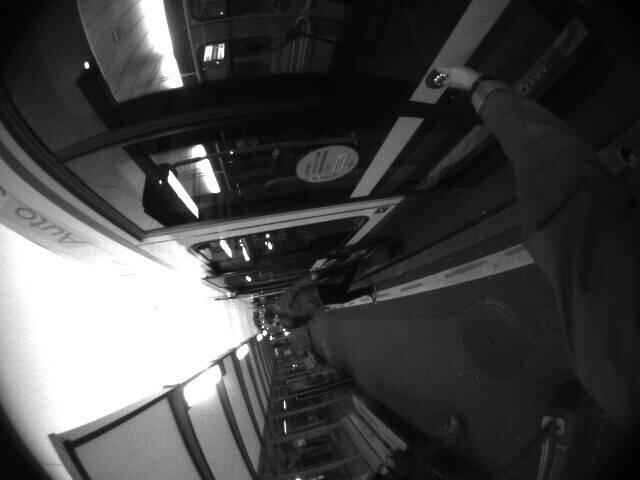} & 
\includegraphics[height=0.108\linewidth,angle=-90]{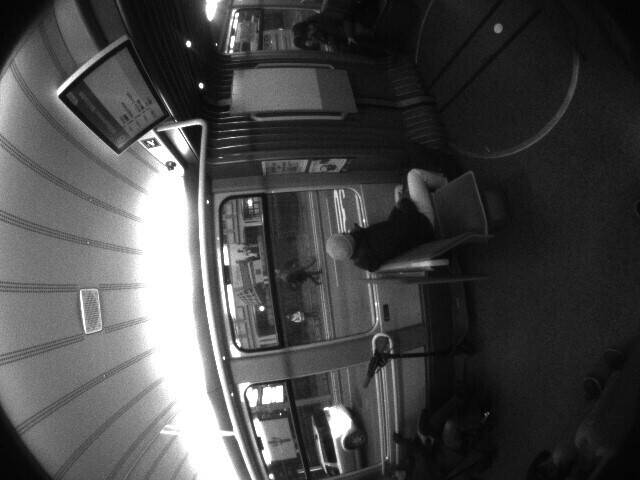} & 
\includegraphics[height=0.108\linewidth,angle=-90]{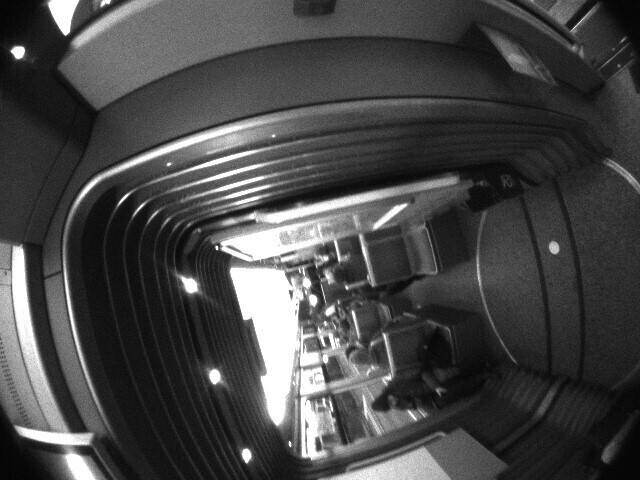} & 
\includegraphics[height=0.108\linewidth,angle=-90]{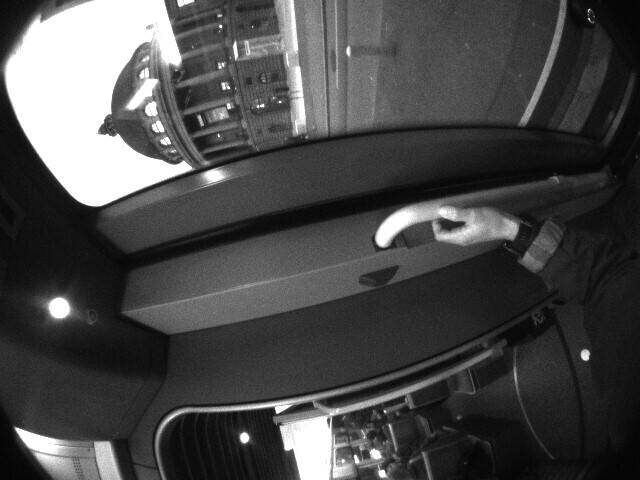} & 
\includegraphics[height=0.108\linewidth,angle=-90]{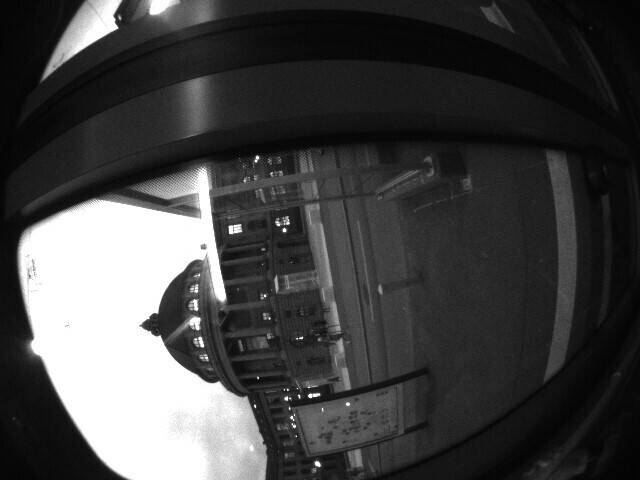} & 
\includegraphics[height=0.108\linewidth,angle=-90]{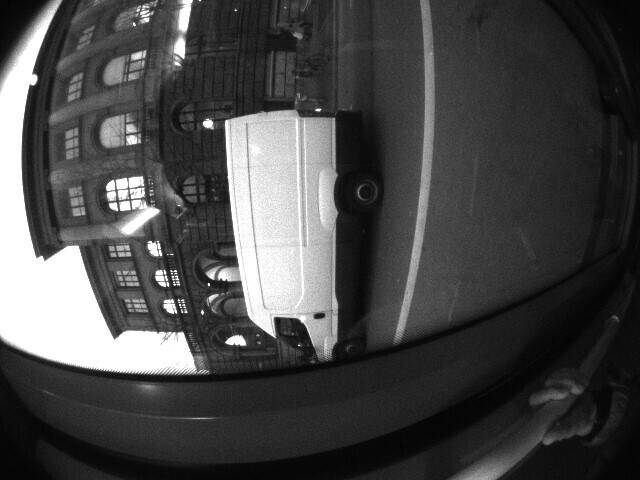} & 
\includegraphics[height=0.108\linewidth,angle=-90]{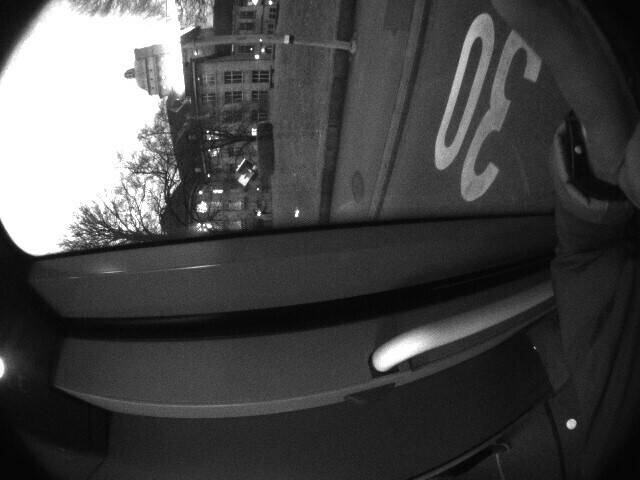} \\ [-5pt]

\includegraphics[height=0.108\linewidth,angle=-90]{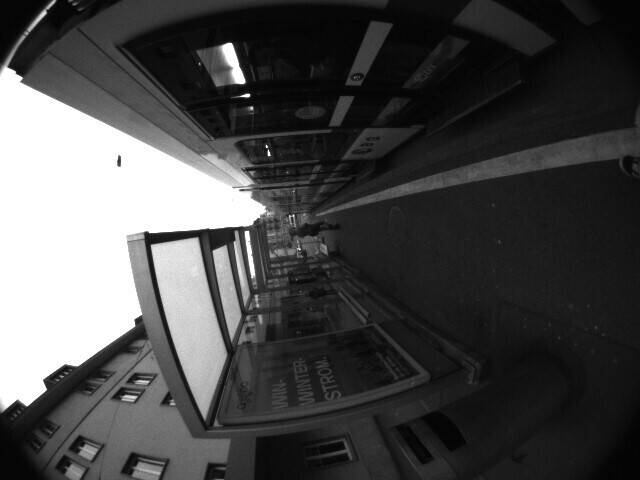} & 
\includegraphics[height=0.108\linewidth,angle=-90]{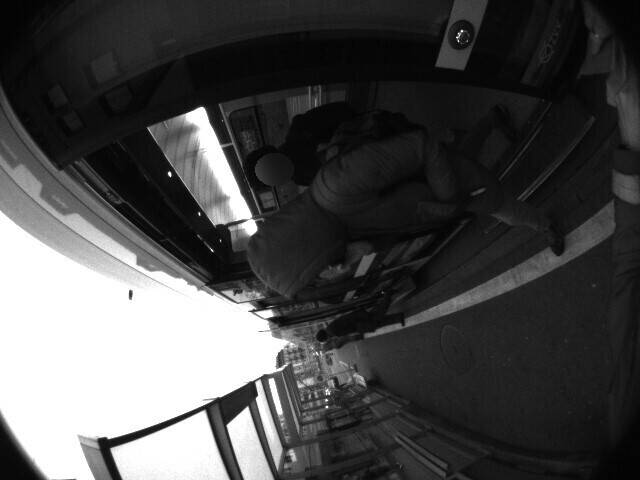} & 
\includegraphics[height=0.108\linewidth,angle=-90]{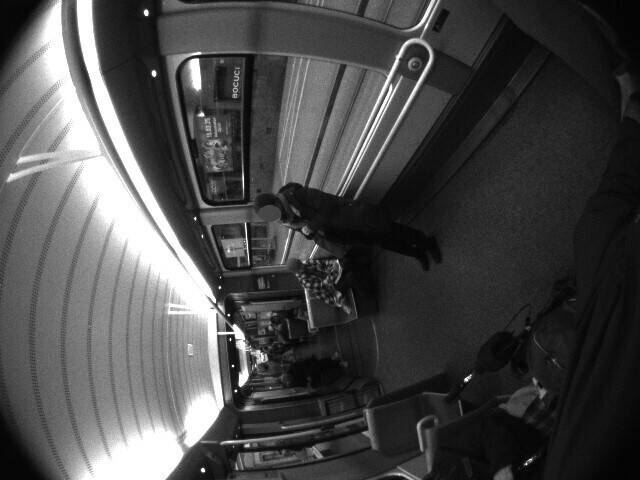} & 
\includegraphics[height=0.108\linewidth,angle=-90]{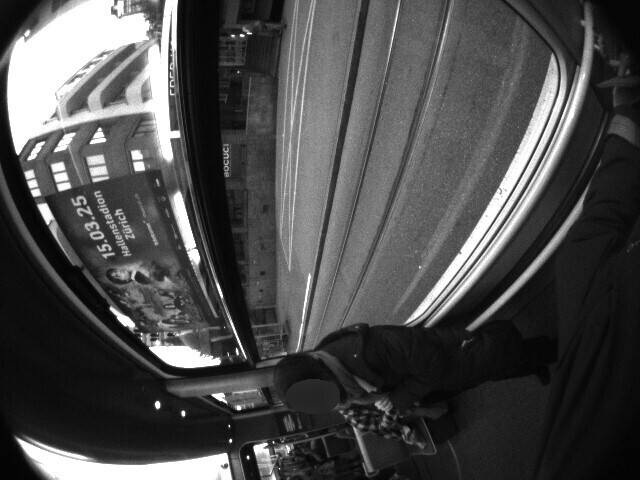} & 
\includegraphics[height=0.108\linewidth,angle=-90]{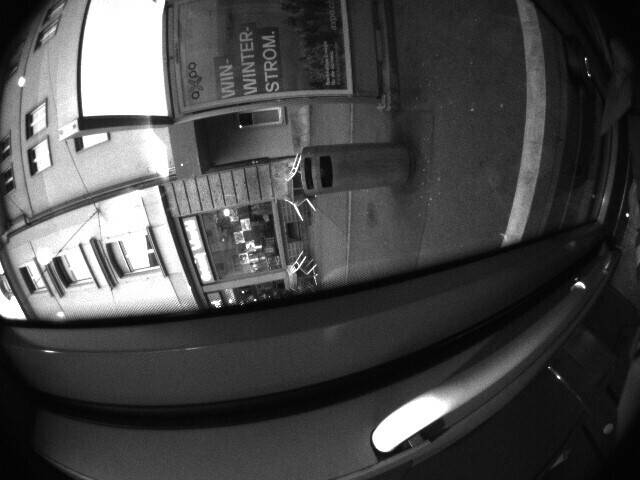} & 
\includegraphics[height=0.108\linewidth,angle=-90]{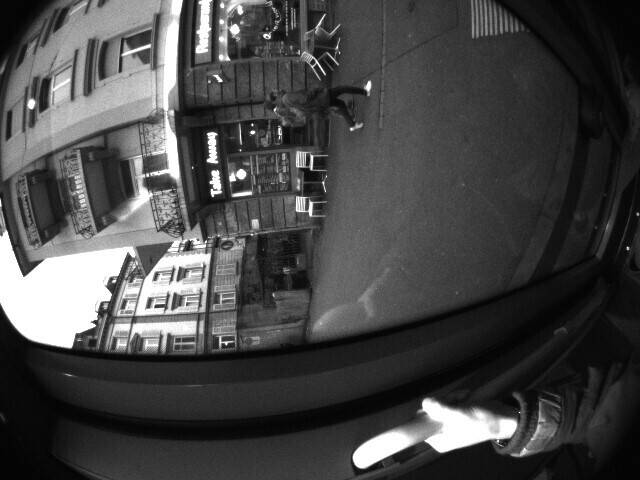} & 
\includegraphics[height=0.108\linewidth,angle=-90]{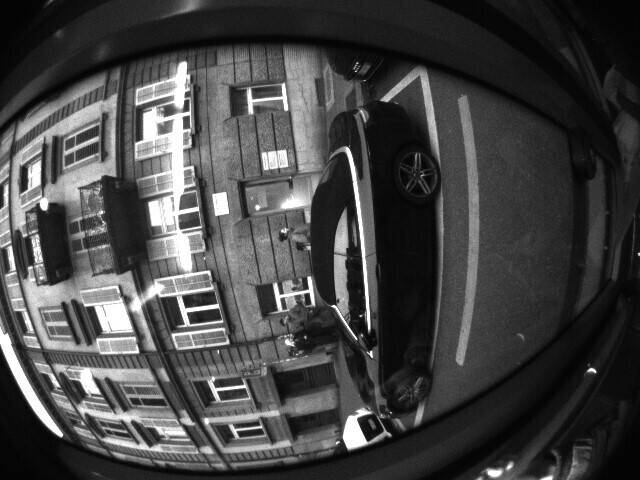} & 
\includegraphics[height=0.108\linewidth,angle=-90]{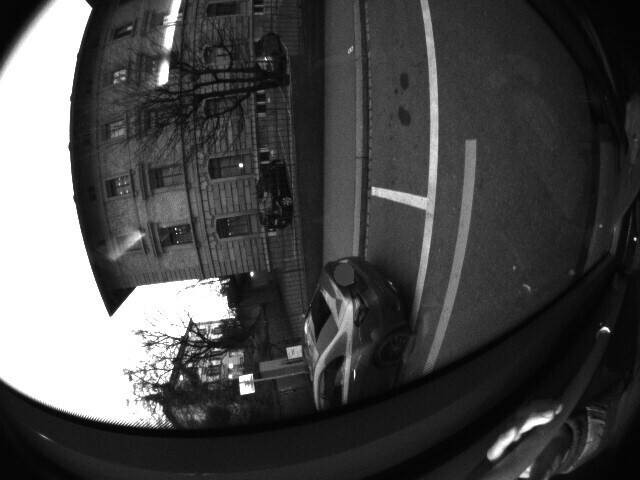} \\ [-5pt]

\includegraphics[height=0.108\linewidth,angle=-90]{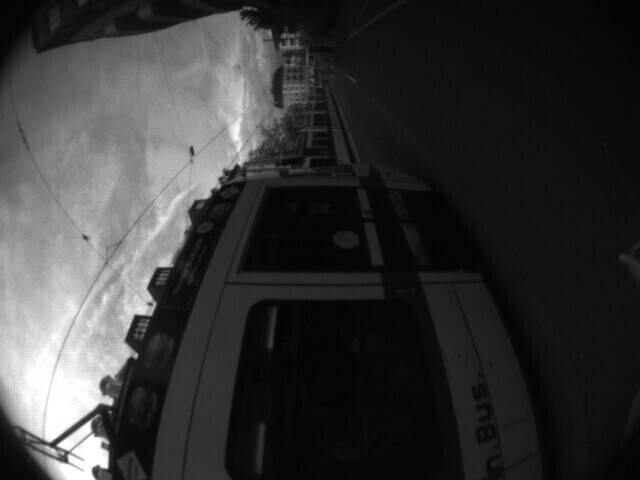} & 
\includegraphics[height=0.108\linewidth,angle=-90]{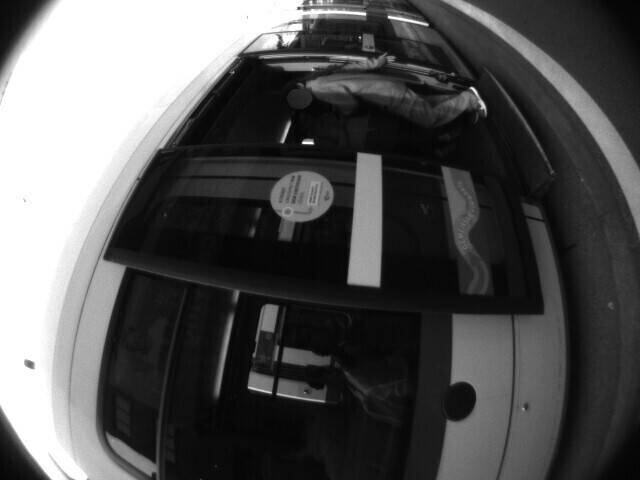} & 
\includegraphics[height=0.108\linewidth,angle=-90]{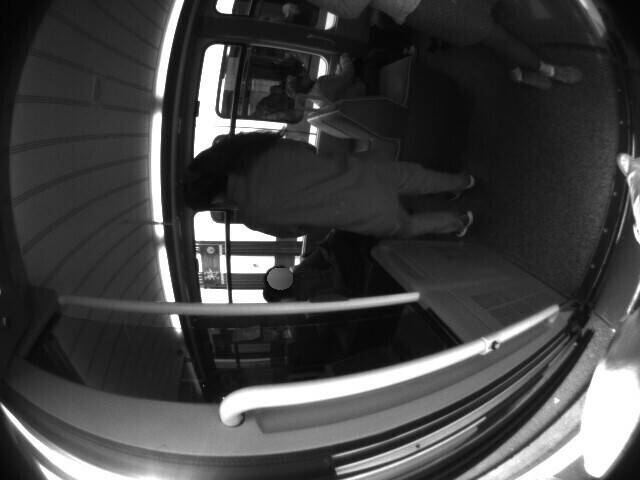} & 
\includegraphics[height=0.108\linewidth,angle=-90]{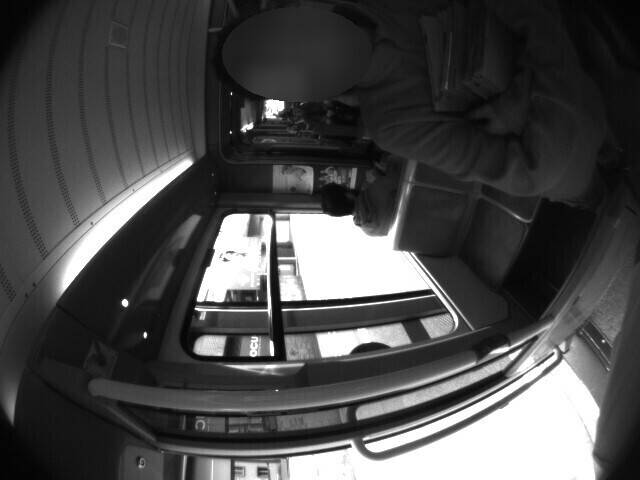} & 
\includegraphics[height=0.108\linewidth,angle=-90]{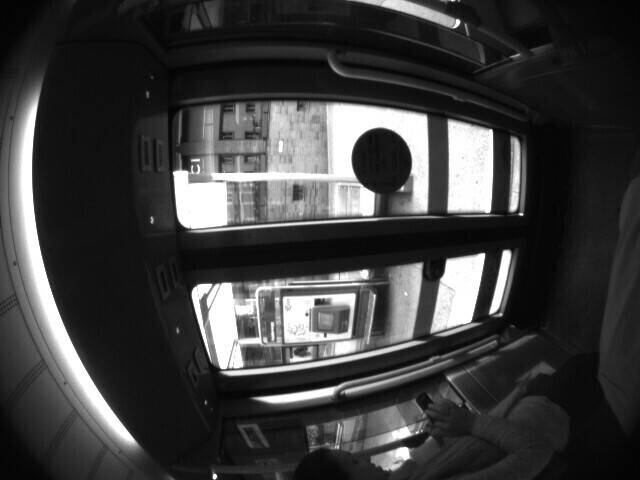} & 
\includegraphics[height=0.108\linewidth,angle=-90]{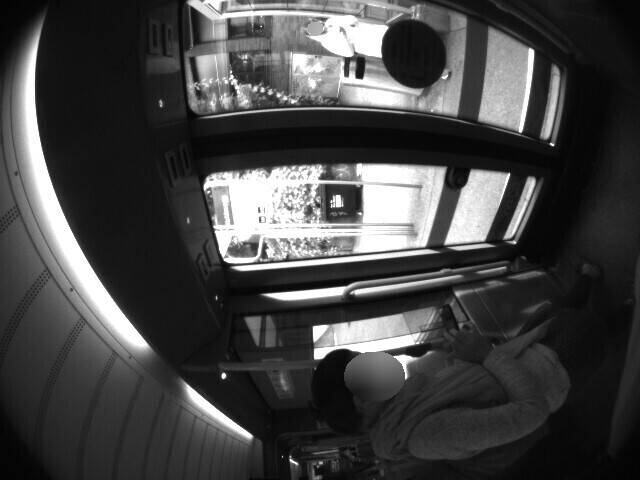} & 
\includegraphics[height=0.108\linewidth,angle=-90]{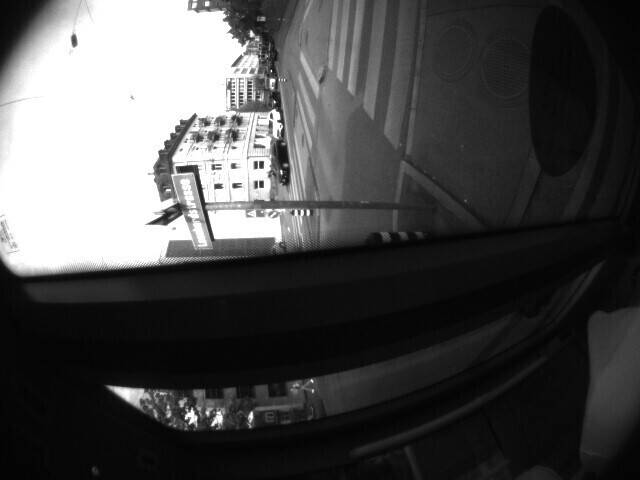} & 
\includegraphics[height=0.108\linewidth,angle=-90]{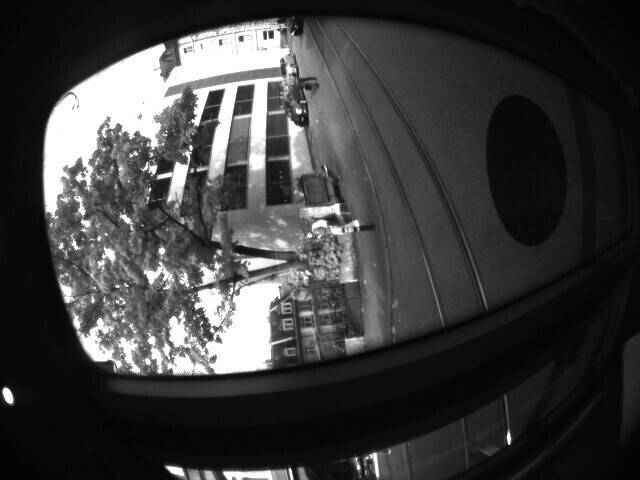} \\
\end{tabular}
\centering
\caption{\textbf{Visualizations of the egocentric recordings in our dataset.}}
\label{fig:supp_examples_full_set_2}%
\end{figure*}

\begin{figure*}[tb]
\centering
\setlength\tabcolsep{1pt}
\begin{tabular}{cccccccc}
\multicolumn{8}{c}{\textbf{long outdoor trajectories}} \\ [-5pt]
\includegraphics[height=0.108\linewidth,angle=-90]{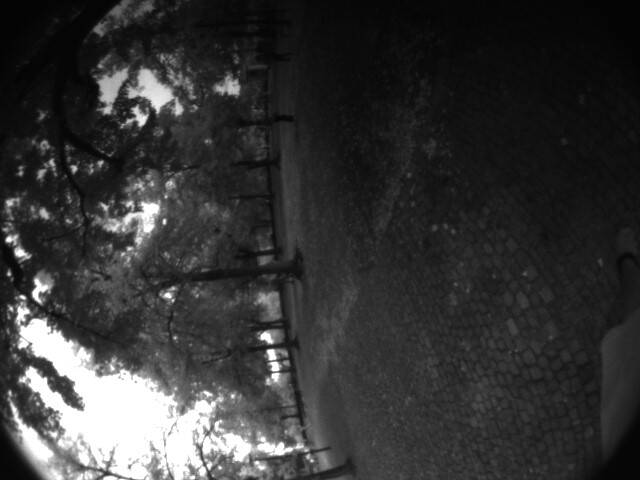} & 
\includegraphics[height=0.108\linewidth,angle=-90]{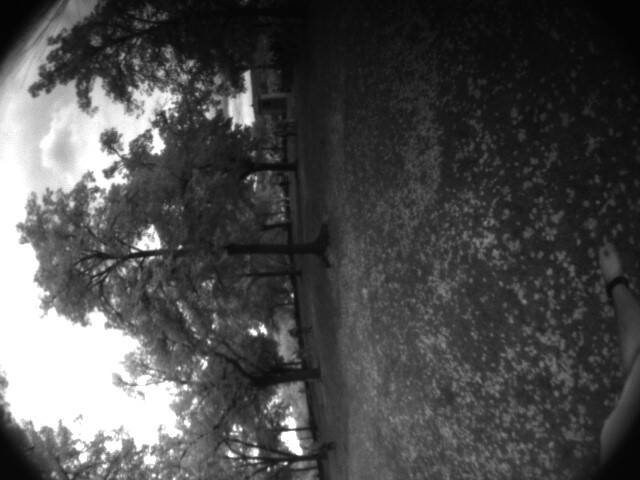} & 
\includegraphics[height=0.108\linewidth,angle=-90]{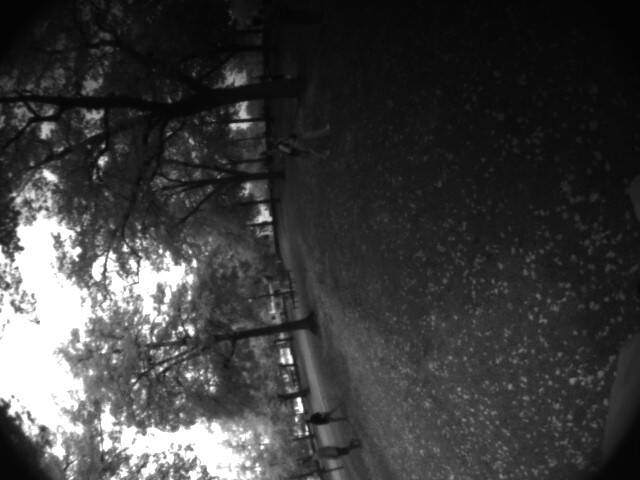} & 
\includegraphics[height=0.108\linewidth,angle=-90]{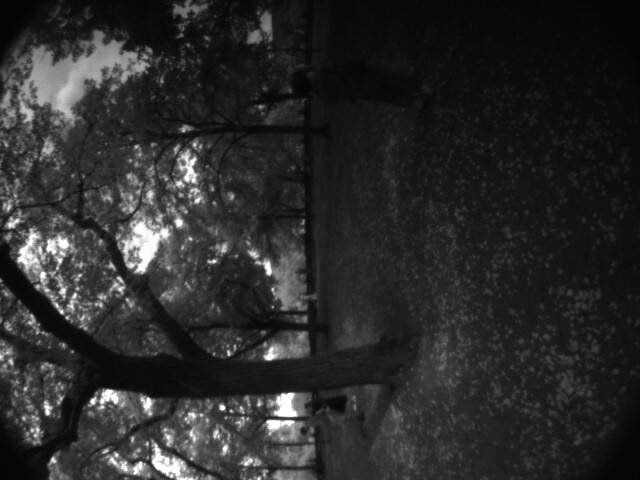} & 
\includegraphics[height=0.108\linewidth,angle=-90]{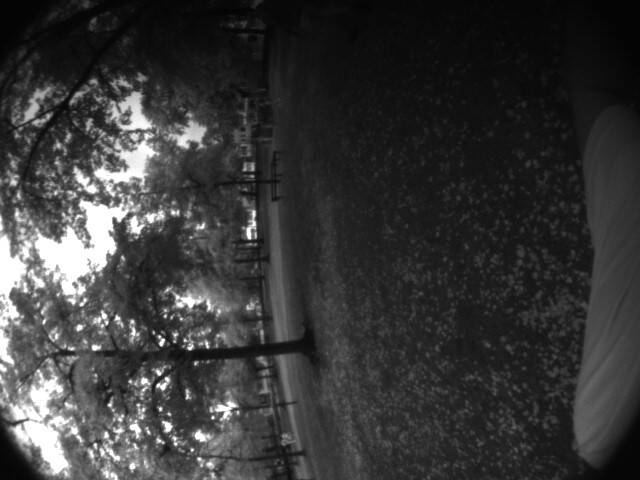} & 
\includegraphics[height=0.108\linewidth,angle=-90]{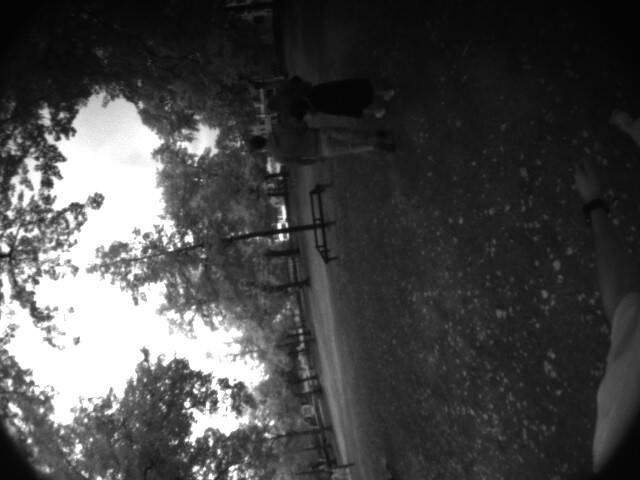} & 
\includegraphics[height=0.108\linewidth,angle=-90]{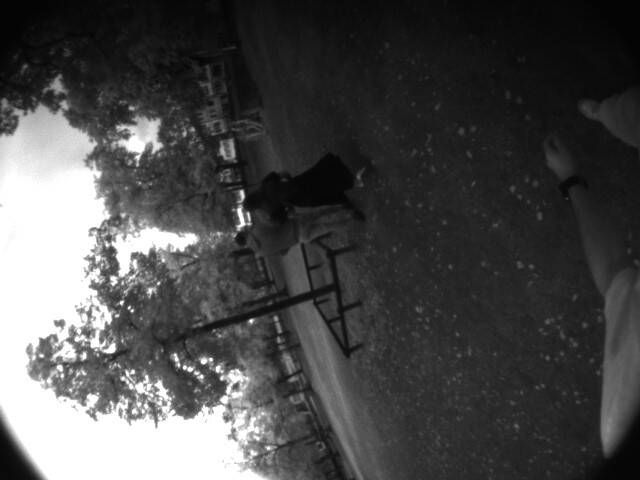} & 
\includegraphics[height=0.108\linewidth,angle=-90]{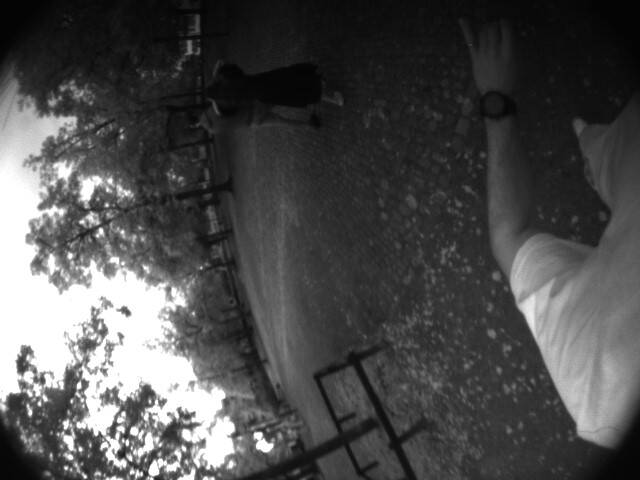} \\ [-5pt]

\includegraphics[height=0.108\linewidth,angle=-90]{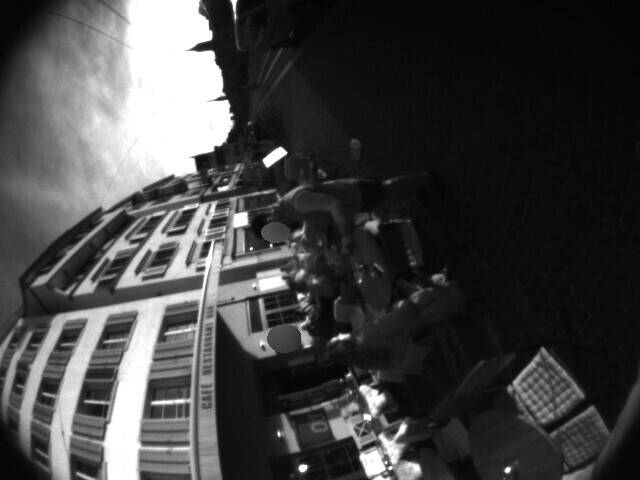} & 
\includegraphics[height=0.108\linewidth,angle=-90]{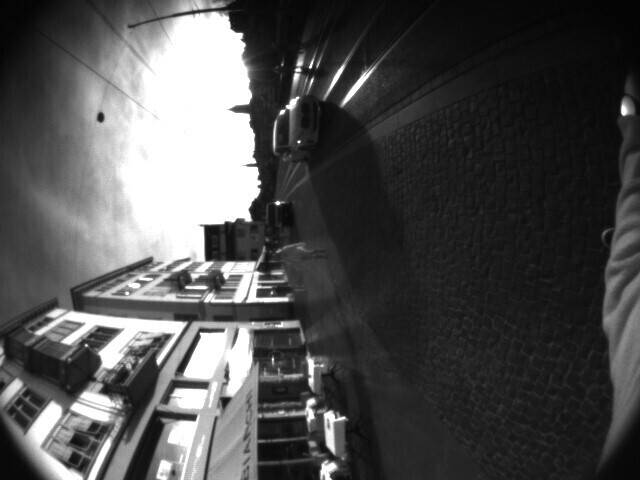} & 
\includegraphics[height=0.108\linewidth,angle=-90]{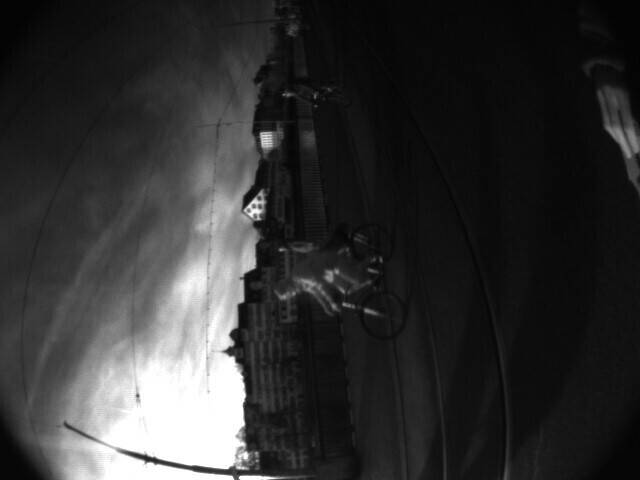} & 
\includegraphics[height=0.108\linewidth,angle=-90]{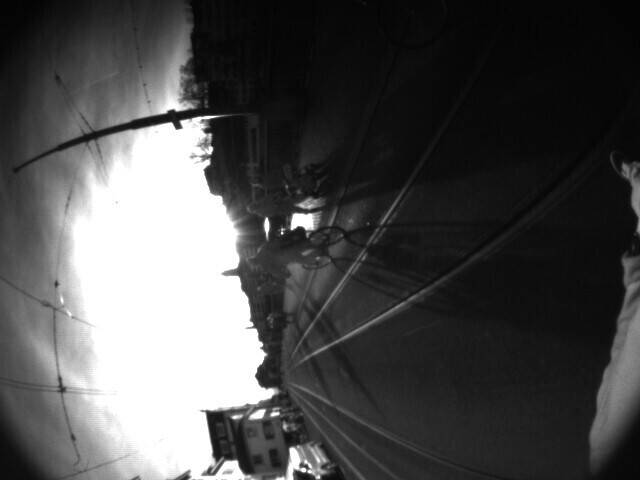} & 
\includegraphics[height=0.108\linewidth,angle=-90]{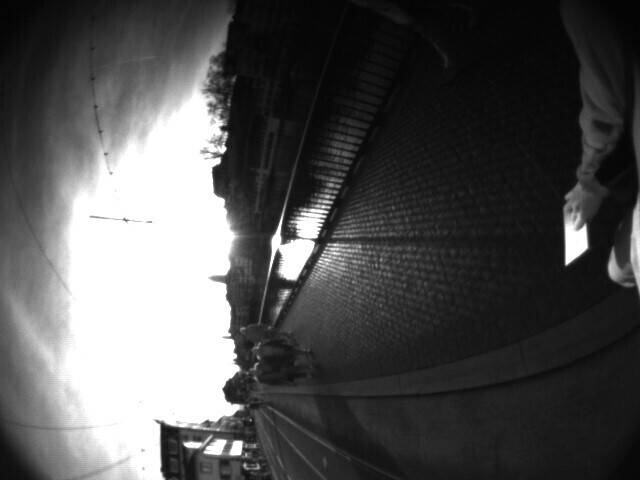} & 
\includegraphics[height=0.108\linewidth,angle=-90]{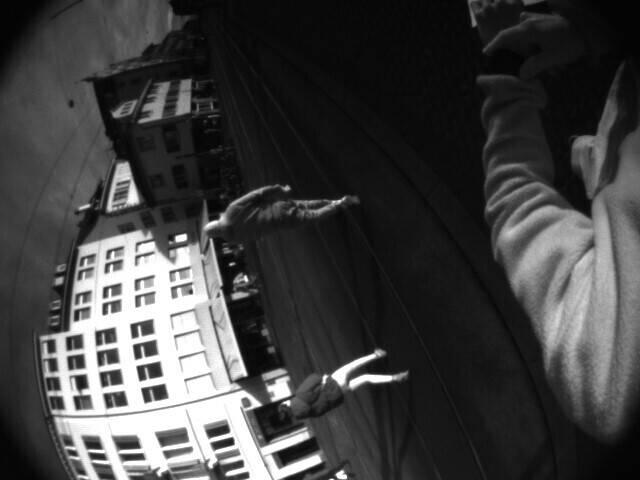} & 
\includegraphics[height=0.108\linewidth,angle=-90]{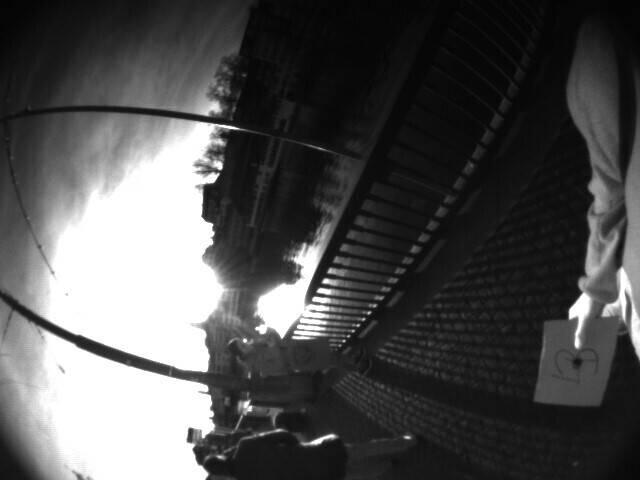} & 
\includegraphics[height=0.108\linewidth,angle=-90]{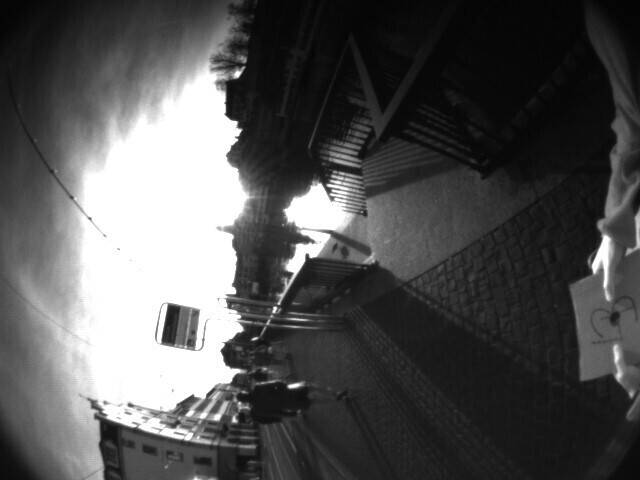} \\ [-5pt]

\includegraphics[height=0.108\linewidth,angle=-90]{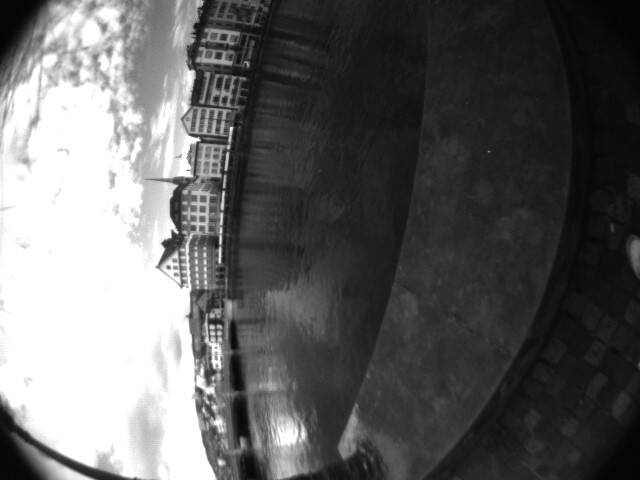} & 
\includegraphics[height=0.108\linewidth,angle=-90]{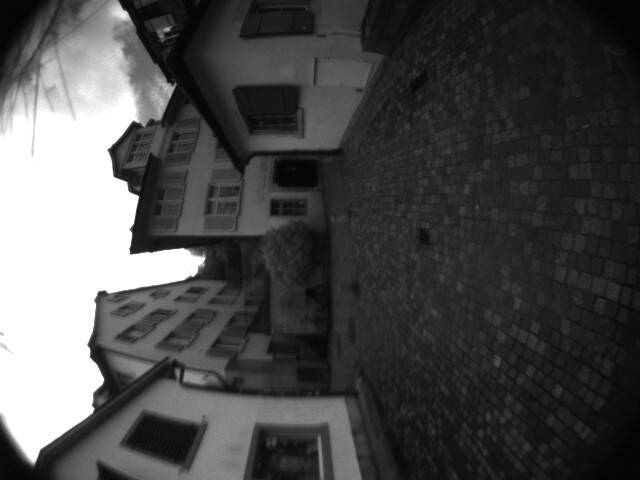} & 
\includegraphics[height=0.108\linewidth,angle=-90]{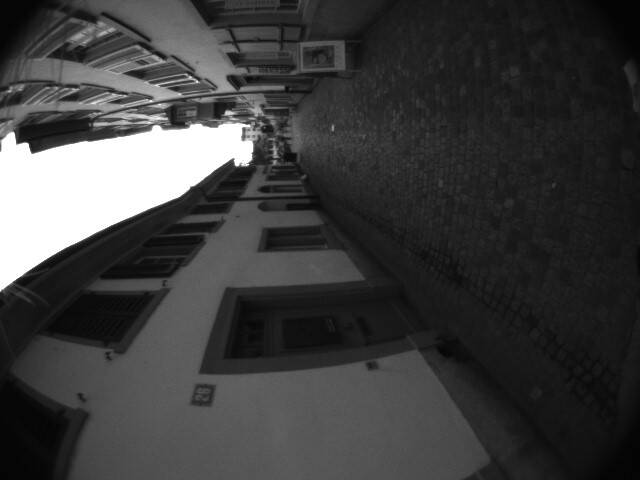} & 
\includegraphics[height=0.108\linewidth,angle=-90]{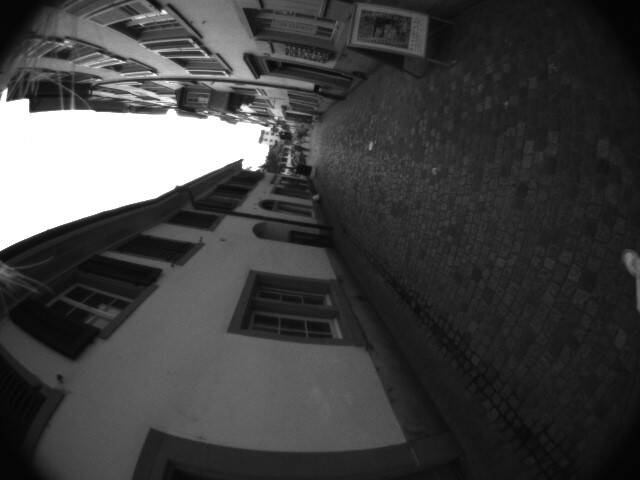} & 
\includegraphics[height=0.108\linewidth,angle=-90]{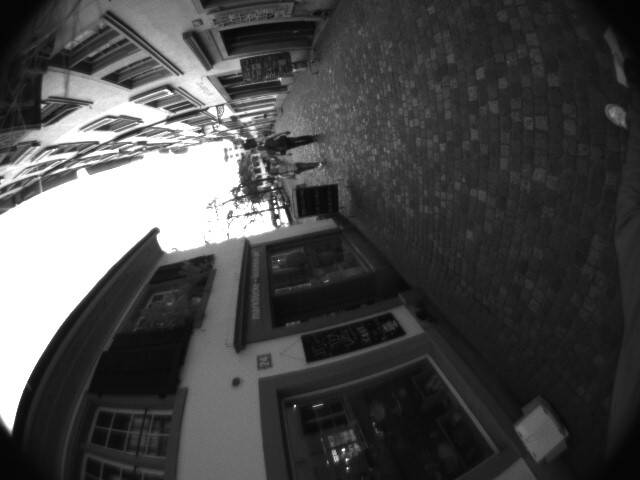} & 
\includegraphics[height=0.108\linewidth,angle=-90]{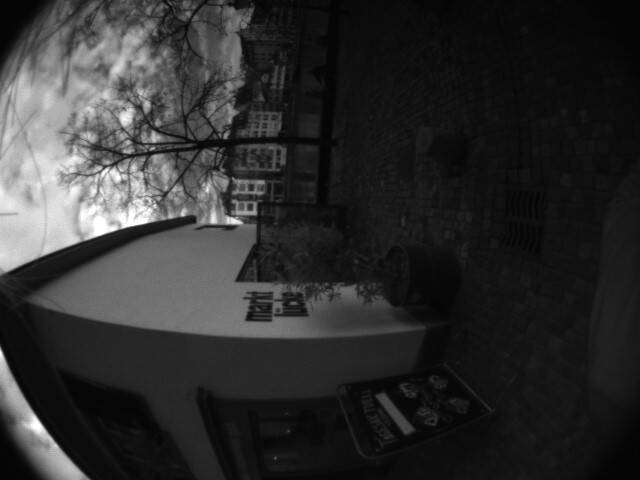} & 
\includegraphics[height=0.108\linewidth,angle=-90]{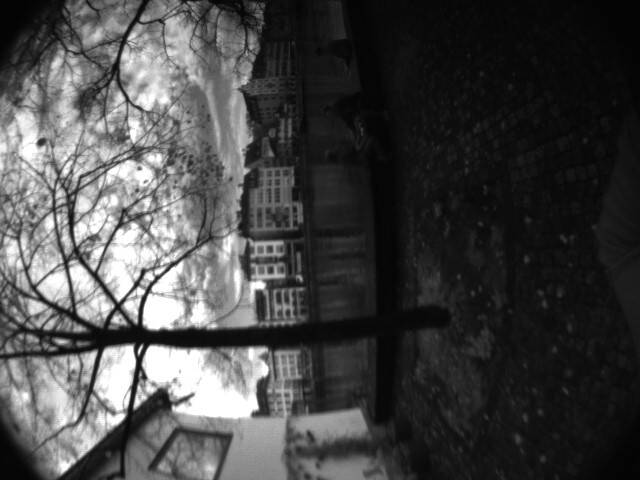} & 
\includegraphics[height=0.108\linewidth,angle=-90]{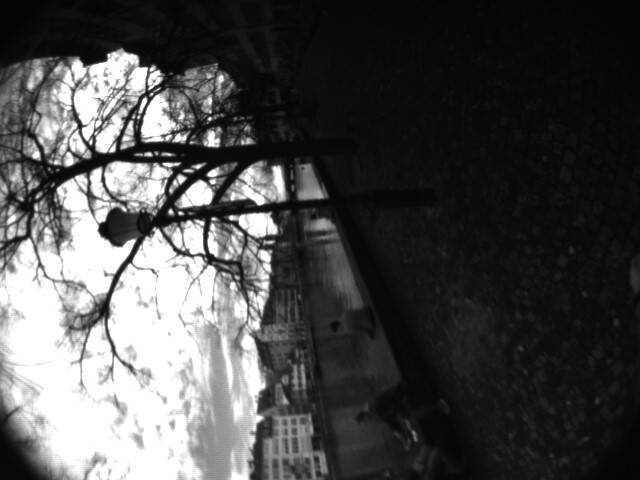} \\ [-5pt]

\includegraphics[height=0.108\linewidth,angle=-90]{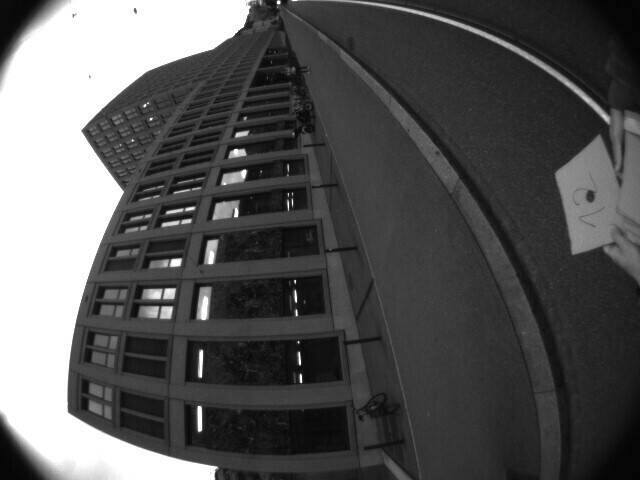} & 
\includegraphics[height=0.108\linewidth,angle=-90]{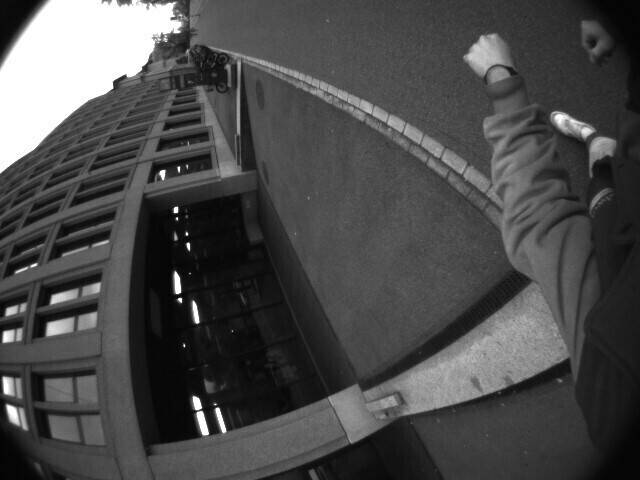} & 
\includegraphics[height=0.108\linewidth,angle=-90]{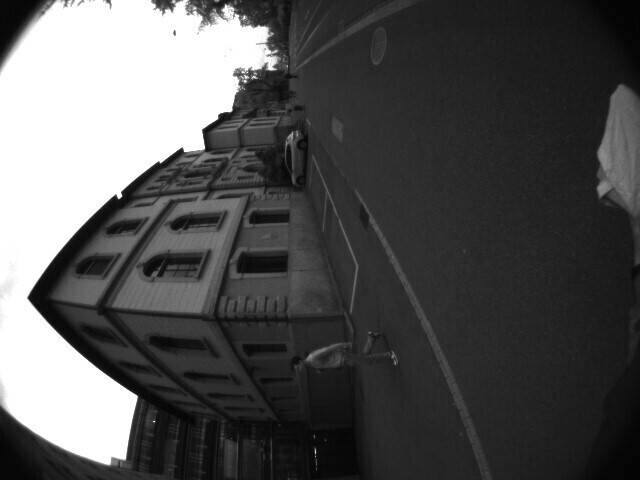} & 
\includegraphics[height=0.108\linewidth,angle=-90]{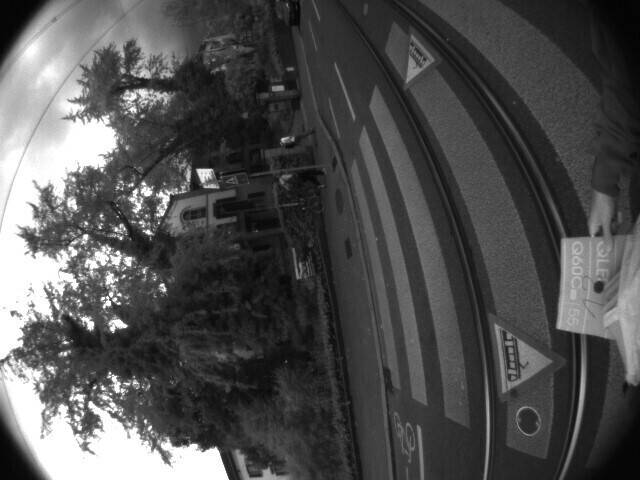} & 
\includegraphics[height=0.108\linewidth,angle=-90]{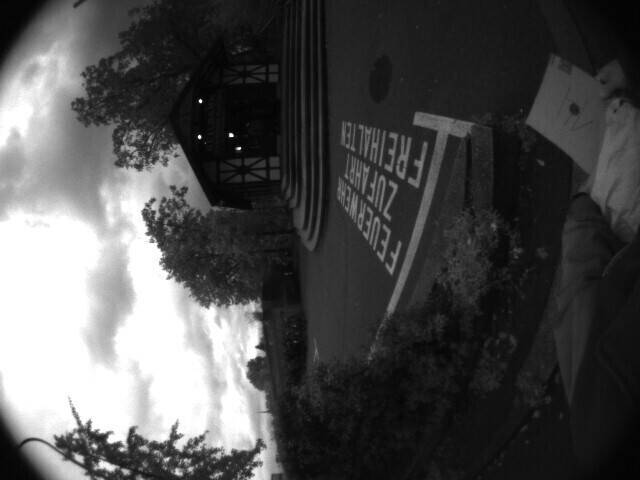} & 
\includegraphics[height=0.108\linewidth,angle=-90]{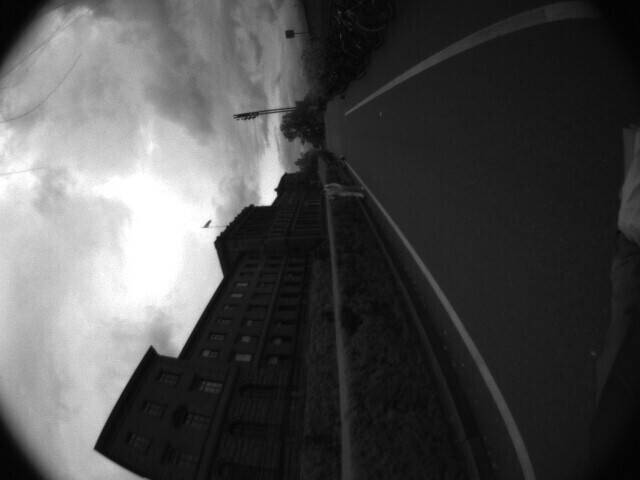} & 
\includegraphics[height=0.108\linewidth,angle=-90]{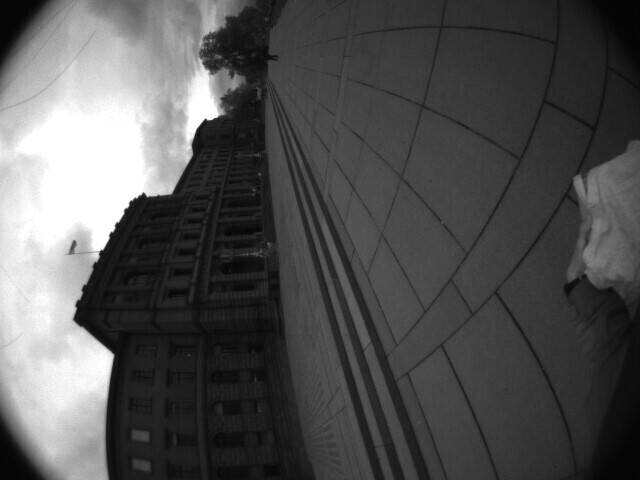} & 
\includegraphics[height=0.108\linewidth,angle=-90]{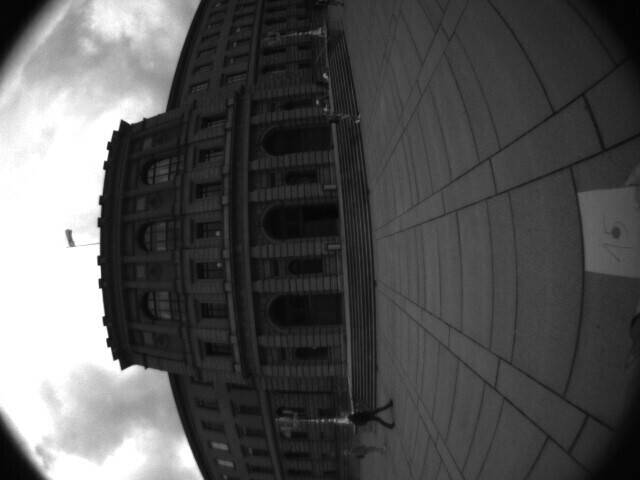} \\ [-5pt]

\includegraphics[height=0.108\linewidth,angle=-90]{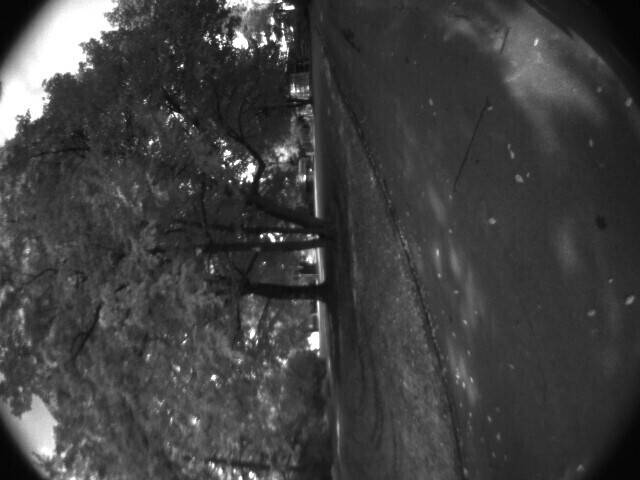} & 
\includegraphics[height=0.108\linewidth,angle=-90]{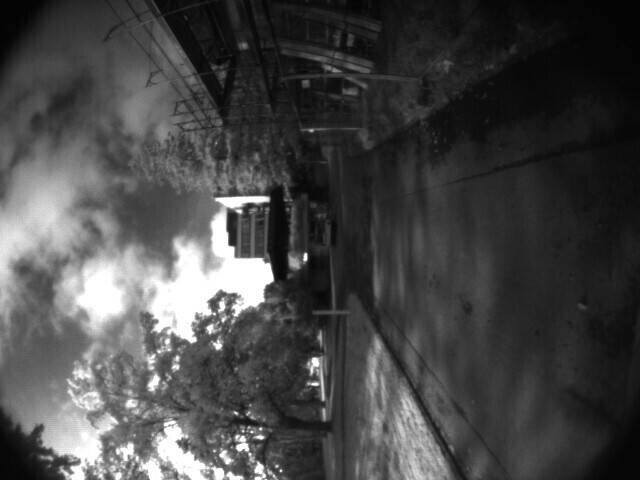} & 
\includegraphics[height=0.108\linewidth,angle=-90]{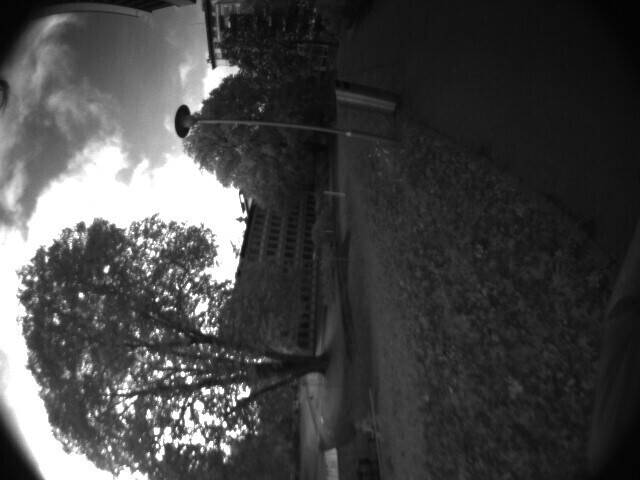} & 
\includegraphics[height=0.108\linewidth,angle=-90]{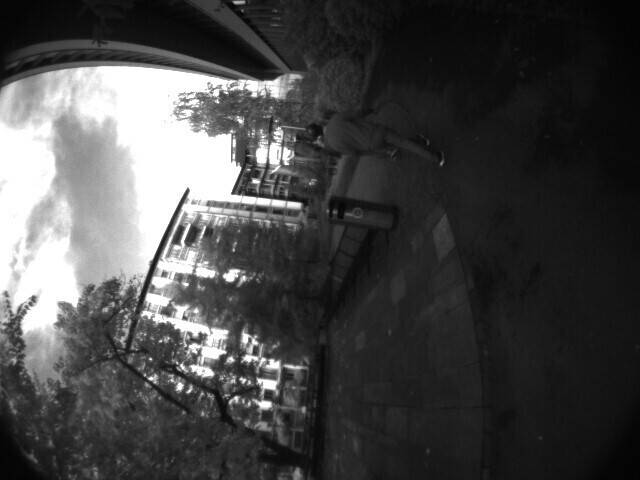} & 
\includegraphics[height=0.108\linewidth,angle=-90]{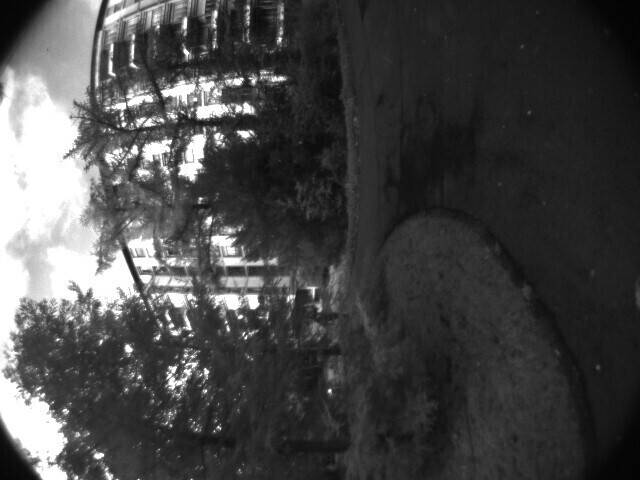} & 
\includegraphics[height=0.108\linewidth,angle=-90]{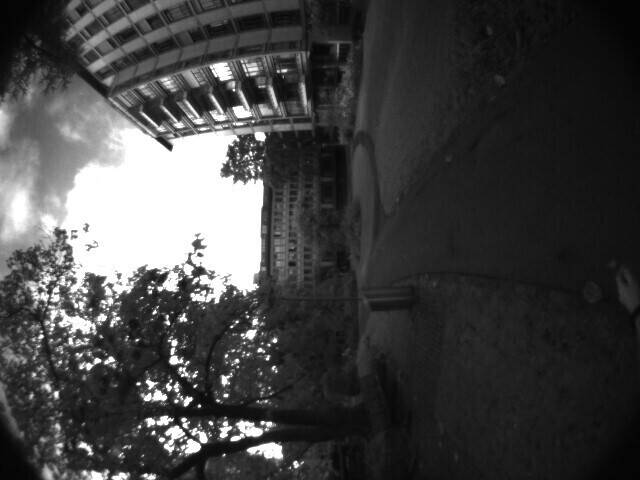} & 
\includegraphics[height=0.108\linewidth,angle=-90]{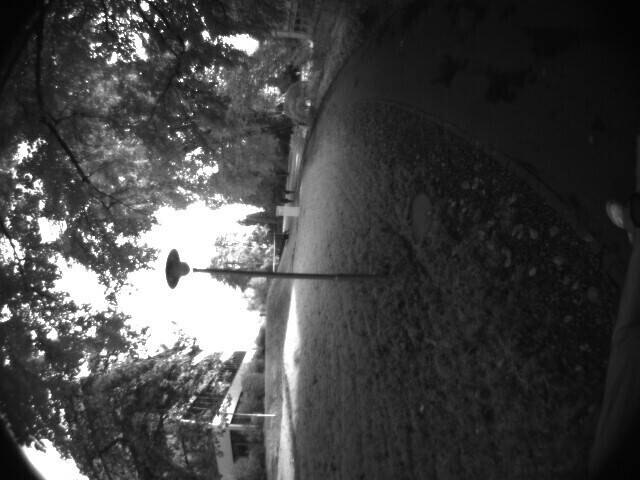} & 
\includegraphics[height=0.108\linewidth,angle=-90]{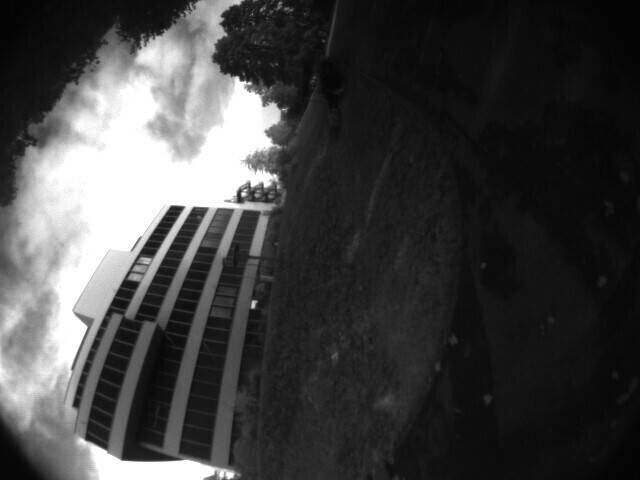} \\ [-5pt]

\includegraphics[height=0.108\linewidth,angle=-90]{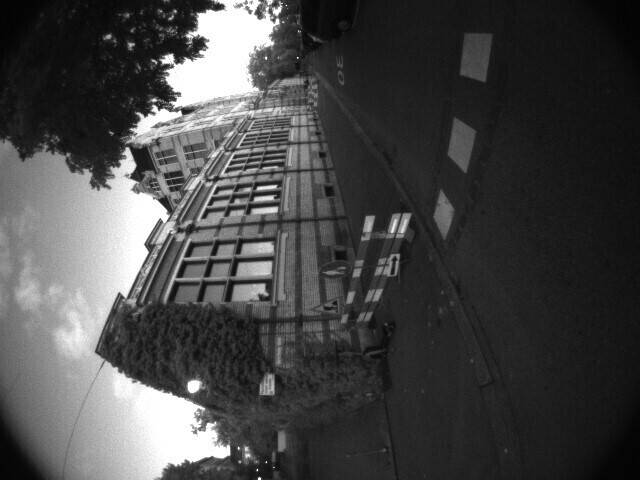} & 
\includegraphics[height=0.108\linewidth,angle=-90]{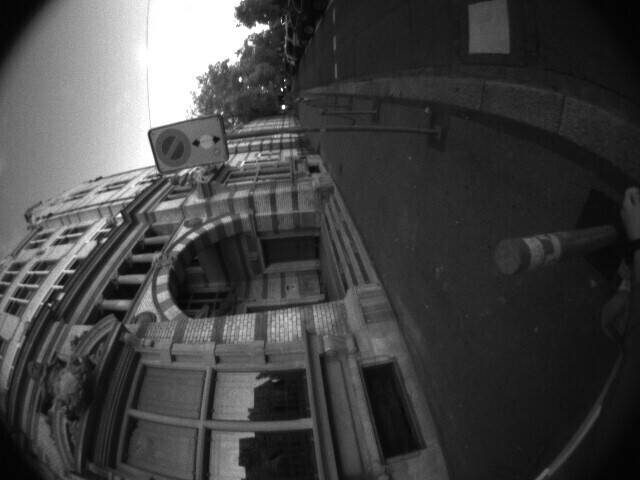} & 
\includegraphics[height=0.108\linewidth,angle=-90]{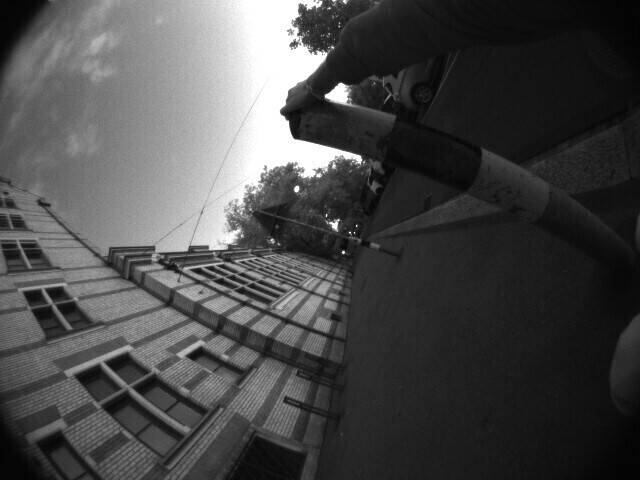} & 
\includegraphics[height=0.108\linewidth,angle=-90]{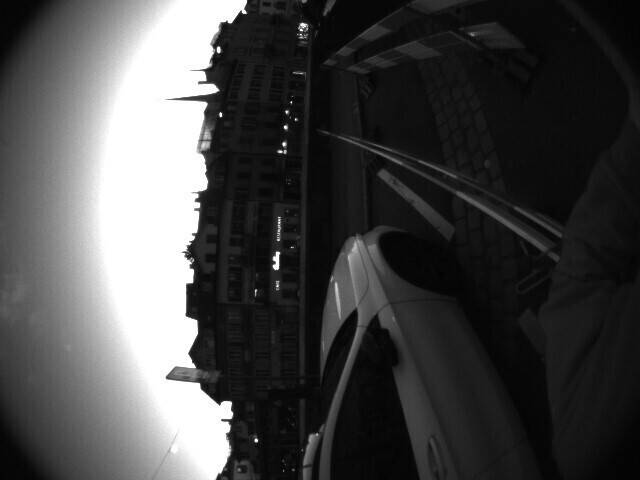} & 
\includegraphics[height=0.108\linewidth,angle=-90]{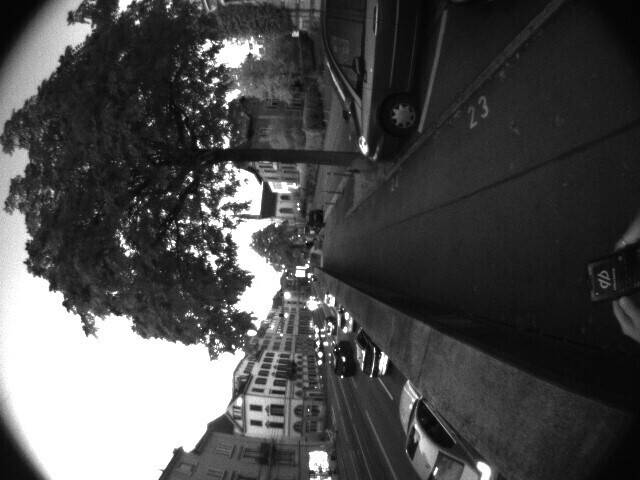} & 
\includegraphics[height=0.108\linewidth,angle=-90]{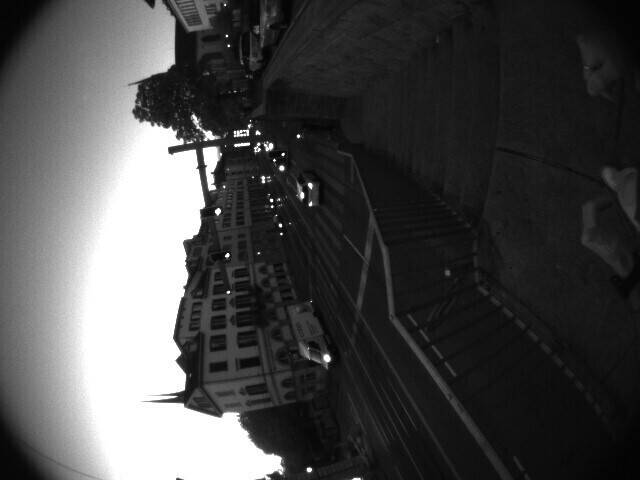} & 
\includegraphics[height=0.108\linewidth,angle=-90]{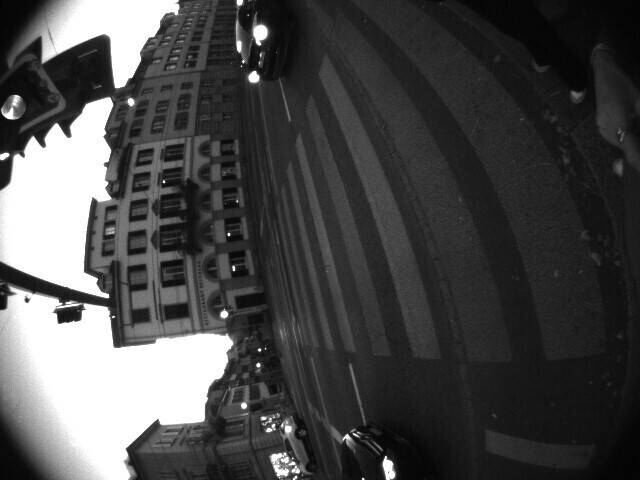} & 
\includegraphics[height=0.108\linewidth,angle=-90]{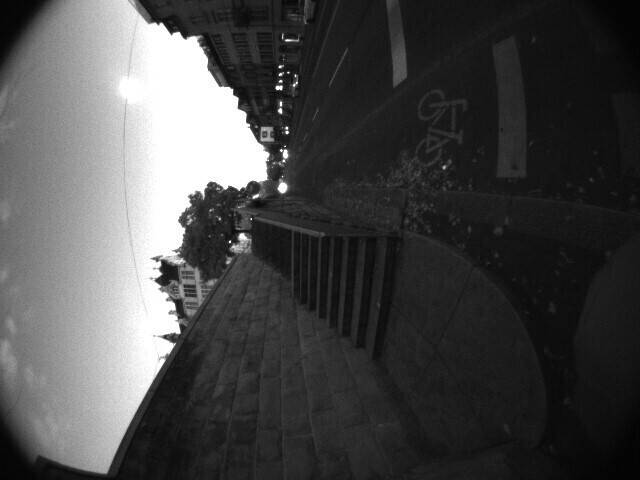} \\
\end{tabular}
\centering
\caption{\textbf{Visualizations of the egocentric recordings in our dataset.}}
\label{fig:supp_examples_full_set_3}%
\end{figure*}

\clearpage
{
    \small
    \bibliographystyle{ieeenat_fullname}
    \bibliography{abbreviations,main}
}

\end{document}